\documentclass{article}
\usepackage{microtype}
\usepackage{graphicx}
\usepackage{booktabs}
\usepackage{hyperref}

\usepackage[preprint]{icml2026}
\usepackage[T1]{fontenc}
\usepackage{amsmath,amssymb,mathtools,amsthm}
\usepackage{tabularx}
\usepackage{xltabular}
\usepackage{placeins}
\usepackage{multirow}
\usepackage{enumitem}
\usepackage[capitalize,noabbrev]{cleveref}

\newtheorem{proposition}{Proposition}
\newcommand{\E}{\mathbb{E}}
\newcommand{\KL}{\mathrm{KL}}
\newcommand{\cL}{\mathcal{L}}
\newcommand{\bpc}{\,bpc}
\setlist{leftmargin=1.1em,itemsep=1pt,topsep=2pt,parsep=0pt}

\icmltitlerunning{Train the Sampler You Run}
\hypersetup{pdftitle={Train the Sampler You Run: Few-Step Masked Diffusion Language Models via Mixture-Likelihood Fine-Tuning}, pdfauthor={Amin Najafgholizadeh}, pdfsubject={}}

\begin{document}

\twocolumn[
  \icmltitle{Train the Sampler You Run: Few-Step Masked Diffusion Language Models\\ via Mixture-Likelihood Fine-Tuning}
  \begin{icmlauthorlist}
    \icmlauthor{Amin Najafgholizadeh}{znu}
  \end{icmlauthorlist}
  \icmlaffiliation{znu}{University of Zanjan}
  \icmlcorrespondingauthor{Amin Najafgholizadeh}{amin.ngz@znu.ac.ir}
  \icmlkeywords{Masked diffusion, discrete diffusion, language models, few-step sampling, importance weighting}
  \vskip 0.3in
]

\printAffiliationsAndNotice{}

\begin{abstract}
When a masked diffusion language model is sampled in a few parallel steps in a random order, the model actually run is a mixture over reveal plans, while cross-entropy training fits token conditionals rather than this mixture.
We introduce mixture-likelihood fine-tuning (Mix-FT), which fine-tunes a pretrained model on an importance-weighted bound of its own sampler's likelihood, with no teacher and no change to the architecture or the sampler.
On LLaDA-8B, a LoRA fine-tune for 0.002\% of the pretraining compute writes better OpenWebText continuations than continued cross-entropy and Alpha diffusion given the same compute, at every tested step count.
With half the model evaluations, 16 rather than 32 steps per block, its continuations score better than the original model's in every seed under three judges, with confidence-based decoding as well as random order, and under confidence-based decoding two blinded LLM raters from different model families prefer them in 69 and 71\% of pairs.
On a public 110M-parameter block diffusion model, Mix-FT takes under four T4-hours and beats continued cross-entropy, Alpha diffusion and SDTT distillation at 4, 8 and 16 steps per block, under two judges in every seed.
Exact likelihoods on text8, with a control that weights every plan equally, attribute about three quarters of the likelihood gain over continued cross-entropy to weighting each plan by how well it explains the sequence.
\end{abstract}

\vspace{0.5ex}

\section{Introduction}
\label{sec:intro}

Masked diffusion language models \citep{sahoo2024mdlm,shi2024md4,lou2024sedd,nie2025llada} learn to fill masked positions given the revealed ones. They are attractive because they can be sampled in a few parallel steps: a block of $B$ positions is filled in $K<B$ steps, each revealing $B/K$ positions drawn independently given what is already revealed. The factorization error this causes is well known. Current fixes act at sampling time, through temperature \citep{amin2026simple} or better schedules \citep{luxembourg2025dilated,benhamu2025ebsampler}, add heads or latents \citep{kim2026tensortrain,xie2026vadd,ivanov2026emoe}, or distill from a many-step teacher \citep{deschenaux2024sdtt,hayakawa2025di4c,zhang2026t3d}.

\begin{figure}[t]
\centering
\includegraphics[width=\linewidth]{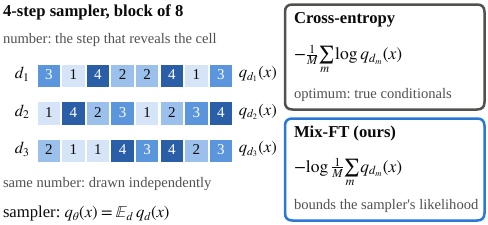}
\caption{A $K$-step random-order sampler draws a plan, then fills each group of positions independently given the groups before it. Its law is the average of the per-plan products. CE averages the per-plan log-likelihoods and is minimized by the true conditionals. Mix-FT fine-tunes on an importance-weighted bound of the log of the average.}
\label{fig:method}
\end{figure}

We change the loss instead, and the objective follows from one observation. Under a random reveal order, the deployed sampler draws from a mixture over reveal plans $d$,
\[
q_\theta(x) \;=\; \textstyle\sum_{d} \mu(d)\, q_{\theta,d}(x),
\]
where each component is a product of factorized steps along plan $d$. The standard cross-entropy (CE) loss averages $\log q_{\theta,d}$ over plans; it separates over steps and is minimized by the true conditionals, which do not minimize $-\E_p\log q_\theta$. So we fine-tune a CE-trained model on the likelihood of the sampler we run, through an importance-weighted bound \citep{burda2016iwae} over $M$ random plans (\cref{fig:method}). We call the result Mix-FT.

This change of loss alone, for under 0.02\% of the pretraining compute, makes the same network with the same sampler write better text. On LLaDA-8B \citep{nie2025llada}, a LoRA fine-tune for 0.002\% of the pretraining compute beats continued cross-entropy and Alpha diffusion \citep{gushchin2026alpha} at every tested step count in every seed. With 16 steps per block, half the model evaluations, it is ahead of LLaDA's 32 in every seed under three judges, with LLaDA's confidence planner as well as random order, and under the confidence planner blinded LLM raters from two model families prefer its continuations in 69 and 71\% of pairs (\cref{sec:8b}). On the public 110M BD3-LM model of OpenWebText \citep{arriola2025bd3lm}, four T4-hours of Mix-FT beat continued cross-entropy, Alpha diffusion and SDTT distillation \citep{deschenaux2024sdtt} at 4, 8 and 16 steps per block, all given the same extra compute: two judge models score its samples better than the strongest baseline's at each step count in every seed, by 0.25 to 0.45 nats/token on average at matched per-sample entropy (\cref{tab:main110,fig:time}). Baselines given eight times the budget, more updates or BD3-LM's own objective stay well behind (\cref{sec:scale}).

Exact experiments on text8 explain why this works. We compute the sampler's likelihood exactly with a dynamic program over reveal sets, so we can split the gain. Part of it is the within-step dependence that independent filling loses; the other part appears even at one token per step, where the fine-tuned network is better under the mixture over orders. A control with the same plans, batches and updates but equal plan weights recovers only a quarter of the gain.

\paragraph{Contributions.}
\begin{itemize}
\item A teacher-free fine-tune on the random-order sampler's mixture likelihood that improves sample quality over every baseline we ran at 4, 8 and 16 steps per block, on public 110M, 1.1B and 8B models. On LLaDA-8B its 16 steps per block are ahead of the original's 32; at 110M, with 256 plans per sequence and 8$\times$ the fine-tune budget, it beats the original model by more than 0.5 nats/token at equal steps, and its 8 steps per block close 95\% of the gap between the original's 8 and 16 steps.
\item The finding that training on coarse two-step plans improves 8- and 16-step sampling more than training on four- or eight-step plans at equal compute, and not because coarse plans allow more updates (\cref{sec:scale}).
\item An account of why CE is not the sampler's optimum, building on \citet{amin2026simple}, tested with exact $K$-step likelihoods and a control that isolates the plan weighting (\cref{sec:theory,sec:text8}).
\end{itemize}

\section{Setup: the $K$-step random-order sampler}
\label{sec:setup}

We use block diffusion \citep{arriola2025bd3lm}: a sequence is split into blocks of $B$ positions, generated left to right, each by a masked diffusion sampler conditioned on the clean previous blocks. All statements below concern one block given its prefix, which we drop from the notation.

A plan $d=(G_1,\dots,G_K)$ is an ordered partition of the $B$ positions into $K$ groups of size $g=B/K$, drawn uniformly. The denoiser $r_\theta(x_i \mid x_S)$ is a distribution over the token at masked position $i$ given the revealed set $S$. At step $k$ the sampler draws every position of $G_k$ independently from $r_\theta(\cdot \mid x_{G_{<k}})$. Under a fixed plan and under the sampler,
\begin{gather}
q_{\theta,d}(x) = \prod_{k=1}^{K} \prod_{i\in G_k} r_\theta\big(x_i \mid x_{G_{<k}}\big), \nonumber\\
q_\theta(x) = \sum_d \mu(d)\, q_{\theta,d}(x).
\label{eq:sampler}
\end{gather}
A temperature $\beta$ replaces $r_\theta$ by $\mathrm{softmax}(\beta \cdot \text{logits})$; we sweep $\beta$ for every model. Masked CE pretraining draws a mask rate and minimizes the CE of masked positions. Our stronger CE baseline draws a plan and a step $k$, reveals $G_{<k}$ and minimizes the CE of all masked positions, so it trains exactly at the states the sampler visits. Both have the true conditionals $p(x_i\mid x_S)$ as their optimum.

\section{Fine-tuning on the sampler's mixture likelihood}
\label{sec:theory}

\paragraph{Objective.} We fine-tune on the random-order sampler's mixture likelihood, through the importance-weighted (IWAE) bound over $M$ plans drawn from $\mu$:
\begin{equation}
\cL_M(\theta) = \E_{x,\,d_{1:M}}\Big[-\log \frac{1}{M}\sum_{m=1}^{M} q_{\theta,d_m}(x)\Big],
\label{eq:iwae}
\end{equation}
an upper bound on $-\E_p\log q_\theta(x)$ that tightens monotonically in $M$ \citep{burda2016iwae}. Evaluating the bound for a sequence costs $1+M(K-1)$ forward passes: one with the block fully masked, shared by all plans, and $K-1$ per plan. Blocks share passes, since they are conditionally independent given the clean prefix. With self-normalized weights $\tilde w_m \propto q_{\theta,d_m}(x)$ the gradient is
\[
\nabla_\theta \cL_M = -\E\Big[\sum_{m}\tilde w_m \sum_{k}\sum_{i\in G^{(m)}_k}\nabla_\theta\log r_\theta\big(x_i\mid x_{G^{(m)}_{<k}}\big)\Big],
\]
cross-entropy at the sampler's own states with each plan weighted by how well it explains $x$. At $M=1$ the weight is 1, and $\cL_1$ is the per-path loss below, whose optimum is the true conditionals, so the method needs $M>1$. As $M\to\infty$ the weights approach the plan posterior and the gradient approaches that of $-\log q_\theta(x)$. In exact toys, training a free table on $\cL_1$ leaves it where it started, while $M=16$ removes 94\% of the error of the true conditionals at $K=3$ (\cref{tab:iwaeM}, \cref{app:toys}). Training targets one random-order sampler at one step count ($K=2$ or 4); other step counts and confidence decoding are transfers we measure.

\paragraph{Why CE is not the sampler's optimum.}
\begin{proposition}[Fixed or state-determined plans]
\label{prop:fixed}
If the plan is fixed, or the next group is a deterministic function of the revealed tokens that does not depend on $\theta$, then
$-\E_p \log q_{\theta,d}(x) = \sum_k \E_p\big[-\sum_{i\in G_k}\log r_\theta(x_i\mid x_{G_{<k}})\big]$,
and every term is minimized by $r_\theta(x_i\mid x_S)=p(x_i\mid x_S)$.
\end{proposition}
This is the chain rule. For such samplers nothing done to the denoiser helps beyond learning the conditionals; the remaining error is the plan's factorization error. Planners that read the denoiser's own outputs, such as confidence-based ones, are not covered. For a random plan, let $\pi_\theta(d\mid x)=\mu(d)q_{\theta,d}(x)/q_\theta(x)$ be the posterior over plans.
\begin{proposition}[Per-path loss = mixture loss + plan information]
\label{prop:gap}
For every $x$ (a special case of Theorem 1 of \citet{amin2026simple}, whose mixing term is this KL),
\[
\E_{d\sim\mu}\big[-\log q_{\theta,d}(x)\big] \;=\; -\log q_\theta(x) \;+\; \KL\big(\mu \,\|\, \pi_\theta(\cdot\mid x)\big).
\]
\end{proposition}
\begin{proof}
$\E_\mu \log\big(q_\theta(x)/q_{\theta,d}(x)\big) = \E_\mu \log\big(\mu(d)/\pi_\theta(d\mid x)\big)$.
\end{proof}
The left side, averaged over $p$, is the per-path loss: CE at the sampler's states up to a per-position weighting. It charges the sampler for any information a sample carries about its own plan. No reweighting of masks, steps or orders removes this term, since every such loss keeps the logarithm inside the plan average.

\paragraph{Two effects.} \Cref{prop:gap} holds at every $K$, and it allows two ways in which the mixture optimum differs from what CE learns.
\emph{(i) Within-step dependence.} When a step reveals $g>1$ positions, the mixture can let plans specialize, for example by committing early to a mode so that later same-step tokens are nearly determined. On small exact problems, re-optimizing a free table of conditionals for the mixture removes nearly all of the error that the true conditionals leave, more than a tuned per-step temperature does (\cref{tab:toy}).
\emph{(ii) Fit under the order mixture at one token per step.} At $K=B$ there is no within-step factorization, and exact conditionals make every order exact, so CE would be optimal with unlimited capacity. A finite network cannot make every order agree. The per-path loss, here the any-order negative ELBO (NELBO), asks it to fit every order equally; the mixture likelihood needs only the average over orders to be good, and the weights move effort toward the orders that explain each sequence. Two observations fit this: on text8 the exact likelihood at one token per step improves, by an amount similar to effect (i), and at 110M the tighter IWAE-64 bound of the one-token-per-step sampler improves while the NELBO, the per-path bound, gets worse (\cref{sec:text8,sec:scale}).

\paragraph{Exact evaluation.} Because $r_\theta$ depends on the revealed set and not on the order it was revealed in, the sum over plans in \cref{eq:sampler} collapses to a dynamic program over reveal sets (\cref{app:dp}). At $B=8$, $K=4$ it needs 127 forward passes per sequence for 2,520 plans. We use it for all block-8 text8 results, since IWAE-64 is looser for Mix-FT than for CE and so understates the gain.

\section{Main result: a public 110M model}
\label{sec:scale}

\paragraph{Setup.} We fine-tune the public BD3-LM OpenWebText checkpoint with block 16 \citep{arriola2025bd3lm}, a 110M-parameter model trained on sequences of 1,024 GPT-2 tokens, and study its natural sampler. Every arm starts from these weights and gets the same extra compute, counted in forward-equivalents of the network (\cref{app:hparams}). This budget is under 0.02\% of the pretraining compute and under 4 T4-hours per arm. Mix-FT trains on two-step plans ($K=2$) with $M=8$ plans per sequence. The baselines are continued CE at the four-step sampler's states; Alpha diffusion \citep{gushchin2026alpha}, trained exactly like continued CE with only the loss changed, at its best setting $\kappa=0.5$; and SDTT \citep{deschenaux2024sdtt} distilled from 16 to 4 steps per block in two rounds (SDTT-4) and from 16 to 8 in one (SDTT-8). Each arm has three seeds (\cref{app:hparams}).

\paragraph{Evaluation.} For each step count $K\in\{2,4,8,16\}$ per block we draw 512 samples of 1,024 tokens at each of six temperatures. GPT-2-large and Pythia-1.4B \citep{radford2019gpt2,biderman2023pythia} score them on content tokens, leaving out newline and end-of-text targets, so that a model's newline rate, which few-step sampling inflates and which differs between methods, does not decide a comparison (\cref{tab:newline}). Scored on all tokens instead, Mix-FT still beats the start and every baseline at $K=4$, 8 and 16 in every seed under both judges (\cref{app:110m}). Temperature trades quality for diversity, so we compare models at equal per-sample entropy, the entropy of each sample's own token frequencies averaged over samples. We interpolate judge NLL along each model's temperature sweep to four entropy levels, from real text's value to 0.3 nats below it, and average them. This compares whole temperature sweeps, which covers the temperature-tuned baseline of \citet{amin2026simple}. Unless stated otherwise, every claim below that one model beats another holds in every seed under both judges, and controls run in one seed use seed 0. We wrote down the criterion for each comparison before running it; \cref{app:rules} gives each with its outcome.

\paragraph{The first token.} BD3-LM never trains position 0 of a window, and its sampler fixes that token to end-of-text; our fine-tunes train it, and our sampler draws it for every model. Drawing it costs the start 0.05 to 0.11 nats/token at 8 and 16 steps, since that output never received a gradient, and saves it 0.04 to 0.07 at 4 steps; it moves the fine-tuned models by at most 0.07. Every table draws it, and every comparison with the start is stated under whichever convention favors the start, with both numbers where they differ (\cref{app:conv}).

\begin{table*}[t]
\centering
\caption{Public BD3-LM OpenWebText model (110M, block 16), all arms from the same weights at the same extra compute (about 0.02\% of pretraining). Left: IWAE-32, the bound of \cref{eq:iwae} with 32 evaluation plans, for the $K$-step sampler (nats/token, best of three temperatures, 256 held-out sequences), and NELBO perplexity, which bounds the one-token-per-step sampler. Right: GPT-2-large NLL of the samples on content tokens at matched per-sample entropy (nats/token, lower is better), with the time per 512 samples of 1,024 tokens on one T4. Means of 3 seeds (start: one model); per seed in \cref{tab:perseed110}, Pythia-1.4B in \cref{tab:pythia}. Bold: best entry. --: not measured. Ablations of Mix-FT in \cref{tab:ablate110}. $^\dagger$The first token of each sequence fixed to end-of-text, as BD3-LM's own sampler does; the other rows draw it (\cref{app:conv}). Real text: 256 held-out windows at their own per-sample entropy, the top matched level.}
\label{tab:main110}
\small
\setlength{\tabcolsep}{3.2pt}
\begin{tabular}{lcccccccc}
\toprule
 & & \multicolumn{2}{c}{sampler bound} & & \multicolumn{4}{c}{judge NLL, content tokens, at steps per block (time)} \\
\cmidrule(lr){3-4}\cmidrule(lr){6-9}
model & opt.\ steps & $K=4$ & $K=2$ & NELBO PPL & $K=2$ (55 s) & $K=4$ (93 s) & $K=8$ (162 s) & $K=16$ (305 s) \\
\midrule
Start (BD3-LM) & -- & 3.223 & 3.775 & 23.00 & 7.31 & 6.10 & 4.60 & 3.71 \\
\quad first token fixed$^\dagger$ & & & & & 7.46 & 6.17 & 4.54 & 3.63 \\
Continued CE & 5,760 & 3.209 & 3.764 & \textbf{22.41} & -- & 6.20 & 4.82 & 3.89 \\
SDTT-4 & 2$\times$1,920 & 3.245 & 3.796 & 24.41 & \textbf{6.68} & 5.83 & 4.75 & 4.02 \\
SDTT-8 & 3,840 & 3.228 & 3.775 & 23.42 & -- & -- & 4.58 & 3.80 \\
Alpha ($\kappa=0.5$) & 5,760 & 3.202 & 3.758 & 23.16 & 7.16 & 5.79 & 4.54 & 3.78 \\
Mix-FT (ours) & 1,152 & \textbf{3.181} & \textbf{3.723} & 24.51 & 7.07 & \textbf{5.54} & \textbf{4.10} & \textbf{3.37} \\
\quad first token fixed$^\dagger$ & & & & & -- & 5.58 & 4.10 & 3.39 \\ %
Real held-out text & & & & & \multicolumn{4}{c}{2.78} \\
\bottomrule
\end{tabular}
\end{table*}

\begin{figure*}[t]
\centering
\includegraphics[width=\textwidth]{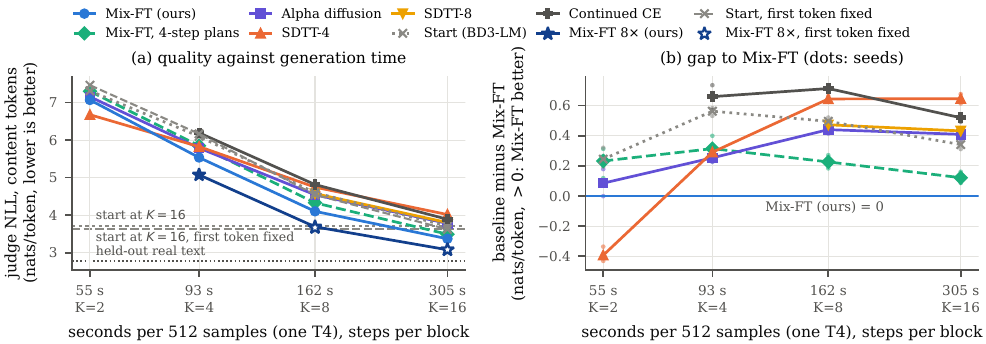}
\caption{Sample quality against measured generation time at 110M (block 16, 512 samples per temperature, one T4). (a) GPT-2-large content-token NLL at matched per-sample entropy, seed means; bottom dotted line: held-out real text. With 8$\times$ the budget, Mix-FT at 8 steps per block comes within 0.04 nats/token of the start's 16-step quality in about half the time when both fix the first token, the start's own convention (hollow star, dashed line), and reaches it when both draw it (filled star, dotted line). (b) Each model's NLL minus Mix-FT's at the same step count (positive: Mix-FT better); lines are seed means, dots are the three paired seeds. From 4 steps per block up every baseline lies above zero in every seed; at 2 steps SDTT-4 lies below.}
\label{fig:time}
\end{figure*}

\paragraph{Mix-FT gives the best judged samples from 4 to 16 steps per block.} At every step count from 4 to 16, Mix-FT's samples score better than every baseline's (\cref{tab:main110,fig:time}; per seed in \cref{tab:perseed110}): by 0.25 to 0.45 nats/token over the strongest fine-tuned baseline at each step count, and by 0.52 to 0.72 over continued CE, the usual way to spend extra compute. All methods take the same time per step, so Mix-FT is the best model we measured at every generation budget from 93 seconds up. At two steps per block it still beats the start, by 0.18 to 0.32 nats/token in every seed under both judges with the first token drawn, the start's better convention there, while SDTT-4, distilled for few steps, leads. It is not merely the model that loses least. Every fine-tuned baseline ends behind the untouched start at $K=16$, continued CE by 0.15 to 0.25 nats/token at $K=8$ and 16 (0.22 to 0.31 in every seed when both fix the first token), and neither a tenth of the learning rate nor EMA weights bring it level (\cref{tab:controls110}e). Even BD3-LM's own objective, continued for the same budget on its own windows, ends 0.18 to 0.43 behind the start, and Mix-FT on those windows beats it by 0.50 to 0.96 (\cref{tab:controls110}g). Mix-FT beats the start at $K=16$ by more than 0.3 nats/token, and by 0.22 to 0.26 in every seed when both fix the first token. Per-step temperature schedules, tuned on a development split, improve Mix-FT, Alpha diffusion and SDTT-8 at $K=8$ by the same 0.3 nats/token, which leaves their order unchanged.

\paragraph{Two-step training plans beat finer ones.} Training on two-step plans beats the same recipe on four-step plans at every step count, by 0.12 to 0.31 nats/token on average (\cref{tab:ablate110}). A two-step plan costs fewer passes, so at equal compute this run takes 1,152 optimizer steps instead of 443, but the extra updates are not the reason (\cref{tab:controls110}a): a two-step run of only 443 steps still beats the four-step recipe, while four-step runs given about as many updates as the two-step run stay 0.25 to 0.48 nats/token behind at $K=4$. Even training on the eight-step plans that the $K=8$ sampler runs is worse at $K=8$ than training on two-step plans. Our untested hypothesis is that coarse plans give the objective the most to correct: the start's plan weights are most concentrated at $K=2$ (median effective sample size of eight plans 1.32, against 1.54 at $K=4$ and 1.95 at $K=8$; \cref{app:110m}), so the plan information that \cref{prop:gap} charges for is largest there. On text8, where the likelihood is exact, two-step plans also beat the four-step recipe at equal compute on the exact $K=4$ likelihood and on samples in every seed, though each step reveals half a block (\cref{tab:t8abl}b). How the two-step model was chosen is in \cref{app:rules}.

\paragraph{Likelihood and its price.} Mix-FT has the best sampler bound of all arms at $K=4$ and $K=2$, below continued CE's in every seed (\cref{tab:main110}). The NELBO, the per-path bound of the one-token-per-step sampler, gets worse: its perplexity rises from 23.0 to 24.5, while continued CE lowers it to 22.4 under our scoring of the held-out windows (\cref{app:conv}). The tighter IWAE-64 bound of the same sampler does not get worse, and ends 0.002 to 0.003 nats/token below continued CE's in every seed. So the NELBO rise widens its gap to the tighter bound, the looser per-path fit that effect (ii) predicts.

\begin{figure*}[t]
\centering
\includegraphics[width=\textwidth]{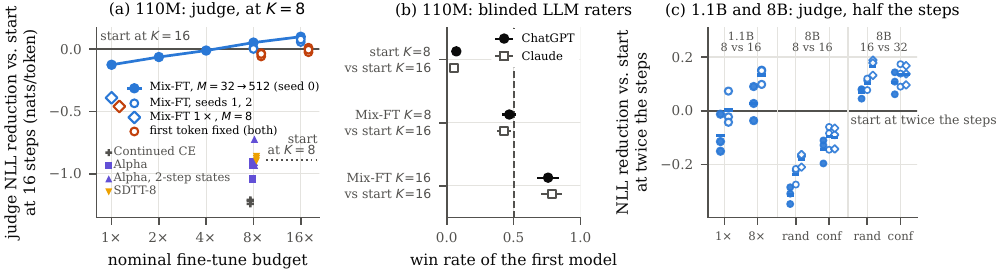}
\caption{Half the steps. (a) 110M: judge NLL of the untouched start at 16 steps per block minus that of each model at 8 (GPT-2-large, content tokens, matched entropy, one temperature per model; $>0$: the 8-step model is better). Doubling plans and nominal budget together at about 350 updates narrows the gap (as executed, 8$\times$ is 0.18\% of pretraining and 16$\times$ 0.25\%); at 8$\times$ the seeds are 0.03 to 0.06 below the start when both fix the first token (orange), the start's own convention, and at or above it when both draw it; 16$\times$ adds 0.03 to 0.08, while every baseline given 8$\times$ stays more than 0.7 below in every seed, near or below the start's own 8 steps. (b) Blinded pairwise raters, win rate of the first model, both models fixing the first token (both orders, ties half, 95\% intervals; \cref{tab:reprorater}). (c) The same gap per seed (markers) and seed mean (bars) under GPT-2-large (filled circles), Pythia-1.4B (open circles) and Qwen2.5-7B (open diamonds): at 1.1B for the 1$\times$ fine-tune and the 8$\times$ recipe of (a); at 8B (block 32, 1$\times$) for random order and LLaDA's confidence planner, at 8 steps against the start's 16 and at 16 against its 32 (\cref{tab:results8b}).}
\label{fig:half}
\end{figure*}

\paragraph{Half the steps.} At equal compute, more plans help little beyond $M=32$ because the update count falls. Holding the update count near 350 and doubling the plans together with the compute narrows the gap between Mix-FT at $K=8$ and the start at $K=16$ one doubling at a time (\cref{fig:half}a). With $M=256$ at 8$\times$ the budget (0.18\% of the pretraining compute as executed, 0.13\% as counted; \cref{app:hparams}), Mix-FT at $K=8$ closes 95\% of the start's own gap between 8 and 16 steps under GPT-2-large (bootstrap interval 92 to 99\%) and 97\% under Pythia, in about half the generation time (\cref{fig:time}a). This is under the convention that favors the start, with the first token fixed for both: there the 8-step model stays 0.04 nats/token behind under GPT-2-large and 0.02 under Pythia, whose interval includes zero, and 0.06 behind under both judges with per-step temperature schedules (\cref{tab:h8}). With the first token drawn, it is level with or ahead of the start in every seed under both judges, with either temperature setting. Its MAUVE at $K=8$ is within 0.02 of the start's at $K=16$ in every setting. At equal steps it is far ahead, by more than 0.5 nats/token at $K=16$ under either convention. At 16$\times$ ($M=512$, 0.25\% of pretraining) every seed gains another 0.03 to 0.08: its 8 steps beat the start's 16 in every seed with the first token drawn, and with it fixed the seed means are $-0.005$ (GPT-2-large) and $+0.016$ (Pythia; \cref{tab:h8}).

We also put the comparison to blinded pairwise raters from two model families, ChatGPT and Claude Sonnet, with the first token fixed (\cref{fig:half}b; protocol in \cref{app:repro}). Both pass a built-in control that an open-weights Qwen2.5-7B rater failed. The start's $K=8$ text wins only 6 to 7\% of its pairs against its own $K=16$ text; the fine-tuned model at $K=8$ wins 45\% with the raters pooled, so it recovers 88\% of the win rate lost by halving the steps (95\% interval 79 to 97\%; 90\% with the first token drawn). Against the start at $K=16$ ChatGPT's reading is inconclusive (0.47) and Claude's is worse (0.43). At equal steps both raters prefer the fine-tuned model's text, in 76 and 79\% of pairs and in every seed, on coherence, readability and substance (88\% with the first token drawn; \cref{tab:reprorater}).

\paragraph{Under a confidence planner the ordering is the same.} LLaDA's default low-confidence remasking \citep{nie2025llada} samples every masked token at each step and keeps the $B/K$ most probable. Mix-FT never trained on it, but at 110M it beats every baseline and the start at $K=4$, 8 and 16 under this planner too, and the 8$\times$ model at $K=8$ beats the start's confidence sampler at $K=16$ in every seed, by 0.02 to 0.07 nats/token with the first token fixed and at least 0.15 with it drawn (\cref{tab:mauve110}).

\paragraph{Compute alone does not close the gap.} More compute without more plans does not help Mix-FT: at 3$\times$ the budget with $M=8$ or 32, it gains under 0.05 nats/token at $K=8$ (\cref{tab:controls110}b). More compute does not help the baselines either. Given the full 8$\times$ budget (three seeds each), continued CE, Alpha diffusion on four-step and on two-step states, and SDTT-8, with teacher passes counted and checkpoints chosen on a development split, all stay more than 0.7 nats/token behind the start at $K=16$, and their 8-step samples get worse or stay level, except Alpha on two-step states in one seed (0.13 to 0.16 better than 1$\times$ Alpha at $K=8$, still 0.72 behind; \cref{fig:half}a, \cref{tab:repro8xbase}). They see 65 to 129 times as many distinct sequences as the 8$\times$ Mix-FT, and their bound barely moves: more data and more updates do not reproduce what the objective gives. On text8, continued CE given the compute of 256 plans, as 769 times the updates, reaches a much lower exact NLL than Mix-FT but worse samples (seed 0; \cref{tab:t8abl}c).

\paragraph{MAUVE and diversity agree with the judges.} MAUVE \citep{pillutla2021mauve} on GPT-2-large features with the boundary tokens removed (\cref{app:110m}) is 0.86 to 0.89 for Mix-FT at $K=8$, against 0.68 for Alpha diffusion and 0.55 for the start (0.86 against 0.62 in seed 0 with the first token fixed; \cref{tab:mauve110}). The 8$\times$ model at $K=8$ is also more diverse across samples than the start at $K=16$ (\cref{tab:div}), with a 6 to 9\% higher Vendi score \citep{friedman2023vendi} on GPT-2-large features and a lower self-BLEU-4 in every seed, with the first token drawn or fixed. At equal steps it repeats more 4-grams per sample than the start (6.3\% against 3.8\%; real text 3.6\%), which the raters did not penalize and the judge NLL barely notices (\cref{app:110m}).

\paragraph{At 8$\times$ the likelihood bound is level with the start's.} With 256 evaluation plans the 8$\times$ model's bound is within about 0.01 nats/token of the start's at $K=8$ and 16, while its NELBO perplexity rises to 28.8 and its plan weights concentrate on an effective 4 to 6 of the 256 plans (\cref{tab:bounds8x}). Its bound still tightens faster than the start's as plans are added, so its exact likelihood is not known. On text8, where it is exact, more plans at fixed updates lower the sampler's NLL in every seed up to $M=128$ and then flatten, while the gap between the eight-plan bound and the exact NLL widens by 0.03 bpc (\cref{tab:t8abl}c): a loose bound can hide a real gain.

\section{At scale: LLaDA-8B}
\label{sec:8b}

\begin{table}[t]
\caption{LLaDA-8B fine-tuned with LoRA for 0.0018\% of its pretraining compute (block 32, 192-token continuations of a 64-token prompt). Top: GPT-2-large NLL at matched per-sample entropy (nats/token, 3-seed mean, lower is better); at $K=32$ the start reveals one token per step. Middle: the start at $K=32$ minus Mix-FT at $K=16$, seed mean and 95\% interval (positive: Mix-FT better). Bottom: share of 288 blinded pairwise judgments preferring Mix-FT at $K=16$ to the start at $K=32$, low-confidence decoding only (ties half, 95\% interval; \cref{app:8b}). Per seed and $K=4$: \cref{tab:results8b,tab:perseed8b}.}
\label{tab:8b}
\small
\setlength{\tabcolsep}{2pt}
\begin{tabular}{lcccccc}
\toprule
 & \multicolumn{3}{c}{random order} & \multicolumn{3}{c}{low confidence} \\
\cmidrule(lr){2-4}\cmidrule(lr){5-7}
model & $K=8$ & 16 & 32 & $K=8$ & 16 & 32 \\
\midrule
Start & 5.01 & 3.98 & 3.54 & 5.72 & 5.02 & 4.48 \\
Continued CE & 4.88 & 3.89 & -- & 5.69 & 4.83 & -- \\
Alpha ($\kappa=0.5$) & 4.67 & 3.74 & -- & 5.55 & 4.76 & -- \\
Mix-FT (ours) & \textbf{4.29} & \textbf{3.48} & -- & \textbf{5.17} & \textbf{4.38} & -- \\
\midrule
\multicolumn{7}{l}{\emph{start at $K=32$ minus Mix-FT at $K=16$}} \\
GPT-2-large & \multicolumn{3}{c}{$+0.07$ $[-0.02, +0.15]$} & \multicolumn{3}{c}{$+0.10$ $[+0.01, +0.21]$} \\
Pythia-1.4B & \multicolumn{3}{c}{$+0.10$ $[+0.02, +0.19]$} & \multicolumn{3}{c}{$+0.13$ $[+0.04, +0.23]$} \\
Qwen2.5-7B & \multicolumn{3}{c}{$+0.17$ $[+0.09, +0.25]$} & \multicolumn{3}{c}{$+0.13$ $[+0.04, +0.22]$} \\
\midrule
ChatGPT rater & \multicolumn{3}{c}{--} & \multicolumn{3}{c}{69\% $[63, 75]$} \\ %
Claude rater & \multicolumn{3}{c}{--} & \multicolumn{3}{c}{71\% $[66, 77]$} \\ %
\bottomrule
\end{tabular}
\end{table}

We fine-tuned LLaDA-8B-Base \citep{nie2025llada} with LoRA on OpenWebText for prefix continuation, with three seeds per arm, and sampled it with its own sampler: after the prompt, six blocks of 32 tokens from left to right, with LLaDA's default low-confidence remasking or with random order (\cref{app:8b}). Mix-FT uses our 1$\times$ recipe ($M=8$, four-step plans) and beats continued CE and Alpha diffusion at the same compute at every step count, in every seed under three judges and both planners, by 0.22 to 0.60 and 0.18 to 0.40 nats/token, though their IWAE-8 bounds are within 0.011. At equal steps it beats the start by 0.50 to 0.85 at $K=8$ and 16 (\cref{tab:8b}).

\paragraph{Half the model evaluations.} At $K=16$, two tokens per step as in our 110M comparison, Mix-FT is ahead of LLaDA at $K=32$, one token per step, in every seed under the three judges with either planner, so it writes better continuations with 96 instead of 192 forward passes (\cref{tab:8b}, \cref{fig:half}c). Under LLaDA's own confidence decoding every judge's interval is above zero, MAUVE rises from 0.63 to 0.73--0.77 (0.85 to 0.95 under random order), and blinded pairwise raters from two model families, ChatGPT and Claude Sonnet, prefer Mix-FT's continuations in 69 and 71\% of 288 pairs, in every seed (\cref{tab:8b}); both pass a built-in control (\cref{app:8b}). At $K=8$, four tokens per step, Mix-FT closes 66 to 91\% of the start's own gap between 8 and 16 steps in every seed, and a run with 32 plans (seed 0) 78 to 99\%, but stays 0.10 to 0.31 behind, where continued CE and Alpha diffusion stay 0.49 to 0.91 behind. At 110M and 1.1B the 8-step result needed 8$\times$ the budget, untried at 8B.

\paragraph{1.1B.} On the 1.1B SMDM base model \citep{nie2025smdm} (LoRA, block 16, under 0.01\% of pretraining), Mix-FT also wins at every step count, and with the 110M half-step recipe at 8$\times$ its $K=8$ is ahead of the start's $K=16$ on average under both judges (\cref{tab:results11b}).

\section{Why it works: exact experiments on text8}
\label{sec:text8}

\paragraph{Protocol.} We use text8 \citep{mahoney2011text8} in sequences of 256 characters, a 4.82M-parameter block diffusion transformer, block 8 and $K=4$ steps (two characters per step) unless stated. Compute is counted in sequence passes; one unit is 691,200 passes. The base model is masked-CE pretraining followed by four units of CE at the sampler's states (``Base 4$\times$''). Every fine-tune starts from Base 4$\times$ and gets one more unit, with three seeds. We report the exact $K$-step NLL in bits per character (bpc) on 2,000 held-out sequences at the best of ten temperatures, and $G$, the gap in a character-level judge's NLL to continued CE at matched distinct-4, the share of distinct 4-grams among all 4-grams of the samples, along a sweep of ten temperatures (positive: better than continued CE).

\begin{table}[t]
\centering
\caption{Text8, block 8, $K=4$: exact $K$-step NLL (bpc, best temperature, 3-seed mean) and differences against continued CE from the same Base 4$\times$ weights at equal extra passes (negative $\Delta$, positive $G$: better). $\Delta_8$: exact NLL at $K=8$ (one token per step) against CE trained at the $K=8$ states. $\Delta_{16}$: block 16, $K=4$, IWAE-64. Hybrid: Mix-FT plus the CE loss on the same forward passes. Per seed in \cref{app:text8}.}
\label{tab:text8}
\small
\setlength{\tabcolsep}{2.6pt}
\begin{tabular}{lccccc}
\toprule
arm & NLL & $\Delta$ & $G$ & $\Delta_8$ & $\Delta_{16}$ \\
\midrule
Base 4$\times$ (start) & 1.804 & & & & \\
Continued CE & 1.792 & 0 & 0 & 0 & 0 \\
SDTT & 1.802 & $+0.009$ & $-0.038$ & -- & $+0.009$ \\
Per-path FT & 1.782 & $-0.010$ & $+0.014$ & -- & -- \\
Hybrid & 1.764 & $-0.028$ & $+0.076$ & -- & $-0.031$ \\
Mix-FT & \textbf{1.751} & $\mathbf{-0.042}$ & $\mathbf{+0.156}$ & $\mathbf{-0.025}$ & $\mathbf{-0.039}$ \\
\bottomrule
\end{tabular}
\end{table}

\paragraph{A second stage, not a replacement.} Used in place of CE from the start of training, the mixture objective is worse on the sampler's likelihood, by about 0.06\bpc{} after four units (\cref{fig:scratch}), with 25$\times$ fewer optimizer steps and a gradient that grows noisier once the weights move (\cref{app:text8}). CE learns the conditionals cheaply, and the mixture objective then corrects them.

\paragraph{From a CE start it wins.} From Base 4$\times$, Mix-FT lowers the exact NLL by 0.039 to 0.043\bpc{} relative to continued CE in every seed (\cref{tab:text8,fig:t8lik}), about three times what continued CE gains from its own fifth unit. Its samples are better at matched diversity in every seed, with equal or slightly higher coverage of real 4-grams (\cref{fig:t8frontier}). At block 16, where we can only bound the likelihood, the gain on IWAE-64 and IWAE-512 is of the same size, and SDTT from the same weights barely moves the bound and gives worse samples than continued CE (\cref{app:text8}).

\paragraph{The plan weighting is the main cause.} Per-path FT keeps everything else in which Mix-FT differs from continued CE (the same sequences, 8 plans and 25 states per sequence, optimizer and 25$\times$ fewer steps) and only replaces $-\log$ of the plan average by the average of $-\log$, giving every plan weight $1/8$. It is better than continued CE but worse than Mix-FT in every seed, so the weighting accounts for about three quarters of the likelihood gain (73 to 80\%; \cref{tab:control}). On samples, Mix-FT beats per-path FT in every seed, while per-path FT's gain over continued CE is small and inconsistent; at 110M the same control ends behind continued CE on the bound and 0.53 to 0.62 nats/token behind Mix-FT on samples (\cref{tab:controls110}c). The weighting also lets plans specialize, which is the plan information that \cref{prop:gap} charges for: the effective sample size of eight plan weights falls from 4.3 under continued CE to 3.6 (\cref{tab:control}).

\paragraph{Splitting the gain.} At $K=8$ each step reveals one character, so effect (i) is absent. Mix-FT still beats CE trained at the $K=8$ states in every seed, by 0.022 to 0.028\bpc{} on the exact likelihood, and on samples. The model fine-tuned at $K=4$ is as good at $K=8$ as the one fine-tuned there, so its $K=4$ gain of 0.042\bpc{} splits into two parts of similar size: 0.025 that persists at one token per step, effect (ii), and 0.016 by which the gap between $K=4$ and $K=8$ narrows, effect (i) (\cref{app:text8}). Widening the network from 4.8M to 43M parameters shrinks the first part by 0.010 to 0.011\bpc{} in every seed and leaves the second level, as a capacity account of effect (ii) predicts (\cref{tab:t8abl}a).

\section{Related work}
\label{sec:related}

\paragraph{Masked, block and any-order models.} Masked diffusion models \citep{sahoo2024mdlm,shi2024md4,lou2024sedd,nie2025llada} and block diffusion \citep{arriola2025bd3lm} train with CE variants optimized by the true conditionals. Any-order autoregressive models \citep{hoogeboom2022ardm,shih2022aoarm} are the one-token-per-step case.

\paragraph{Orders as latent variables.} \citet{amin2026simple} analyze the few-step sampler as a mixture over paths; their Theorem 1 contains \cref{prop:gap}. They recover much of the gap at sampling time, with temperature 0.9 and float32 instead of float64 categorical sampling, without retraining; the diagnosis is theirs, and we train the denoiser on the mixture. \citet{fox2026orders} learn a token selector and a variational posterior over unmasking orders that sees the clean sequence, and LO-ARM \citep{wang2025loarm} learns a stochastic order policy with such a posterior for one-token-per-step generation; we learn no order policy and fine-tune the denoiser on a multi-sample bound of a fixed uniform plan distribution. Eso-LMs \citep{sahoo2025esolm} use a multi-order bound for evaluation and train with one order.

\paragraph{Trajectory-aware training.} PUMBA \citep{madeira2026pumba} trains on the few-step policy's own trajectories with a carry state. Its confidence planner and carry state put it outside \cref{prop:fixed} and our dynamic program, though not outside \cref{prop:gap} (\cref{app:related}). T3D \citep{zhang2026t3d} distills the model's own trajectories with a reverse-KL-like loss and MDPO \citep{he2025mdpo} uses policy gradients with intermediate rewards, both under confidence decoding, scoring a sample with its plan. We analyze only the idealized path-space reverse KL with plans independent of $\theta$: like trajectory balance with a fixed backward policy \citep{malkin2023gfnvi,sanokowski2025sdds}, it pushes a parallel sampler toward plan-invariant behavior, in exact toys toward independent filling (\cref{app:toys}).

\paragraph{Added latents, heads and teachers.} Di4C \citep{hayakawa2025di4c} distills into a mixture of products over an auxiliary variable, E-MoE \citep{ivanov2026emoe} mixes over expert routes with a single-route bound, VADD \citep{xie2026vadd} adds a Gaussian latent and uses IWAE only for evaluation, and tensor-train heads \citep{kim2026tensortrain} model the joint of masked tokens. Our mixture latent is the plan the sampler already draws. SDTT \citep{deschenaux2024sdtt} matches a few-step student to its many-step self. Alpha diffusion \citep{gushchin2026alpha} fine-tunes teacher-free with a sequence-level $\alpha$-divergence at one masking state; it changes the divergence, not the plan marginal.

\paragraph{Theory of parallel samplers.} With exact conditionals, \citet{wu2026klmixing} separate schedule-averaged from mixture KL, \citet{li2026randomization} show that a uniform mixture of fixed schedules can beat any deterministic adaptive policy, and \citet{chen2025schedules} analyze random-partition schedules. DUEL \citep{turok2026duel} computes exact likelihoods for deterministic unmasking. Sampling-only methods \citep{benhamu2025ebsampler,luxembourg2025dilated} are complementary.

\section{Limitations and conclusion}
\label{sec:limits}

\begin{itemize}
\item \emph{Scale.} Our full set of baselines was run at 110M. At 1.1B and 8B we compared with continued CE and Alpha diffusion only, with LoRA and 256-token sequences. At 8B, Mix-FT at $K=8$ stays behind LLaDA at $K=16$. On a 1.1B GSM8K model fine-tuned further on its own verified solutions, Mix-FT did not differ from continued CE, as the fixed point of the mixture refit predicts for self-generated data (\cref{app:gsm8k}).
\item \emph{Two steps per block.} At $K=2$, SDTT-4's samples beat Mix-FT's under both judges in every seed, and SDTT also ranks ahead at matched distinct-4, partly by repairing a newline excess, more than twice real text's rate, that Mix-FT keeps (\cref{tab:perseed110,tab:newline}). At 110M the 8$\times$ model closes most of the start's gap between 8 and 16 steps, but its 4-step samples stay about 0.5 nats/token behind the start at 8.
\item \emph{Block size.} We tested blocks of 8, 16 and 32, LLaDA-8B's. With one block over 1,024 tokens, as in MDLM, eight plan weights have a median effective sample size of 1.00, against 1.54 for the start at block 16 and $K=4$ (\cref{tab:block32}), so a bound with few plans is loose.
\item \emph{Price.} Mix-FT trades the per-path bound for the sampler's: the NELBO perplexity rises from 23.0 to 24.5 at 1$\times$ and to 28.8 at 8$\times$ (\cref{tab:bounds8x}). It costs $1+M(K-1)$ passes per sequence and needs a CE-trained start: from under-trained weights the mixture arms lose to continued CE (text8 from scratch and a 34M model, \cref{app:text8,app:110m}).
\item \emph{Conventions.} BD3-LM trained on raw windows with the first token clean, we on wrapped ones. Claims against the start use the first-token convention that favors it; the window convention moves BD3-LM's objective by up to 0.38, and Mix-FT leads on either (\cref{tab:controls110}g).
\item \emph{Scope.} We train on uniform random-order samplers, and \cref{prop:fixed} does not cover confidence planners. We did not run PUMBA, Di4C or added-head methods. Judges and raters are models.
\end{itemize}

Few-step masked diffusion samplers are mixtures over reveal plans, and fine-tuning on that mixture's likelihood is a cheap, teacher-free change of loss. On LLaDA-8B it puts 16 steps per block ahead of the original's 32. On a public 110M model it gives the best samples from 4 to 16 steps per block, and with more plans its 8 steps close 95\% of the gap to the original's 16. The two-step regime, where distillation still wins, and full fine-tuning at scale remain open.

\raggedbottom %
\section*{Impact Statement}
This paper presents work whose goal is to advance the field of machine learning, specifically the efficiency of sampling from diffusion language models. Faster and better few-step generation lowers the cost of running such models, with the same potential for misuse as other text generators. We do not see consequences specific to this method that need discussion here.

\bibliographystyle{icml2026}
\bibliography{refs}

\clearpage
\onecolumn
\raggedbottom
\makeatletter\setlength{\@fptop}{0pt}\setlength{\@fpbot}{0pt plus 1fil}\makeatother %
\renewcommand{\topfraction}{0.9}\renewcommand{\bottomfraction}{0.8}\renewcommand{\textfraction}{0.07}\renewcommand{\floatpagefraction}{0.8}
\appendix

\section{The 110M experiments in more detail}
\label{app:110m}

\begin{table}[h]
\centering
\caption{Ablations of Mix-FT at 110M at the compute of \cref{tab:main110} (training plans $K$, number of plans $M$), with the same columns; seed 0 except the four-step recipe (3 seeds). The main recipe ($K=2$, $M=8$) is repeated for reference.}
\label{tab:ablate110}
\small
\setlength{\tabcolsep}{3.2pt}
\begin{tabular}{lcccccccc}
\toprule
 & & \multicolumn{2}{c}{sampler bound} & & \multicolumn{4}{c}{judge NLL at steps per block} \\
\cmidrule(lr){3-4}\cmidrule(lr){6-9}
training plans & opt.\ steps & $K=4$ & $K=2$ & NELBO PPL & $K=2$ & $K=4$ & $K=8$ & $K=16$ \\
\midrule
$K=2$, $M=8$ (main) & 1,152 & 3.181 & 3.723 & 24.51 & 7.07 & 5.54 & 4.10 & 3.37 \\
$K=4$, $M=8$ (3 seeds) & 443 & 3.187 & 3.741 & 23.29 & 7.30 & 5.85 & 4.33 & 3.49 \\
$K=4$, $M=3$ & 1,047 & 3.194 & 3.747 & 22.73 & -- & 6.01 & 4.53 & -- \\
$K=4$, $M=2$ & 1,440 & 3.197 & 3.751 & 22.54 & -- & 6.09 & 4.60 & -- \\
$K=2$, $M=4$ & 1,920 & 3.181 & 3.731 & 23.55 & -- & 5.67 & 4.27 & -- \\
\bottomrule
\end{tabular}
\end{table}

\begin{table}[h]
\centering
\caption{As the right half of \cref{tab:main110}, scored by Pythia-1.4B on content tokens (nats/token, lower is better).}
\label{tab:pythia}
\small
\begin{tabular}{lcccc}
\toprule
model & $K=2$ & $K=4$ & $K=8$ & $K=16$ \\
\midrule
Start (BD3-LM) & 7.18 & 6.08 & 4.59 & 3.69 \\
Continued CE & -- & 6.14 & 4.79 & 3.86 \\
SDTT-4 & \textbf{6.70} & 5.82 & 4.72 & 3.98 \\
SDTT-8 & -- & -- & 4.55 & 3.77 \\
Alpha diffusion ($\kappa=0.5$) & 6.99 & 5.75 & 4.52 & 3.75 \\
Mix-FT ($K=2$ plans) & 6.96 & \textbf{5.50} & \textbf{4.07} & \textbf{3.33} \\
Real held-out text & \multicolumn{4}{c}{2.65} \\
\midrule
$K=4$ plans, $M=8$ & 7.16 & 5.80 & 4.30 & 3.45 \\
$K=4$ plans, $M=3$ & -- & 5.96 & 4.50 & -- \\
$K=4$ plans, $M=2$ & -- & 6.04 & 4.57 & -- \\
$K=2$ plans, $M=4$ & -- & 5.63 & 4.24 & -- \\
\bottomrule
\end{tabular}
\end{table}

\begin{table}[h]
\centering
\caption{Per-seed gains of Mix-FT at 110M: each model's content-token judge NLL minus Mix-FT's at matched per-sample entropy (nats/token, 512 samples per temperature; positive: Mix-FT better). Mix-FT is the two-step model of \cref{tab:main110}. Means are over the three seeds and are computed before rounding, as are the margins quoted in \cref{sec:scale}.}
\label{tab:perseed110}
\small
\setlength{\tabcolsep}{4pt}
\begin{tabular}{lcrrrrrrrr}
\toprule
 & & \multicolumn{4}{c}{GPT-2-large} & \multicolumn{4}{c}{Pythia-1.4B} \\
\cmidrule(lr){3-6}\cmidrule(lr){7-10}
Mix-FT against & $K$ & seed 0 & seed 1 & seed 2 & mean & seed 0 & seed 1 & seed 2 & mean \\
\midrule
Start (BD3-LM) & 2 & 0.218 & 0.320 & 0.195 & 0.244 & 0.205 & 0.268 & 0.176 & 0.216 \\
 & 4 & 0.573 & 0.584 & 0.532 & 0.563 & 0.595 & 0.587 & 0.560 & 0.581 \\
 & 8 & 0.521 & 0.497 & 0.469 & 0.496 & 0.544 & 0.518 & 0.492 & 0.518 \\
 & 16 & 0.348 & 0.364 & 0.312 & 0.341 & 0.368 & 0.388 & 0.334 & 0.363 \\
\addlinespace[2pt]
Continued CE & 4 & 0.569 & 0.735 & 0.673 & 0.659 & 0.573 & 0.709 & 0.660 & 0.647 \\
 & 8 & 0.693 & 0.729 & 0.718 & 0.713 & 0.708 & 0.743 & 0.716 & 0.722 \\
 & 16 & 0.535 & 0.531 & 0.492 & 0.519 & 0.545 & 0.538 & 0.490 & 0.525 \\
\addlinespace[2pt]
SDTT-4 & 2 & $-$0.431 & $-$0.336 & $-$0.412 & $-$0.393 & $-$0.296 & $-$0.238 & $-$0.261 & $-$0.265 \\
 & 4 & 0.292 & 0.292 & 0.287 & 0.290 & 0.333 & 0.314 & 0.331 & 0.326 \\
 & 8 & 0.664 & 0.625 & 0.644 & 0.644 & 0.668 & 0.628 & 0.644 & 0.647 \\
 & 16 & 0.647 & 0.676 & 0.614 & 0.646 & 0.649 & 0.686 & 0.623 & 0.652 \\
\addlinespace[2pt]
SDTT-8 & 8 & 0.497 & 0.456 & 0.465 & 0.473 & 0.505 & 0.466 & 0.472 & 0.481 \\
 & 16 & 0.447 & 0.450 & 0.399 & 0.432 & 0.452 & 0.460 & 0.406 & 0.439 \\
\addlinespace[2pt]
Alpha diffusion ($\kappa=0.5$) & 2 & $-$0.001 & 0.178 & 0.078 & 0.085 & $-$0.031 & 0.087 & 0.028 & 0.028 \\
 & 4 & 0.226 & 0.298 & 0.235 & 0.253 & 0.239 & 0.271 & 0.236 & 0.249 \\
 & 8 & 0.437 & 0.414 & 0.470 & 0.440 & 0.447 & 0.417 & 0.473 & 0.446 \\
 & 16 & 0.435 & 0.420 & 0.372 & 0.409 & 0.444 & 0.438 & 0.379 & 0.421 \\
\midrule
Four-step Mix-FT (ablation) & 2 & 0.265 & 0.313 & 0.119 & 0.232 & 0.200 & 0.274 & 0.108 & 0.194 \\
 & 4 & 0.313 & 0.399 & 0.230 & 0.314 & 0.317 & 0.377 & 0.228 & 0.307 \\
 & 8 & 0.224 & 0.271 & 0.183 & 0.226 & 0.223 & 0.274 & 0.180 & 0.225 \\
 & 16 & 0.117 & 0.161 & 0.086 & 0.121 & 0.115 & 0.163 & 0.087 & 0.122 \\
\bottomrule
\end{tabular}
\end{table}

\paragraph{Per-seed gaps.} \Cref{tab:perseed110} gives every gain of Mix-FT by seed and judge. From $K=4$ to 16 the smallest gain over any baseline is 0.226, under GPT-2-large against Alpha diffusion at $K=4$ in seed 0. At $K=2$ SDTT-4 is better in every seed, and Mix-FT is level with or slightly ahead of Alpha diffusion.

\paragraph{Controls.} \Cref{tab:controls110} collects the 110M controls that \cref{sec:scale} summarizes. Removing the plan weights costs Mix-FT more than any other change we tried, and two-step training plans beat finer plans whether the compute or the number of updates is held fixed. More plans per sequence matter less: at equal compute the gain levels off beyond $M=32$, while doubling the plans together with the compute adds about 0.06 nats/token at $K=8$ per doubling (seed 0).

\begin{table*}[p]
\centering
\caption{Controls at 110M. Each row gives the content-token judge gain of the first arm over the second at matched per-sample entropy (nats/token; positive: first arm better) under GPT-2-large and Pythia-1.4B, with each arm's compute in multiples of the 1$\times$ budget and its number of optimizer updates. Mix-FT runs are named by the plans they train on (two-, four- or eight-step) and by $M$, the number of plans per sequence ($M=8$ unless stated). Rows with three seeds give seed means whose sign holds in every seed under both judges, except cells marked $\dagger$; rows with one seed are seed 0. --: not measured.}
\label{tab:controls110}
\footnotesize
\setlength{\tabcolsep}{2.8pt}
\begin{tabular}{llcccrrrrrr}
\toprule
 & & & & & \multicolumn{2}{c}{$K=4$} & \multicolumn{2}{c}{$K=8$} & \multicolumn{2}{c}{$K=16$} \\
\cmidrule(lr){6-7}\cmidrule(lr){8-9}\cmidrule(lr){10-11}
first arm & second arm & compute & updates & seeds & GPT-2 & Pythia & GPT-2 & Pythia & GPT-2 & Pythia \\
\midrule
\multicolumn{11}{l}{\emph{(a) plan coarseness and update count}} \\
two-step & four-step & 1 / 1 & 1,152 / 443 & 3 & $+0.31$ & $+0.31$ & $+0.23$ & $+0.23$ & $+0.12$ & $+0.12$ \\ %
two-step, 443 steps & four-step & 0.38 / 1 & 443 / 443 & 3 & $+0.23$ & $+0.23$ & $+0.22$ & $+0.22$ & -- & -- \\ %
two-step, 443 steps & two-step & 0.38 / 1 & 443 / 1,152 & 3 & $-0.09$ & $-0.08$ & $-0.00^{\dagger}$ & $-0.01^{\dagger}$ & -- & -- \\ %
two-step & four-step, 1,152 steps & 1 / 2.6 & 1,152 / 1,152 & 1 & $+0.25$ & $+0.26$ & $+0.20$ & $+0.20$ & -- & -- \\ %
two-step & four-step, $M=3$ & 1 / 1 & 1,152 / 1,047 & 1 & $+0.48$ & $+0.48$ & $+0.45$ & $+0.46$ & -- & -- \\ %
two-step & eight-step & 1 / 1 & 1,152 / 198 & 1 & $+0.44$ & $+0.42$ & $+0.26$ & $+0.26$ & $+0.08$ & $+0.08$ \\ %
\midrule
\multicolumn{11}{l}{\emph{(b) number of plans and compute (two-step plans)}} \\
$M=8$ & $M=4$ & 1 / 1 & 1,152 / 1,920 & 1 & $+0.14$ & $+0.15$ & $+0.20$ & $+0.20$ & -- & -- \\ %
$M=16$ & $M=8$ & 1 / 1 & 640 / 1,152 & 1 & $+0.09$ & $+0.10$ & $+0.11$ & $+0.11$ & $+0.09$ & $+0.09$ \\ %
$M=32$ & $M=8$ & 1 / 1 & 338 / 1,152 & 3 & $+0.18$ & $+0.19$ & $+0.24$ & $+0.24$ & $+0.16$ & $+0.16$ \\ %
$M=64$ & $M=32$ & 1 / 1 & 174 / 338 & 1 & $+0.02$ & $+0.02$ & $+0.03$ & $+0.04$ & $+0.01$ & $+0.01$ \\ %
$M=128$ & $M=64$ & 1 / 1 & 88 / 174 & 1 & $-0.04$ & $-0.02$ & $-0.01$ & $-0.02$ & $-0.01$ & $-0.02$ \\ %
$M=8$ at 3$\times$ & $M=8$ at 1$\times$ & 3 / 1 & 3,456 / 1,152 & 1 & $-0.01$ & $-0.01$ & $+0.04$ & $+0.04$ & $-0.00$ & $+0.01$ \\ %
$M=32$ at 3$\times$ & $M=32$ at 1$\times$ & 3 / 1 & 1,016 / 338 & 1 & $+0.06$ & $+0.05$ & $-0.01$ & $+0.00$ & $-0.03$ & $-0.03$ \\ %
$M=64$ at 2$\times$ & $M=32$ at 1$\times$ & 2 / 1 & 349 / 338 & 1 & $+0.11$ & $+0.11$ & $+0.06$ & $+0.07$ & $+0.02$ & $+0.03$ \\ %
$M=128$ at 4$\times$ & $M=64$ at 2$\times$ & 4 / 2 & 354 / 349 & 1 & $+0.08$ & $+0.08$ & $+0.05$ & $+0.05$ & $+0.04$ & $+0.04$ \\ %
$M=256$ at 8$\times$ & $M=128$ at 4$\times$ & 8 / 4 & 357 / 354 & 1 & $+0.07$ & $+0.08$ & $+0.06$ & $+0.06$ & $+0.05$ & $+0.05$ \\ %
\midrule
\multicolumn{11}{l}{\emph{(c) per-path controls: the same runs with equal plan weights}} \\
four-step Mix-FT & four-step per-path FT & 1 / 1 & 443 / 443 & 3 & $+0.53$ & $+0.53$ & $+0.61$ & $+0.62$ & -- & -- \\ %
four-step per-path FT & continued CE & 1 / 1 & 443 / 5,760 & 3 & $-0.18$ & $-0.19$ & $-0.12$ & $-0.12$ & -- & -- \\ %
two-step Mix-FT & two-step per-path FT & 1 / 1 & 1,152 / 1,152 & 1 & $+0.80$ & $+0.79$ & $+0.78$ & $+0.79$ & $+0.57$ & $+0.59$ \\ %
two-step per-path FT & continued CE & 1 / 1 & 1,152 / 5,760 & 1 & $-0.23$ & $-0.21$ & $-0.08$ & $-0.08$ & $-0.04$ & $-0.04$ \\ %
\midrule
\multicolumn{11}{l}{\emph{(d) per-step temperature schedule against one temperature at $K=8$ (same weights)}} \\
Mix-FT, schedule & Mix-FT, one temp. & 1 / 1 & 1,152 / 1,152 & 1 & -- & -- & $+0.30$ & $+0.31$ & -- & -- \\ %
Alpha, schedule & Alpha, one temp. & 1 / 1 & 5,760 / 5,760 & 1 & -- & -- & $+0.29$ & $+0.30$ & -- & -- \\ %
SDTT-8, schedule & SDTT-8, one temp. & 1 / 1 & 3,840 / 3,840 & 1 & -- & -- & $+0.29$ & $+0.30$ & -- & -- \\ %
\midrule
\multicolumn{11}{l}{\emph{(e) learning rate and weight averaging for continued CE}} \\
Mix-FT & continued CE, 0.1$\times$ lr & 1 / 1 & 1,152 / 5,760 & 1 & $+0.76$ & $+0.76$ & $+0.80$ & $+0.81$ & $+0.51$ & $+0.52$ \\ %
Mix-FT & continued CE, EMA & 1 / 1 & 1,152 / 5,760 & 1 & $+0.63$ & $+0.62$ & $+0.68$ & $+0.69$ & $+0.51$ & $+0.52$ \\ %
continued CE, 0.1$\times$ lr & untouched start & 1 / 0 & 5,760 / 0 & 1 & $-0.19$ & $-0.16$ & $-0.28$ & $-0.26$ & $-0.16$ & $-0.15$ \\ %
continued CE, EMA & untouched start & 1 / 0 & 5,760 / 0 & 1 & $-0.06$ & $-0.03$ & $-0.16$ & $-0.15$ & $-0.17$ & $-0.15$ \\ %
\midrule
\multicolumn{11}{l}{\emph{(f) baselines at the 8$\times$ budget}} \\
8$\times$ Mix-FT & continued CE 8$\times$ & 8 / 8 & 357 / 46,080 & 3 & $+1.19$ & $+1.21$ & $+1.24$ & $+1.27$ & $+0.87$ & $+0.89$ \\ %
8$\times$ Mix-FT & Alpha 8$\times$ & 8 / 8 & 357 / 46,080 & 3 & $+0.95$ & $+0.97$ & $+0.98$ & $+1.01$ & $+0.81$ & $+0.84$ \\ %
8$\times$ Mix-FT & Alpha 8$\times$, two-step states & 8 / 8 & 357 / $\le$23,040$^{\ddagger}$ & 3 & $+0.78$ & $+0.81$ & $+0.88$ & $+0.90$ & $+0.77$ & $+0.79$ \\ %
8$\times$ Mix-FT & SDTT-8 8$\times$ & 8 / 8 & 357 / $\le$30,720$^{\ddagger}$ & 3 & $+0.85$ & $+0.90$ & $+0.90$ & $+0.92$ & $+0.72$ & $+0.74$ \\ %
\midrule
\multicolumn{11}{l}{\emph{(g) BD3-LM's own objective (null fine-tune) and BD3-LM's raw windows}} \\
null, raw windows & untouched start & 1 / 0 & 5,760 / 0 & 3 & $-0.25$ & $-0.24$ & $-0.35$ & $-0.36$ & $-0.23$ & $-0.23$ \\ %
Mix-FT, raw windows & null, raw windows & 1 / 1 & 1,152 / 5,760 & 3 & $+0.72$ & $+0.72$ & $+0.84$ & $+0.86$ & $+0.53$ & $+0.55$ \\ %
continued CE, raw windows & null, raw windows & 1 / 1 & 5,760 / 5,760 & 3 & $+0.09$ & $+0.08$ & $+0.01^{\dagger}$ & $+0.01^{\dagger}$ & $-0.02^{\dagger}$ & $-0.02^{\dagger}$ \\ %
Mix-FT, raw windows & continued CE, raw windows & 1 / 1 & 1,152 / 5,760 & 3 & $+0.64$ & $+0.65$ & $+0.83$ & $+0.85$ & $+0.55$ & $+0.56$ \\ %
Mix-FT, raw windows & untouched start & 1 / 0 & 1,152 / 0 & 3 & $+0.47$ & $+0.48$ & $+0.49$ & $+0.50$ & $+0.31$ & $+0.32$ \\ %
null, raw windows & null, our windows & 1 / 1 & 5,760 / 5,760 & 3 & $-0.21$ & $-0.22$ & $-0.25$ & $-0.27$ & $-0.16$ & $-0.17$ \\ %
null, our windows & untouched start & 1 / 0 & 5,760 / 0 & 3 & $-0.04^{\dagger}$ & $-0.02^{\dagger}$ & $-0.10$ & $-0.09$ & $-0.07$ & $-0.06$ \\ %
\bottomrule
\end{tabular}
\par\smallskip
\begin{minipage}{\textwidth}\scriptsize
In (a), ``443 steps'' and ``1,152 steps'' are separate runs with their own learning-rate schedule over that many steps, not snapshots of a longer run. In (b), the $M=8$ run at 3$\times$ continues the 1$\times$ run for 2,304 more steps with a fresh warmup and cosine; the $M=32$ run at 3$\times$ is one run of 1,016 steps. In (c), per-path FT recovers none of Mix-FT's gain on the $K=4$ sampler bound: its gain over continued CE is $-0.14$ of Mix-FT's for four-step plans and $-0.07$ for two-step plans. In (d), schedules are chosen on a development split and scored on fresh samples. In (e), one run uses a tenth of the peak learning rate and the other is evaluated with an exponential moving average of its weights (decay 0.999). In (f), $^{\ddagger}$evaluated at the checkpoint chosen on a development split: half of Alpha's 46,080 steps on two-step states in seed 0 and a quarter in seeds 1 and 2, all of SDTT-8's 30,720 in seed 0, a quarter in seed 1 and half in seed 2; SDTT-8 counts its teacher's passes. Rows in (f) pair each seed of the 8$\times$ Mix-FT with the same seed of the baseline. In (g), the null fine-tune continues BD3-LM's own training objective (its loss and clipped masking schedule, the first position kept clean) for the 1$\times$ budget with continued CE's schedule; ``raw windows'' are BD3-LM's training windows, cut from concatenated documents with nothing added, instead of ours. With the first token fixed for both arms, the null is 0.24 to 0.40 behind the start on average and Mix-FT 0.54 to 0.91 ahead of the null. On our windows and with the first token fixed, the null is level with the start at $K=4$ and 0.18 to 0.23 behind it at $K=8$ and 16.
\end{minipage}
\end{table*}

\paragraph{MAUVE without boundary tokens.} The standard MAUVE feature is GPT-2-large's state at the final token. Every real sequence ends with an end-of-text token and no block-sampler sample does, and with that feature every block-sampler model scores about 0.004 regardless of quality. We drop the first and last position of every sequence, real and generated, before featurizing (real against real: 0.97).

\paragraph{Newline rates.} Few-step samplers from the BD3-LM start over-produce newlines, which make up 3.35\% of real held-out text (\cref{tab:newline}). Sharper sampling emits more of them, and every model except Alpha diffusion needs $\beta>1$ to match real text's per-sample entropy, so for those models the rate at matched entropy, where the judges compare models, is higher than at $\beta=1$. Only SDTT stays near or below real text's rate at both operating points.

\paragraph{All tokens.} Scored on every token, newline and end-of-text targets included, every comparison of Mix-FT with the start and the baselines at $K=4$, 8 and 16 keeps its direction in every seed under both judges. Its margin over the strongest fine-tuned baseline becomes 0.24, 0.43 and 0.40 nats/token at $K=4$, 8 and 16 under GPT-2-large (content tokens: 0.25, 0.44 and 0.41), and over continued CE 0.51 to 0.68 (content tokens: 0.52 to 0.72). A judge predicts a newline at well under 1 nat, against about 4 for a content token, so all-token scores favor whichever model emits more newlines; scoring content tokens removes that advantage for every model alike. %

\begin{table}[h]
\centering
\caption{Share of newline tokens in samples at 110M (percent; real held-out text: 3.35\%). Matched entropy: interpolated along each model's temperature sweep at the matched per-sample entropy, as for the judges. $\beta=1$: the untempered sampler. The 512-sample files are the main protocol of \cref{sec:scale}; the 128-sample files are the earlier protocol at $K=2$, 4 and 8. Means over the three fine-tune seeds. --: not measured.}
\label{tab:newline}
\footnotesize
\setlength{\tabcolsep}{3.5pt}
\begin{tabular}{lcccccccccccc}
\toprule
 & \multicolumn{6}{c}{matched entropy} & \multicolumn{6}{c}{$\beta=1$} \\
\cmidrule(lr){2-7}\cmidrule(lr){8-13}
 & \multicolumn{3}{c}{512 samples} & \multicolumn{3}{c}{128 samples} & \multicolumn{3}{c}{512 samples} & \multicolumn{3}{c}{128 samples} \\
\cmidrule(lr){2-4}\cmidrule(lr){5-7}\cmidrule(lr){8-10}\cmidrule(lr){11-13}
model & $K=4$ & $K=8$ & $K=16$ & $K=2$ & $K=4$ & $K=8$ & $K=4$ & $K=8$ & $K=16$ & $K=2$ & $K=4$ & $K=8$ \\
\midrule
Start (BD3-LM) & 8.0 & 5.4 & 4.2 & 10.9 & 8.7 & 5.2 & 4.3 & 3.5 & 3.3 & 6.3 & 4.4 & 3.3 \\
Continued CE & 9.4 & 6.1 & 4.7 & 12.5 & 9.6 & 6.1 & 4.9 & 3.8 & 3.5 & 6.5 & 5.0 & 3.6 \\
Alpha ($\kappa=0.5$) & 7.3 & 5.5 & 4.8 & 11.1 & 7.5 & 5.6 & 7.3 & 5.3 & 4.6 & 12.2 & 7.5 & 5.3 \\
SDTT-4 & 2.4 & 1.8 & 2.2 & 3.6 & 2.6 & 1.8 & 1.5 & 1.4 & 1.9 & 1.7 & 1.6 & 1.3 \\
SDTT-8 & -- & 3.5 & 3.2 & -- & -- & -- & -- & 2.5 & 2.5 & -- & -- & -- \\
Four-step Mix-FT & 8.5 & 5.8 & 4.7 & 11.2 & 8.5 & 5.6 & 5.2 & 3.9 & 3.6 & 7.0 & 5.4 & 3.7 \\
Mix-FT (two-step) & 7.3 & 5.5 & 4.8 & 9.2 & 7.1 & 5.6 & 4.7 & 4.2 & 4.1 & 5.4 & 4.7 & 4.2 \\
Mix-FT 8$\times$ & 7.5 & 5.5 & 4.7 & -- & -- & -- & 5.8 & 4.5 & 4.3 & -- & -- & -- \\
Mix-FT after SDTT-4 & 7.7 & 5.6 & 4.5 & -- & -- & -- & 4.8 & 3.8 & 3.6 & -- & -- & -- \\
\bottomrule
\end{tabular}
\end{table}

\begin{table}[h]
\centering
\caption{Diversity across samples and repetition within them, at matched per-sample entropy (110M, block 16, 512 samples per file; each value is the mean over four entropy levels, from real text's down to 0.3 nats below it). Vendi score on GPT-2-large features (higher: more diverse); self-BLEU-4 (lower: more diverse); distinct-4 over the set of samples; rep-4, the share of 4-grams repeated within a sample. Mix-FT ranges are over three seeds; baselines are seed 0, at 1$\times$ unless marked. Rows marked fixed fix the first token to end-of-text for both models (\cref{app:conv}); the others draw it. No sample of any model here has a near duplicate in its file (8-gram Jaccard similarity above 0.5).}
\label{tab:div}
\small
\begin{tabular}{lccccc}
\toprule
model & $K$ & Vendi & self-BLEU-4 & distinct-4 & rep-4 (\%) \\
\midrule
Real held-out text & -- & 172 & 0.341 & 0.898 & 3.6 \\
Start (BD3-LM) & 8 & 115 & 0.292 & 0.923 & 3.1 \\
Start (BD3-LM) & 16 & 137 & 0.353 & 0.891 & 3.8 \\
Continued CE & 16 & 129 & 0.358 & 0.892 & 3.6 \\
Alpha diffusion & 16 & 133 & 0.363 & 0.869 & 5.3 \\
SDTT-8 & 16 & 138 & 0.339 & 0.916 & 2.5 \\
Continued CE 8$\times$ & 8 & 100 & 0.280 & 0.928 & 3.0 \\
Mix-FT 1$\times$ & 8 & 133--136 & 0.316--0.323 & 0.900--0.902 & 4.1--4.4 \\
Mix-FT 1$\times$ & 16 & 147--148 & 0.365--0.373 & 0.866--0.867 & 5.4--5.6 \\
Mix-FT 8$\times$ & 8 & 146--148 & 0.319--0.334 & 0.891 & 4.7--5.1 \\
Mix-FT 8$\times$ & 16 & 147--152 & 0.363--0.370 & 0.858--0.859 & 6.2--6.4 \\
\midrule
Start, fixed & 8 & 115 & 0.302 & 0.918 & 3.3 \\
Start, fixed & 16 & 136 & 0.363 & 0.887 & 3.9 \\
Mix-FT 8$\times$, fixed & 8 & 144--148 & 0.326--0.336 & 0.890--0.891 & 4.7--5.1 \\
\bottomrule
\end{tabular}
\end{table}

\paragraph{Diversity and repetition.} \Cref{tab:div} gives the measures behind the summary in \cref{sec:scale}. At $K=8$ the 8$\times$ model is more diverse than the start at $K=16$ on Vendi and self-BLEU-4 and level on distinct-4. At equal steps Mix-FT has the higher Vendi score but slightly higher self-BLEU-4 and lower distinct-4, and like Alpha diffusion it repeats more 4-grams within a sample than the start or real text. Repetition does not flatter the judges here. Within the 8$\times$ model's files we regressed each sample's judge NLL on its per-sample entropy (quadratic) and its rep-4: at $K=8$, more repetition at the same entropy goes with a worse NLL in every seed under both judges, and at $K=16$ it changes the NLL by at most 0.006 nats/token per percentage point. The blinded raters, asked about repetition among other criteria, find it level at equal steps (\cref{sec:scale}).

\begin{figure}[h]
\centering
\includegraphics[width=0.8\linewidth]{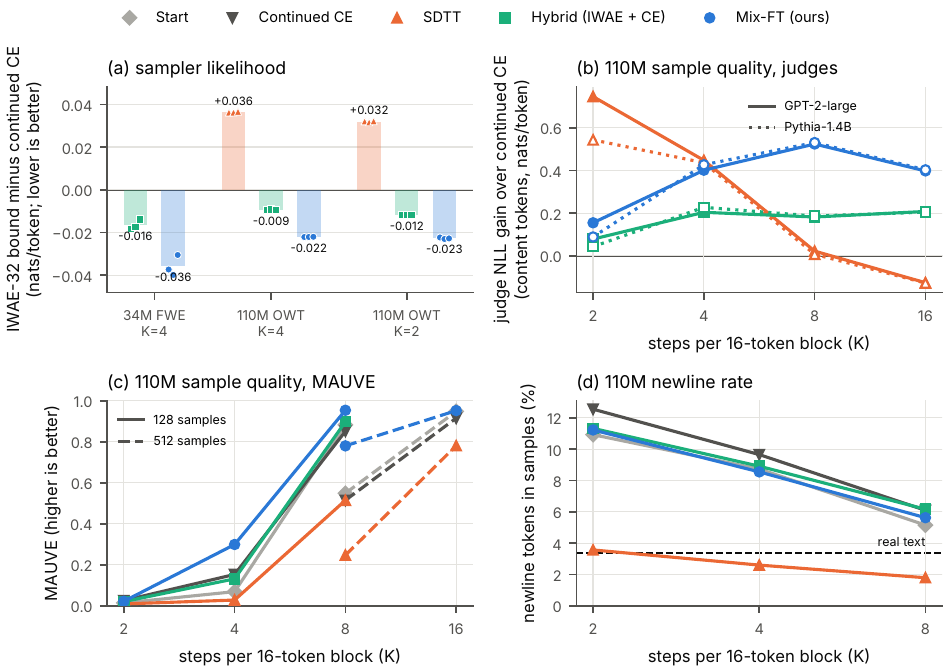}
\caption{Four-step Mix-FT at 110M (and 34M FineWeb-Edu). (a) Sampler likelihood relative to continued CE at equal extra compute (bars: mean of 3 seeds; markers: seeds). (b) Content-token judge gain over continued CE at matched per-sample entropy (solid: GPT-2-large, dotted: Pythia-1.4B; $K=16$ from 512 samples). (c) Boundary-free MAUVE at matched entropy (solid: 128 samples, dashed: 512). (d) Newline rate at matched entropy; real text has 3.35\%.}
\label{fig:scale}
\end{figure}

\paragraph{The four-step Mix-FT.} Our first 110M model used four-step plans (443 steps). Its sampler bound was already better than every baseline's in every seed, including Alpha diffusion at every $\kappa$ (\cref{tab:ablate110,tab:scale}). Its samples beat continued CE on average at every step count, and SDTT-4, SDTT-8 and Alpha diffusion at $K=8$ and 16 in every seed. At fewer steps it was weaker: SDTT-4 was better at $K=2$ and mixed at $K=4$, and Alpha diffusion was ahead in at least two of three seeds at $K=2$ and 4 with either sample count. \Cref{fig:scale,tab:scale} collect these results with the 128-sample protocol; \cref{tab:main110} supersedes them for the two-step model.

\begin{table}[h]
\centering
\caption{Stage 5 to 8 summary at 110M (128 samples per temperature at $K=2,4,8$). $G$: judge NLL gain over continued CE at matched per-sample entropy, mean over $K=2,4,8$ (positive: better than CE); ``all'' scores every token with GPT-2-large, ``content'' excludes newline and end-of-text targets. Bounds: IWAE-32, nats/token; seed spreads at most 0.0013.}
\label{tab:scale}
\small
\begin{tabular}{lcccccc}
\toprule
 & \multicolumn{2}{c}{sampler bound} & & \multicolumn{3}{c}{$G$ vs continued CE} \\
\cmidrule(lr){2-3}\cmidrule(lr){5-7}
model & $K=4$ & $K=2$ & NELBO PPL & GPT-2-L all & GPT-2-L content & Pythia content \\
\midrule
Start (BD3-LM) & 3.223 & 3.775 & 23.00 & & & \\
Continued CE & 3.209 & 3.764 & 22.41 & 0 & 0 & 0 \\
SDTT-4 & 3.245 & 3.796 & 24.41 & $+0.091$ & $+0.405$ & $+0.329$ \\
Hybrid & 3.200 & 3.752 & 22.69 & $+0.113$ & $+0.155$ & $+0.152$ \\
Alpha diffusion & 3.202 & 3.758 & 23.16 & $+0.255$ & $+0.324$ & $+0.321$ \\
Mix-FT, $K=4$ plans & 3.187 & 3.741 & 23.29 & $+0.295$ & $+0.360$ & $+0.349$ \\
Mix-FT, $K=2$ plans & 3.181 & 3.723 & 24.51 & $+0.442$ & $+0.577$ & $+0.563$ \\
\bottomrule
\end{tabular}
\end{table}

\paragraph{Hybrid.} Adding the CE loss on the same forward passes ($\lambda=1$) gives about 40\% of four-step Mix-FT's sample gain over continued CE (\cref{tab:scale}). At $K=8$ and 16 it also beats continued CE and SDTT-4 in every seed under both content judges, by 0.14 to 0.35 nats/token under GPT-2-large.

\paragraph{Mix-FT after SDTT.} Fine-tuning the SDTT-4 student with Mix-FT, against giving it the same compute of further SDTT, lowers its $K=4$ bound by 0.063 nats/token in every seed and improves its samples at $K=8$ and 16 by about 0.5 nats/token. But it lands almost on the model that Mix-FT reaches from the original weights, within 0.003 on the bound and 0.04 on samples on average, and the newline excess that SDTT had removed returns (\cref{tab:newline}). Mix-FT and SDTT are therefore alternatives rather than stages of one recipe. Further SDTT makes the bound and the samples at $K\ge8$ slightly worse.

\paragraph{Alpha diffusion.} It reweights each block's CE by $\exp(\kappa\,\bar s_b)$, where $\bar s_b$ is the model's mean log-likelihood of the block's masked tokens. It is a strong baseline: it beats continued CE at every $K$, most at $K=4$ ($+0.47$ content, 128 samples), but its gain fades with more steps ($+0.10$ to $+0.11$ at $K=16$), where it falls below the untouched start. Larger $\kappa$ degrades the model: its $K=4$ bound is 3.466 at $\kappa=2$ and 5.706 at $\kappa=8$, against 3.202 at $\kappa=0.5$. Smaller values do not help: $\kappa=0.1$ is worse than 0.5 on samples at every $K$, and $\kappa=0.25$ is worse at $K=4$ and 8 and about level at $K=2$ and 16 (seed 0).

\paragraph{A 34M model trained by us.} The recipe needs a CE-trained start. We trained a 34M block diffusion model (8 layers, width 384, 256 GPT-2 tokens, block 16) on 150M tokens of FineWeb-Edu \citep{penedo2024fineweb} with the text8 protocol. From Base 1$\times$, where a further unit of CE still lowered the bound by 0.12 nats/token, both mixture arms ended 0.022 nats/token behind continued CE (one seed). From Base 2$\times$ both beat continued CE in all three seeds, on the bound and on GPT-2-large NLL at matched entropy, and Mix-FT ($M=4$) gains 0.030 to 0.040 nats/token on the bound, about twice the hybrid's gain (\cref{fig:scale}). At matched distinct-4 the gains are near zero on average at this scale.

\paragraph{Bounds of the 8$\times$ model.} \Cref{tab:bounds8x} gives the bounds of the 8$\times$ model and its comparators with up to 256 evaluation plans. With all 256, the 8$\times$ model is level with continued CE at $K=8$ and about 0.03 nats/token behind it at $K=16$, in every seed. Its plan weights are much more concentrated than continued CE's, so its bounds tighten faster as plans are added, and the gap at $K=16$ is still closing at 256 plans. The NELBO, which weights every plan equally, is the loosest bound and the one Mix-FT raises most. The 1$\times$ two-step model is below continued CE and the start at both step counts. Its bound also tightens faster, so its margin over continued CE at $K=16$ is larger with 256 plans than with the 64 plans of \cref{sec:scale}.

\begin{table}[h]
\centering
\caption{Likelihood bounds of the 8$\times$ model at 110M (nats/token, on 256 held-out sequences). IWAE-$M$ bounds the $K$-step random-order sampler with $M$ evaluation plans, at each model's best of $\beta\in\{1,1.1,1.25\}$; IWAE-64 is the mean over four disjoint groups of 64 of the 256 plans. The NELBO is the per-path bound of the one-token-per-step sampler. ESS: median effective sample size of the 256 self-normalized plan weights at $\beta=1$. Tightening: how much more the bound falls than continued CE's when the evaluation plans go from 64 to 256. Mix-FT 1$\times$ is the two-step model; it and both continued CE models are seed 0.}
\label{tab:bounds8x}
\footnotesize
\setlength{\tabcolsep}{4pt}
\begin{tabular}{lccccccccc}
\toprule
 & \multicolumn{2}{c}{IWAE-64} & \multicolumn{2}{c}{IWAE-256} & \multicolumn{2}{c}{ESS of 256 plans} & \multicolumn{2}{c}{tightening} & \\
\cmidrule(lr){2-3}\cmidrule(lr){4-5}\cmidrule(lr){6-7}\cmidrule(lr){8-9}
model & $K=8$ & $K=16$ & $K=8$ & $K=16$ & $K=8$ & $K=16$ & $K=8$ & $K=16$ & NELBO \\
\midrule
Start (BD3-LM) & 3.027 & 2.966 & 3.008 & 2.955 & 10.8 & 18.0 & 0.002 & 0.001 & 3.135 \\
Continued CE & 3.010 & 2.943 & 2.993 & 2.933 & 11.7 & 19.6 & \multicolumn{2}{c}{reference} & 3.109 \\
Continued CE 8$\times$ & 3.001 & 2.934 & 2.984 & 2.924 & 11.8 & 19.8 & 0.000 & 0.000 & 3.100 \\
Mix-FT 1$\times$ & 2.994 & 2.940 & 2.969 & 2.924 & 6.6 & 9.7 & 0.007 & 0.006 & 3.199 \\
\midrule
Mix-FT 8$\times$, seed 0 & 3.034 & 2.991 & 2.995 & 2.963 & 4.4 & 5.9 & 0.022 & 0.018 & 3.361 \\
Mix-FT 8$\times$, seed 1 & 3.032 & 2.989 & 2.994 & 2.962 & 4.5 & 6.0 & 0.021 & 0.017 & 3.356 \\
Mix-FT 8$\times$, seed 2 & 3.035 & 2.992 & 2.996 & 2.964 & 4.5 & 5.9 & 0.022 & 0.018 & 3.362 \\
\bottomrule
\end{tabular}
\end{table}

\paragraph{Confidence sampler.} Each step draws a token for every masked position from $\mathrm{softmax}(\beta\,\text{logits})$, scores it by its untempered probability, and keeps the $B/K$ highest. This is LLaDA's \texttt{low\_confidence} rule, and one step matches LLaDA's reference code in distribution. Temperatures run from $\beta=0.5$ to 1.15 so that every model's sweep brackets real text's per-sample entropy, and each file has 512 samples. Mix-FT has three seeds and the baselines one. \Cref{tab:mauve110} gives MAUVE under both planners and the judge gap between them. On this checkpoint the confidence planner is worse than random order at matched entropy for every model at $K=8$ and 16, by at least 0.25 nats/token, and at $K=4$ for every model except continued CE, which is level. So under this planner the 8$\times$ model at $K=8$ beats the start's confidence sampler at $K=16$ but stays 0.29 to 0.35 nats/token behind the start's random-order sampler, and at $K=4$ it does not reach the start's confidence sampler at $K=8$.

\begin{table}[h]
\centering
\caption{MAUVE at 110M under the random-order sampler and under LLaDA's low-confidence planner, and the judge gap between the two planners. MAUVE uses GPT-2-large features with the boundary tokens removed (\cref{app:110m}), at matched per-sample entropy, with one global temperature per model and 512 samples per file. The gap is the same model's content-token GPT-2-large NLL under the confidence planner minus its NLL under random order, at matched entropy (nats/token; positive: the confidence planner is worse). Pythia-1.4B gives the same signs, except for continued CE at $K=4$, where both judges find the two planners level. Continued CE, Alpha and SDTT-8 are seed 0. Rows marked fixed fix the first token to end-of-text (\cref{app:conv}); the others draw it. --: not measured.}
\label{tab:mauve110}
\footnotesize
\setlength{\tabcolsep}{4pt}
\begin{tabular}{lccccccccc}
\toprule
 & \multicolumn{3}{c}{MAUVE, random order} & \multicolumn{3}{c}{MAUVE, confidence planner} & \multicolumn{3}{c}{judge NLL gap} \\
\cmidrule(lr){2-4}\cmidrule(lr){5-7}\cmidrule(lr){8-10}
model & $K=4$ & $K=8$ & $K=16$ & $K=4$ & $K=8$ & $K=16$ & $K=4$ & $K=8$ & $K=16$ \\
\midrule
Start (BD3-LM) & 0.04 & 0.55 & 0.95 & 0.01 & 0.11 & 0.72 & 0.04 & 0.56 & 0.50 \\
Continued CE & 0.08 & 0.48 & 0.90 & 0.04 & 0.22 & 0.73 & $-0.01$ & 0.45 & 0.42 \\
Alpha ($\kappa=0.5$) & 0.17 & 0.68 & 0.92 & 0.06 & 0.29 & 0.73 & 0.14 & 0.47 & 0.41 \\
SDTT-8 & -- & 0.52 & 0.92 & 0.01 & 0.06 & 0.51 & -- & 0.60 & 0.52 \\
\midrule
Mix-FT 1$\times$, seed 0 & 0.18 & 0.89 & 0.96 & 0.06 & 0.54 & 0.91 & 0.20 & 0.51 & 0.45 \\
Mix-FT 1$\times$, seed 1 & 0.19 & 0.88 & 0.96 & 0.07 & 0.52 & 0.91 & 0.21 & 0.48 & 0.45 \\
Mix-FT 1$\times$, seed 2 & 0.14 & 0.86 & 0.95 & 0.06 & 0.54 & 0.91 & 0.19 & 0.46 & 0.41 \\
Mix-FT 8$\times$, seed 0 & 0.38 & 0.94 & 0.95 & 0.16 & 0.91 & 0.96 & 0.28 & 0.38 & 0.28 \\
Mix-FT 8$\times$, seed 1 & 0.37 & 0.93 & 0.95 & 0.14 & 0.87 & 0.96 & 0.23 & 0.33 & 0.26 \\
Mix-FT 8$\times$, seed 2 & 0.36 & 0.93 & 0.95 & 0.14 & 0.87 & 0.96 & 0.27 & 0.35 & 0.26 \\
\midrule
Start, fixed & 0.04 & 0.62 & 0.94 & 0.02 & 0.17 & 0.77 & -- & -- & -- \\
Continued CE, fixed & 0.09 & 0.54 & 0.91 & -- & -- & -- & -- & -- & -- \\
Mix-FT 1$\times$, seed 0, fixed & 0.18 & 0.86 & 0.96 & 0.07 & 0.52 & 0.91 & -- & -- & -- \\
Mix-FT 8$\times$, fixed & -- & 0.94 & 0.95--0.96 & -- & 0.84--0.88 & 0.96 & -- & -- & -- \\
\bottomrule
\end{tabular}
\end{table}

\paragraph{Block 32 (exploratory).} No BD3-LM checkpoint at block 32 is public, so we ran the block-16 start with block-32 masks, which raises its NELBO perplexity only from 23.00 to 23.30. Before training, the median effective sample size of eight plan weights at $K=4$ was 1.13, below the 1.25 our pilot rule asked for, so we ran the fine-tunes as exploratory follow-ups (\cref{tab:block32}). For the start at block 16 the same median is 1.32, 1.54 and 1.95 at $K=2$, 4 and 8 (1.06, 1.13 and 1.30 at block 32), and for MDLM's single block of 1,024 tokens it is 1.00 at $K=32$ and 128. At the 1$\times$ budget, with 443 Mix-FT updates at $K=4$ and $M=8$, Mix-FT beats continued CE at $K=4$, 8 and 16 under both judges in every seed and has the better bound. In seed 0 it also beats the start at every step count, by 0.14 to 0.30 nats/token with the first token drawn or fixed, while continued CE falls behind the start at $K=8$ and 16 either way. With 8$\times$ the budget and $M=64$ (seed 0), Mix-FT at $K=16$, two tokens per step, matches the start at $K=32$ under both judges when both draw the first token ($+0.02$ and $+0.05$), but falls 0.10 nats/token behind under both when both fix it to end-of-text, the start's better convention here: fixing it helps the start at $K=32$ by 0.11 to 0.13 and leaves the 8$\times$ model where it was (\cref{app:conv}). It beats continued CE given the same budget by at least 0.55 nats/token at $K=4$, 8 and 16.

\begin{table}[h]
\centering
\caption{Block 32 (exploratory). Top: content-token judge gain of the first model over the second at matched per-sample entropy (GPT-2-large / Pythia-1.4B, nats/token; positive: first model better). Mix-FT trains on four-step plans, with $M=8$ at the 1$\times$ budget and $M=64$ at 8$\times$. Rows below the line compare a model at $K$ steps per block with the start at $2K$. Every row draws the first token, except those where both models fix it to end-of-text (\cref{app:conv}). Bottom: IWAE-32 bound of the four-step sampler, NELBO perplexity and plan ESS (ranges over seeds).}
\label{tab:block32}
\small
\begin{tabular}{llccc}
\toprule
comparison & seed & $K=4$ & $K=8$ & $K=16$ \\
\midrule
Mix-FT vs continued CE & 0 & 0.258 / 0.243 & 0.407 / 0.409 & 0.412 / 0.413 \\
 & 1 & 0.310 / 0.290 & 0.455 / 0.475 & 0.430 / 0.436 \\
 & 2 & 0.125 / 0.109 & 0.304 / 0.304 & 0.370 / 0.371 \\
Mix-FT vs start & 0 & 0.265 / 0.264 & 0.267 / 0.296 & 0.255 / 0.278 \\
\quad both fix the first token & 0 & 0.253 / 0.239 & 0.241 / 0.238 & 0.144 / 0.152 \\
continued CE vs start & 0 & 0.006 / 0.021 & $-$0.140 / $-$0.113 & $-$0.157 / $-$0.135 \\
\quad both fix the first token & 0 & 0.020 / 0.004 & $-$0.115 / $-$0.123 & $-$0.252 / $-$0.263 \\
Mix-FT 8$\times$ vs continued CE 8$\times$ & 0 & 0.552 / 0.558 & 0.879 / 0.909 & 0.752 / 0.775 \\
\midrule
Mix-FT at $K$ vs start at $2K$ & 0 & $-$1.048 / $-$0.998 & $-$0.679 / $-$0.650 & $-$0.193 / $-$0.177 \\
Mix-FT 8$\times$ at $K$ vs start at $2K$ & 0 & $-$0.847 / $-$0.786 & $-$0.404 / $-$0.367 & 0.022 / 0.048 \\
\quad both fix the first token & 0 & & & $-$0.097 / $-$0.094 \\
continued CE 8$\times$ at $K$ vs start at $2K$ & 0 & $-$1.399 / $-$1.344 & $-$1.283 / $-$1.275 & $-$0.731 / $-$0.727 \\
\bottomrule
\end{tabular}

\vspace{4pt}
\begin{tabular}{lccc}
\toprule
model & IWAE-32, $K=4$ & NELBO PPL & median ESS of 8 plans, $K=4$ \\
\midrule
Start at block 16 (reference) & 3.223 & 23.00 & 1.54 \\
\midrule
Start (block-32 masks) & 3.387 & 23.30 & 1.13 \\
Continued CE (3 seeds) & 3.364 to 3.365 & 22.66 to 22.67 & 1.14 \\
Continued CE 8$\times$ (seed 0) & 3.354 & 22.44 & 1.14 \\
Mix-FT (3 seeds) & 3.351 & 23.25 to 23.30 & 1.10 \\
Mix-FT 8$\times$ (seed 0) & 3.348 & 24.15 & 1.07 \\
\bottomrule
\end{tabular}
\end{table}

\FloatBarrier
\section{The first token of a sequence}
\label{app:conv}

BD3-LM is trained on 1,024-token windows cut from concatenated documents, each document followed by end-of-text. Position 0 of a window is never masked and never scored, and BD3-LM's sampler sets it to end-of-text before generating. Our fine-tunes follow MDLM's data pipeline instead: every window begins and ends with end-of-text, and position 0 is masked and predicted like any other position. Our sampler draws position 0 for every model. The start therefore draws its first token from an output that never received a gradient, and the rest of the first block, and through the cache every later block, conditions on that token. %

We sampled every model of the main comparison a second time with the first token fixed to end-of-text from the first step, with the same plans and random draws, so that the two files of each pair differ only in that token: the step whose group contains position 0 still draws a token there, which is discarded, and reveals one fewer position. Regenerated with the first token drawn, all 122 files reproduce the files of the main text token for token. \Cref{tab:conv} gives what the fix changes. It helps the start by 0.05 to 0.11 nats/token at $K=8$ and 16 and costs it 0.04 to 0.07 at $K=4$. The fine-tuned models, trained to predict end-of-text at position 0, move by at most 0.07 and mostly by less than 0.03, whether they draw end-of-text there often (continued CE, in 56 to 64\% of samples) or rarely (Mix-FT 1$\times$ in 16 to 21\%, the 8$\times$ model in 1 to 3\%); the start draws it in at most 0.2\%. %

\begin{table}[h]
\centering
\caption{The first token. Top: judge NLL with the first token drawn minus with it fixed to end-of-text, for the same model, $K$ and random draws, at matched per-sample entropy (content tokens, nats/token; positive: fixing helps). Judging the drawn files with end-of-text put in place of their first token, so that only the samples differ, changes these by at most 0.014. Bottom: comparisons with the start, both models drawing or both fixing the first token (positive: first model better). G: GPT-2-large, P: Pythia-1.4B. Per-seed values and intervals for the 8$\times$ model are in \cref{tab:h8}; with the first token fixed, the 1$\times$ model beats the start by 0.54 to 0.62, 0.42 to 0.46 and 0.22 to 0.26 nats/token at $K=4$, 8 and 16 in every seed under both judges.}
\label{tab:conv}
\footnotesize
\setlength{\tabcolsep}{4pt}
\begin{tabular}{lcccccc}
\toprule
 & \multicolumn{2}{c}{$K=4$} & \multicolumn{2}{c}{$K=8$} & \multicolumn{2}{c}{$K=16$} \\
\cmidrule(lr){2-3}\cmidrule(lr){4-5}\cmidrule(lr){6-7}
gain from fixing the first token & G & P & G & P & G & P \\
\midrule
Start (BD3-LM) & $-0.070$ & $-0.035$ & $+0.055$ & $+0.091$ & $+0.087$ & $+0.111$ \\
Continued CE, seed mean & $+0.005$ & $+0.000$ & $-0.013$ & $-0.002$ & $+0.015$ & $+0.019$ \\
Mix-FT 1$\times$, seed mean & $-0.044$ & $-0.021$ & $-0.002$ & $+0.010$ & $-0.016$ & $-0.004$ \\
Mix-FT 8$\times$, seed mean & -- & -- & $+0.022$ & $+0.042$ & $-0.002$ & $+0.016$ \\
\quad with per-step schedules & -- & -- & $-0.003$ & $+0.014$ & -- & -- \\
\bottomrule
\end{tabular}

\vspace{6pt}
\begin{tabular}{lcccc}
\toprule
 & \multicolumn{2}{c}{first token drawn} & \multicolumn{2}{c}{first token fixed} \\
\cmidrule(lr){2-3}\cmidrule(lr){4-5}
comparison & G & P & G & P \\
\midrule
Mix-FT 8$\times$ at $K=8$ vs start at $K=16$, seed mean & $+0.022$ & $+0.047$ & $-0.043$ & $-0.023$ \\
\quad with per-step schedules & $+0.022$ & $+0.039$ & $-0.060$ & $-0.057$ \\
\quad share of the start's gap between $K=8$ and 16 closed & 102\% & 105\% & 95\% & 97\% \\
Mix-FT 8$\times$ at $K=16$ vs start at $K=16$, seed mean & $+0.633$ & $+0.664$ & $+0.545$ & $+0.569$ \\
Mix-FT 1$\times$ vs start, seed mean, $K=4$ & $+0.563$ & $+0.581$ & $+0.588$ & $+0.595$ \\
\quad $K=8$ & $+0.496$ & $+0.518$ & $+0.439$ & $+0.437$ \\
\quad $K=16$ & $+0.341$ & $+0.363$ & $+0.238$ & $+0.248$ \\
Continued CE vs start, seed mean, $K=4$ & $-0.096$ & $-0.067$ & $-0.021$ & $-0.031$ \\
\quad $K=8$ & $-0.218$ & $-0.204$ & $-0.286$ & $-0.297$ \\
\quad $K=16$ & $-0.178$ & $-0.161$ & $-0.250$ & $-0.254$ \\
\bottomrule
\end{tabular}
\end{table}

Every comparison among fine-tuned models in the paper uses one convention for both sides and stands as reported. Comparisons with the start are stated under the convention that favors the start, with both numbers where they differ. At $K=8$ against the start at $K=16$, the 8$\times$ model closes 95\% (GPT-2-large) and 97\% (Pythia) of the start's own gap between 8 and 16 steps with the first token fixed, and all of it with the first token drawn. The 1.1B results are not affected: the SMDM base model was trained to predict position 0, and our 1.1B pipeline follows its convention. %

\paragraph{Bounds.} Our bounds score all 1,024 positions of the held-out windows, including position 0, which the start never learned to predict, and position 1,023, which holds end-of-text only in our windows. BD3-LM's validation scores windows without end-of-text tokens and leaves position 0 out. \Cref{tab:convbound} gives the bound differences to the start under both scorings. Three comparisons change sign, continued CE's IWAE-32 and NELBO and the 8$\times$ model's IWAE-32, and all three are within 0.01 nats/token of the start's under BD3-LM's scoring. The 1$\times$ Mix-FT model's bound stays below the start's under both. %

\begin{table}[h]
\centering
\caption{Bound minus the start's bound (nats/token; negative: better than the start) under our scoring of the held-out windows (Table 1) and under BD3-LM's (no end-of-text inside the windows, position 0 left out).}
\label{tab:convbound}
\footnotesize
\begin{tabular}{lcccc}
\toprule
 & \multicolumn{2}{c}{IWAE-32, $K=4$} & \multicolumn{2}{c}{NELBO} \\
\cmidrule(lr){2-3}\cmidrule(lr){4-5}
model & ours & BD3-LM's & ours & BD3-LM's \\
\midrule
Continued CE, seed 0 & $-0.014$ & $+0.010$ & $-0.026$ & $+0.002$ \\
Mix-FT 1$\times$, seed 0 & $-0.043$ & $-0.024$ & $+0.064$ & $+0.083$ \\
Mix-FT 8$\times$, seeds 0, 1, 2 & $-0.010$ to $-0.009$ & $+0.002$ to $+0.003$ & $+0.221$ to $+0.227$ & $+0.229$ to $+0.234$ \\
\bottomrule
\end{tabular}
\end{table}

\FloatBarrier

\section{LLaDA-8B}
\label{app:8b}

\paragraph{Setting.} \Cref{tab:repro8b} gives the setting. We fine-tune LLaDA-8B-Base \citep{nie2025llada} with LoRA for prefix continuation on OpenWebText and sample it with its own semi-autoregressive sampler. The 256 positions after BOS form eight blocks of 32; the prompt fills the first two, and the model generates the other six from left to right with $K$ steps per block, one full bidirectional forward per step, the later blocks masked and no KV cache. %
We sample with both of LLaDA's planners: random order, the sampler Mix-FT trains on, and LLaDA's default low-confidence remasking, which at each step draws a token for every masked position of the block and keeps the $32/K$ most probable. %
Every arm starts from the same weights and gets the same 480,000 forward-equivalents, 0.0018\% of the pretraining compute, on the same windows. Continued CE and Alpha diffusion train at the four-step random-order sampler's states, as at 1.1B. Mix-FT trains on four-step plans with $M=8$, as in our block-32 runs at 110M (\cref{tab:block32}), because at block 32 a two-step plan reveals 16 tokens per step. %
A second Mix-FT run with $M=32$ (seed 0) tests whether more plans help at equal compute, since the start's plan weights are nearly degenerate at four steps (median ESS of 8 plans 1.19 on the evaluation windows). %
Before any training, our implementation of the network was checked against the Hub's own code and LLaDA's likelihood estimator (\cref{tab:repro8b}). %

\paragraph{Evaluation.} Each model, planner and $K$ gets its own temperature sweep around real text's per-sample entropy, with 256 continuations per file, and all models share the plan and sampling noise. We compare models as at 110M: judge NLL of the continuation's content tokens given the prompt, interpolated along each sweep to four entropy levels from real text's value to 0.3 nats below it and averaged. Three judges are read, GPT-2-large, Pythia-1.4B and Qwen2.5-7B, and a fourth, Qwen2.5-1.5B, is reported. Intervals are a paired bootstrap over prompts with 2,000 replicates, shared by every file and by the real continuations, whose entropy and hence the levels are recomputed in each replicate. %
The start alone is also sampled at $K=32$, one token per step, LLaDA's default full-quality setting and the block-32 analog of our 110M comparison at $K=16$. %
The expectations, written before any run, were: Mix-FT beats continued CE and Alpha diffusion at each planner and $K\in\{4,8,16\}$ under the three judges in every seed; and, separately for each planner, Mix-FT at $K=8$ matches the start at $K=16$ ($H_8\ge0$ under the three judges in every seed), is level with it (seed mean at least $-0.03$ with a 95\% upper bound at least 0 under each judge), or does not reach it; the same rule applies to Mix-FT at $K=16$ against the start at $K=32$. %

\paragraph{Results.} Mix-FT gives the best samples of the three fine-tunes at every planner and $K$, in every seed under the three judges (\cref{tab:results8b}a, \cref{tab:perseed8b}): by 0.22 to 0.60 nats/token over continued CE and 0.18 to 0.40 over Alpha diffusion as seed means, and by at least 0.11 in any seed. %
The bound does not show this. With eight plans at block 32 it stays close to the per-plan bound (median ESS 1.1 to 1.7 for every model), and the three fine-tunes lie within 0.011 nats/token of each other on it, Alpha diffusion lowest, while their samples differ by up to 0.6. %
Mix-FT's MAUVE is also higher than both baselines' at every planner and $K$. %
At equal steps it beats the untouched start by 0.50 to 0.85 at $K=8$ and 16 under either planner and by 0.10 to 0.54 at $K=4$, where continued CE and Alpha diffusion are behind the start. %

With half the steps the result depends on how many tokens each step reveals. At two tokens per step, the setting of our 110M comparison, Mix-FT at $K=16$ is ahead of the start at $K=32$ in every seed under all four judges with either planner, and the seed mean's interval is above zero except under GPT-2-large with random order (\cref{tab:results8b}b); its MAUVE is 0.95 against the start's 0.85 under random order and 0.73 to 0.77 against 0.63 under low confidence. %
At four tokens per step, $K=8$ against the start at $K=16$, it is behind under both planners, but most of the start's own loss from halving the steps is gone: Mix-FT closes 72 to 91\% of it with low confidence and 66 to 84\% with random order, per seed under the three judges, against $-3$ to 47\% for continued CE and Alpha diffusion (seed 0). %
More plans help at equal compute: the $M=32$ run (seed 0) is better than the mean of the $M=8$ runs at every planner and $K$ under GPT-2-large, by 0.04 to 0.12, and closes 78 to 99\% of the start's gap at $K=8$, although its bound is 0.01 worse. %
At 110M and 1.1B the 8-step result needed $M=256$ at eight times the budget (\cref{fig:half}a), which we did not run at 8B.

As at 110M, LLaDA's low-confidence planner is worse than random order at matched entropy for every model and $K$ under GPT-2-large, by 0.2 to 1.0 nats/token, and the ordering of the methods is the same under both. %
Matched entropy per sample does not by itself show diversity across samples (\cref{tab:div8b}). In the comparison of \cref{sec:8b}, Mix-FT at $K=16$ against the start at $K=32$, Mix-FT has the higher Vendi score under both planners (103.2 and 98.8 against 96.9 and 95.3; real text 103.0) and repeats fewer 4-grams within a continuation; its self-BLEU-4 is higher, 0.18 against 0.16, closer to real text's 0.21. At $K=8$ under random order it has a higher Vendi score than the start at $K=16$ (94.4 against 92.8) and than either baseline; under low confidence it is slightly below the start at $K=16$ (80.4 against 82.5) and above both baselines. Its repetition at $K=8$ (2.5\% of 4-grams) is below the start's at $K=16$ and close to real text's 2.8\%, and no model produces near-duplicate continuations. %

\paragraph{Blinded raters.} We put the half-step result at two tokens per step to the same kind of blinded pairwise rater as at 110M (\cref{app:repro}), with the reading rule written down before any of its samples were drawn. Each model was sampled anew with LLaDA's low-confidence decoding at its matched temperature: Mix-FT at $K=16$ (three seeds) and the start at $K=32$. A judgment shows the 64-token prompt and the two 192-token continuations. Each seed contributes 96 pairs, 288 over 191 prompts, every pair rated in both orders in separate files of 40 judgments, with ties counted half and intervals from a bootstrap over prompts. Each model also has 8 control pairs on prompts used by no test pair: its matched-temperature continuation against a hotter one of its own, about 1.35 nats/token worse under GPT-2-large by extrapolation along the temperature sweeps above. A rater whose control interval has its lower bound at or below 0.60 is not read. %
ChatGPT prefers Mix-FT at $K=16$ in 68.8\% of the pairs, 95\% interval 62.9 to 74.5\%, and every seed's interval is above one half (67.4, 70.8 and 68.0\%). It prefers each model's matched-temperature text in 84\% of the control pairs (interval 72 to 95\%), and it agrees with itself across the two orders in 85\% of the test pairs. Mix-FT is preferred on each criterion: coherence 69\%, readability 67\%, substance 69\% and repetition 62\%. %
Claude Sonnet, one fresh instance per file at maximum effort as in our other Claude ratings, prefers Mix-FT at $K=16$ in 71.4\% of the pairs, 95\% interval 65.6 to 77.1\%, and every seed's interval is above one half (72.4, 70.8 and 71.1\%). Its control is 84\% (interval 73 to 95\%), it agrees with itself across the two orders in 85\% of the test pairs, and its verdicts match ChatGPT's in 82\% of the pair orders. It prefers Mix-FT on each criterion as well: coherence 69\%, readability 69\%, substance 67\% and repetition 60\%. %

\begin{table}[h]
\centering
\caption{8B setting (LLaDA-8B-Base, LoRA, OpenWebText prefix continuation).}
\label{tab:repro8b}
\footnotesize
\begin{tabularx}{\linewidth}{@{}l>{\raggedright\arraybackslash}X@{}}
\toprule
start & LLaDA-8B-Base \citep{nie2025llada}, \texttt{GSAI-ML/LLaDA-8B-Base}, revision \texttt{0f2787f2}, six bf16 shards (sha256 checked); 8.0B parameters, pretrained on 2.3T tokens; vocabulary 126,464 \\ %
port & our implementation, checked against the Hub's remote code (transformers 4.46.3) before any training: on 4,104 masked positions its fp32 logits agree within $7\times10^{-4}$ and its masked CE to $10^{-5}$ nats; in fp16, the mode used for training and sampling, its masked CE is within $2\times10^{-4}$ nats of the fp32 reference; LLaDA's own likelihood estimator gives the same mean on 16 prompt--continuation pairs to 0.002 nats (Monte Carlo SE 2.5) \\ %
data & OpenWebText, re-tokenized with LLaDA's tokenizer, documents joined with EOS; 65,536 training windows of 256 tokens from shard 0; evaluation on 256 windows from the last 3,000 documents of shard 79; prompt: BOS and the window's first 64 tokens; the model generates the other 192 \\ %
LoRA & rank 32, alpha 32, fp32 adapters on all attention and MLP linears (83,886,080 parameters); base frozen in fp16 with an fp32 residual stream, layers split at 16 over the two GPUs of a T4$\times$2; per-layer gradient checkpointing \\ %
budget & FE: a 257-position pass counts 1, or 3 with a backward. Mix-FT: $K=4$, $M=8$, 8 examples (window, block) per update with 25 rows each, a plan-weight pass without gradient and a weighted gradient pass (4 FE per row), 600 updates, 480,000 FE; Mix-FT $M=32$ (seed 0): 4 examples per update, 309 updates, 479,568 FE; continued CE and Alpha ($\kappa=0.5$): 267 rows per update at the four-step random-order states, 600 updates, 480,600 FE. Each is 0.0018\% of pretraining, counting a forward and a backward for each of the 2.3T pretraining tokens \\ %
optimizer & AdamW (0.9, 0.999), no weight decay, lr $2\times10^{-4}$, 5\% warmup, cosine to 0, clip 1, fp16 loss scaling from $2^{12}$, micro-batch 8 rows; window order seed $1000+s$, LoRA init seed $7000+s$ \\ %
temperatures & per model, planner and $K$: a bracketing pass on 32 prompts, then four main temperatures where the entropy curve reaches real text's per-sample entropy $+0.10$, $-0.10$, $-0.25$ and $-0.40$ nats, 256 continuations each, and further files until real text's entropy and 0.3 nats below it are both bracketed; the start's $K=32$ sweep is seeded from its $K=16$ sweep \\ %
judges & GPT-2-large, Pythia-1.4B and Qwen2.5-7B (read) and Qwen2.5-1.5B (reported) on the decoded text, re-tokenized, with the prompt as context; content tokens only; Qwen2.5-7B in fp16 after an fp32 check; MAUVE on GPT-2-large features of prompt and continuation \\ %
bound & IWAE-8 of the block-32 random-order sampler at $K=4$, 8, 16 on the 256 evaluation windows (one generated block per window), with the median effective sample size of the 8 plan weights \\ %
\bottomrule
\end{tabularx}
\end{table}

\begin{table*}[tp]
\centering
\caption{LLaDA-8B-Base fine-tuned with LoRA for OpenWebText prefix continuation (setting in \cref{tab:repro8b}). (a) Means of 3 seeds (start: one model; $M=32$: seed 0) of the IWAE-8 bound of the $K$-step random-order sampler (nats/token) and of the GPT-2-large content-token NLL of the continuations at matched per-sample entropy (nats/token, lower is better, 256 continuations per temperature), for each planner. Bold: best entry in the column among the first four rows. (b) Each model at $K$ steps against the start at $2K$, judge NLL of the start minus the model's (positive: the model is better), per seed, with the seed mean, its 95\% bootstrap interval over prompts, and the verdict of the rule fixed before the runs (\cref{app:8b}).}
\label{tab:results8b}
\footnotesize
\setlength{\tabcolsep}{3pt}
\begin{tabular}{lccccccccccc}
\toprule
\multicolumn{12}{l}{\emph{(a) seed means}} \\
 & \multicolumn{3}{c}{sampler bound} & \multicolumn{4}{c}{GPT-2-large, random order} & \multicolumn{4}{c}{GPT-2-large, low confidence} \\
\cmidrule(lr){2-4}\cmidrule(lr){5-8}\cmidrule(lr){9-12}
model & $K=4$ & 8 & 16 & $K=4$ & 8 & 16 & 32 & $K=4$ & 8 & 16 & 32 \\
\midrule
Start (LLaDA-8B-Base) & 3.085 & 2.792 & 2.669 & 5.95 & 5.01 & 3.98 & 3.54 & 6.38 & 5.72 & 5.02 & 4.48 \\
Continued CE & 3.050 & 2.757 & 2.635 & 6.27 & 4.88 & 3.89 & -- & 6.48 & 5.69 & 4.83 & -- \\
Alpha ($\kappa=0.5$) & \textbf{3.045} & \textbf{2.754} & \textbf{2.632} & 6.18 & 4.67 & 3.74 & -- & 6.44 & 5.55 & 4.76 & -- \\
Mix-FT ($K=4$, $M=8$) & 3.048 & 2.761 & 2.643 & \textbf{5.85} & \textbf{4.29} & \textbf{3.48} & -- & \textbf{6.19} & \textbf{5.17} & \textbf{4.38} & -- \\
\midrule
Mix-FT $M=32$ (seed 0) & 3.058 & 2.775 & 2.650 & 5.73 & 4.21 & 3.41 & -- & 6.12 & 5.13 & 4.34 & -- \\
\bottomrule
\end{tabular}

\vspace{6pt}
\begin{tabular}{llcccccl}
\toprule
\multicolumn{8}{l}{\emph{(b) half the steps: each model at $K$ against the start at $2K$}} \\
model, planner, steps & judge & seed 0 & seed 1 & seed 2 & seed mean & 95\% interval & verdict \\
\midrule
Mix-FT, low conf., 8 vs 16 & GPT-2-large & $-0.129$ & $-0.196$ & $-0.111$ & $-0.145$ & $[-0.245, -0.044]$ & not reached \\
 & Pythia-1.4B & $-0.084$ & $-0.145$ & $-0.061$ & $-0.097$ & $[-0.196, -0.001]$ &  \\
 & Qwen2.5-7B & $-0.090$ & $-0.143$ & $-0.065$ & $-0.099$ & $[-0.182, -0.018]$ &  \\
 & Qwen2.5-1.5B & $-0.094$ & $-0.150$ & $-0.077$ & $-0.107$ & $[-0.191, -0.025]$ & (reported) \\
 & $M=32$, seed 0, GPT-2-large & $-0.109$ & & & & $[-0.215, -0.008]$ & \\
\addlinespace[2pt]
Mix-FT, random, 8 vs 16 & GPT-2-large & $-0.348$ & $-0.309$ & $-0.285$ & $-0.314$ & $[-0.413, -0.208]$ & not reached \\
 & Pythia-1.4B & $-0.275$ & $-0.223$ & $-0.217$ & $-0.238$ & $[-0.322, -0.126]$ &  \\
 & Qwen2.5-7B & $-0.210$ & $-0.170$ & $-0.163$ & $-0.181$ & $[-0.253, -0.062]$ &  \\
 & Qwen2.5-1.5B & $-0.219$ & $-0.177$ & $-0.171$ & $-0.189$ & $[-0.264, -0.068]$ & (reported) \\
 & $M=32$, seed 0, GPT-2-large & $-0.230$ & & & & $[-0.337, -0.114]$ & \\
\addlinespace[2pt]
Mix-FT, low conf., 16 vs 32 & GPT-2-large & $+0.108$ & $+0.061$ & $+0.143$ & $+0.104$ & $[+0.010, +0.208]$ & match \\
 & Pythia-1.4B & $+0.138$ & $+0.090$ & $+0.173$ & $+0.134$ & $[+0.040, +0.229]$ &  \\
 & Qwen2.5-7B & $+0.137$ & $+0.096$ & $+0.168$ & $+0.134$ & $[+0.045, +0.224]$ &  \\
 & Qwen2.5-1.5B & $+0.130$ & $+0.088$ & $+0.154$ & $+0.124$ & $[+0.034, +0.216]$ & (reported) \\
 & $M=32$, seed 0, GPT-2-large & $+0.142$ & & & & $[+0.034, +0.254]$ & \\
\addlinespace[2pt]
Mix-FT, random, 16 vs 32 & GPT-2-large & $+0.045$ & $+0.080$ & $+0.077$ & $+0.067$ & $[-0.018, +0.154]$ & match \\
 & Pythia-1.4B & $+0.077$ & $+0.112$ & $+0.120$ & $+0.103$ & $[+0.021, +0.186]$ &  \\
 & Qwen2.5-7B & $+0.132$ & $+0.188$ & $+0.178$ & $+0.166$ & $[+0.088, +0.250]$ &  \\
 & Qwen2.5-1.5B & $+0.136$ & $+0.186$ & $+0.178$ & $+0.167$ & $[+0.090, +0.252]$ & (reported) \\
 & $M=32$, seed 0, GPT-2-large & $+0.137$ & & & & $[+0.036, +0.246]$ & \\
\addlinespace[2pt]
CE, low conf., 8 vs 16 & GPT-2-large & $-0.721$ & $-0.664$ & $-0.618$ & $-0.668$ & -- & \\
 & Pythia-1.4B & $-0.668$ & $-0.609$ & $-0.566$ & $-0.614$ & -- & \\
CE, random, 8 vs 16 & GPT-2-large & $-0.913$ & $-0.905$ & $-0.902$ & $-0.907$ & -- & \\
 & Pythia-1.4B & $-0.845$ & $-0.825$ & $-0.833$ & $-0.834$ & -- & \\
Alpha, low conf., 8 vs 16 & GPT-2-large & $-0.515$ & $-0.545$ & $-0.516$ & $-0.525$ & -- & \\
 & Pythia-1.4B & $-0.477$ & $-0.519$ & $-0.481$ & $-0.492$ & -- & \\
Alpha, random, 8 vs 16 & GPT-2-large & $-0.694$ & $-0.683$ & $-0.705$ & $-0.694$ & -- & \\
 & Pythia-1.4B & $-0.642$ & $-0.641$ & $-0.639$ & $-0.641$ & -- & \\
\bottomrule
\end{tabular}

\end{table*}

\begin{table*}[tp]
\centering
\caption{Per-seed comparisons at 8B: each arm's content-token judge NLL minus Mix-FT's at the same planner and number of steps per block (nats/token, matched per-sample entropy; positive: Mix-FT better). The start is one model, compared with each Mix-FT seed; continued CE and Alpha diffusion are paired by seed.}
\label{tab:perseed8b}
\footnotesize
\setlength{\tabcolsep}{2.6pt}
\begin{tabular}{llcccccccccccc}
\toprule
\multicolumn{14}{l}{\emph{random order}} \\
 & & \multicolumn{4}{c}{GPT-2-large} & \multicolumn{4}{c}{Pythia-1.4B} & \multicolumn{4}{c}{Qwen2.5-7B} \\
\cmidrule(lr){3-6}\cmidrule(lr){7-10}\cmidrule(lr){11-14}
against & $K$ & s0 & s1 & s2 & mean & s0 & s1 & s2 & mean & s0 & s1 & s2 & mean \\
\midrule
Continued CE & 4 & $+0.366$ & $+0.411$ & $+0.480$ & $+0.419$ & $+0.372$ & $+0.403$ & $+0.480$ & $+0.418$ & $+0.336$ & $+0.360$ & $+0.428$ & $+0.375$ \\
 & 8 & $+0.566$ & $+0.596$ & $+0.617$ & $+0.593$ & $+0.570$ & $+0.602$ & $+0.616$ & $+0.596$ & $+0.565$ & $+0.578$ & $+0.599$ & $+0.581$ \\
 & 16 & $+0.472$ & $+0.402$ & $+0.373$ & $+0.416$ & $+0.478$ & $+0.412$ & $+0.392$ & $+0.427$ & $+0.446$ & $+0.406$ & $+0.375$ & $+0.409$ \\
\addlinespace[2pt]
Alpha & 4 & $+0.268$ & $+0.303$ & $+0.411$ & $+0.327$ & $+0.284$ & $+0.329$ & $+0.414$ & $+0.342$ & $+0.243$ & $+0.281$ & $+0.323$ & $+0.282$ \\
 & 8 & $+0.346$ & $+0.374$ & $+0.420$ & $+0.380$ & $+0.366$ & $+0.419$ & $+0.423$ & $+0.403$ & $+0.334$ & $+0.384$ & $+0.370$ & $+0.362$ \\
 & 16 & $+0.247$ & $+0.322$ & $+0.221$ & $+0.263$ & $+0.268$ & $+0.348$ & $+0.255$ & $+0.290$ & $+0.238$ & $+0.342$ & $+0.238$ & $+0.273$ \\
\addlinespace[2pt]
Start & 4 & $+0.058$ & $+0.085$ & $+0.161$ & $+0.101$ & $+0.306$ & $+0.331$ & $+0.413$ & $+0.350$ & $+0.511$ & $+0.522$ & $+0.587$ & $+0.540$ \\
 & 8 & $+0.685$ & $+0.723$ & $+0.747$ & $+0.718$ & $+0.758$ & $+0.811$ & $+0.817$ & $+0.795$ & $+0.820$ & $+0.860$ & $+0.867$ & $+0.849$ \\
 & 16 & $+0.478$ & $+0.513$ & $+0.510$ & $+0.500$ & $+0.547$ & $+0.582$ & $+0.589$ & $+0.573$ & $+0.573$ & $+0.629$ & $+0.619$ & $+0.607$ \\
\addlinespace[2pt]
\bottomrule
\end{tabular}

\vspace{6pt}
\begin{tabular}{llcccccccccccc}
\toprule
\multicolumn{14}{l}{\emph{low confidence}} \\
 & & \multicolumn{4}{c}{GPT-2-large} & \multicolumn{4}{c}{Pythia-1.4B} & \multicolumn{4}{c}{Qwen2.5-7B} \\
\cmidrule(lr){3-6}\cmidrule(lr){7-10}\cmidrule(lr){11-14}
against & $K$ & s0 & s1 & s2 & mean & s0 & s1 & s2 & mean & s0 & s1 & s2 & mean \\
\midrule
Continued CE & 4 & $+0.233$ & $+0.302$ & $+0.330$ & $+0.288$ & $+0.231$ & $+0.284$ & $+0.322$ & $+0.279$ & $+0.187$ & $+0.236$ & $+0.245$ & $+0.223$ \\
 & 8 & $+0.593$ & $+0.468$ & $+0.507$ & $+0.523$ & $+0.583$ & $+0.463$ & $+0.505$ & $+0.517$ & $+0.493$ & $+0.375$ & $+0.430$ & $+0.433$ \\
 & 16 & $+0.460$ & $+0.415$ & $+0.480$ & $+0.452$ & $+0.468$ & $+0.420$ & $+0.489$ & $+0.459$ & $+0.441$ & $+0.380$ & $+0.464$ & $+0.429$ \\
\addlinespace[2pt]
Alpha & 4 & $+0.208$ & $+0.260$ & $+0.277$ & $+0.248$ & $+0.229$ & $+0.307$ & $+0.308$ & $+0.281$ & $+0.115$ & $+0.227$ & $+0.208$ & $+0.183$ \\
 & 8 & $+0.387$ & $+0.348$ & $+0.405$ & $+0.380$ & $+0.393$ & $+0.374$ & $+0.420$ & $+0.396$ & $+0.289$ & $+0.267$ & $+0.309$ & $+0.288$ \\
 & 16 & $+0.331$ & $+0.361$ & $+0.445$ & $+0.379$ & $+0.347$ & $+0.374$ & $+0.467$ & $+0.396$ & $+0.322$ & $+0.333$ & $+0.382$ & $+0.345$ \\
\addlinespace[2pt]
Start & 4 & $+0.163$ & $+0.195$ & $+0.218$ & $+0.192$ & $+0.247$ & $+0.281$ & $+0.314$ & $+0.281$ & $+0.282$ & $+0.317$ & $+0.321$ & $+0.306$ \\
 & 8 & $+0.569$ & $+0.501$ & $+0.587$ & $+0.552$ & $+0.614$ & $+0.552$ & $+0.637$ & $+0.601$ & $+0.554$ & $+0.501$ & $+0.578$ & $+0.544$ \\
 & 16 & $+0.648$ & $+0.601$ & $+0.682$ & $+0.644$ & $+0.703$ & $+0.655$ & $+0.738$ & $+0.699$ & $+0.681$ & $+0.639$ & $+0.711$ & $+0.677$ \\
\addlinespace[2pt]
\bottomrule
\end{tabular}

\end{table*}

\begin{table}[h]
\centering
\caption{Diversity of the 8B continuations at matched per-sample entropy (seed means; each model at $K=8$, Mix-FT also at $K=16$; the mean over the four entropy levels of each measure interpolated along the temperature sweep). rep4: share of repeated 4-grams within a continuation; distinct-4 and self-BLEU-4 over the 256 continuations; near-dup.: share of continuations whose 8-gram Jaccard similarity with another exceeds 0.5; Vendi on GPT-2-large features of prompt and continuation.}
\label{tab:div8b}
\footnotesize
\setlength{\tabcolsep}{3pt}
\begin{tabular}{llcccccc}
\toprule
planner & model & $K$ & rep4 & distinct-4 & self-BLEU-4 & near-dup. & Vendi \\
\midrule
random order & Start & 32 & 0.052 & 0.930 & 0.160 & 0.000 & 96.87 \\
 & Mix-FT & 16 & 0.040 & 0.937 & 0.182 & 0.000 & 103.20 \\
 & Start & 16 & 0.042 & 0.943 & 0.148 & 0.000 & 92.79 \\
 & Start & 8 & 0.024 & 0.965 & 0.115 & 0.000 & 76.70 \\
 & Mix-FT & 8 & 0.025 & 0.958 & 0.157 & 0.000 & 94.39 \\
 & Continued CE & 8 & 0.018 & 0.967 & 0.149 & 0.000 & 81.06 \\
 & Alpha & 8 & 0.024 & 0.958 & 0.157 & 0.000 & 87.62 \\
\addlinespace[2pt]
low confidence & Start & 32 & 0.045 & 0.936 & 0.159 & 0.000 & 95.26 \\
 & Mix-FT & 16 & 0.039 & 0.938 & 0.178 & 0.000 & 98.81 \\
 & Start & 16 & 0.035 & 0.950 & 0.144 & 0.000 & 82.46 \\
 & Start & 8 & 0.018 & 0.972 & 0.102 & 0.000 & 62.29 \\
 & Mix-FT & 8 & 0.025 & 0.960 & 0.142 & 0.000 & 80.43 \\
 & Continued CE & 8 & 0.015 & 0.971 & 0.134 & 0.000 & 67.11 \\
 & Alpha & 8 & 0.025 & 0.959 & 0.142 & 0.000 & 70.37 \\
\addlinespace[2pt]
& real continuations & -- & 0.028 & 0.935 & 0.207 & 0.000 & 103.00 \\
\bottomrule
\end{tabular}

\end{table}

\section{1.1B GSM8K}
\label{app:gsm8k}
We tested the 1.1B masked diffusion model of \citet{nie2025smdm} fine-tuned by its authors for GSM8K, sampled in random order at two steps. LoRA fine-tuning on its own fine-tuning data (one seed) left Mix-FT and continued CE within 0.2 accuracy points, but the model has nearly memorized that data (about 7 nats per answer under the two-step sampler). We therefore fine-tuned on the model's own verified solutions to 3,368 new questions: every fine-tune gained 1 to 3 points over the start, and Mix-FT and continued CE, at equal compute or at equal examples, did not differ (three seeds; paired difference $-0.1$ points, 95\% interval $[-1.6, +1.6]$). We found no gain from the method on this task at this scale. This is what the objective predicts for self-generated data: the sampler's own law is a fixed point of the exact mixture refit (\cref{app:toys}), so on its own samples the mixture likelihood has little to correct beyond the verification filter.

\FloatBarrier
\section{Text8 details}
\label{app:text8}

\begin{table}[h]
\centering
\caption{Text8 mechanism control ($B=8$, $K=4$, exact NLL in bpc at the best temperature). Per-path FT uses Mix-FT's sequences, plans, replicas and step count with equal plan weights. $G$: judge gain at matched distinct-4 against per-path FT. ESS: effective sample size of 8 self-normalized plan weights; gap: per-path bound minus IWAE-8, both on 512 held-out sequences. Lower rows: $M$ ablation at equal passes, seed 0.}
\label{tab:control}
\small
\begin{tabular}{lccccc}
\toprule
arm & seed 0 & seed 1 & seed 2 & $G$ vs per-path & ESS / gap (s0) \\
\midrule
Base 5$\times$ (continued CE) & 1.7944 & 1.7946 & 1.7878 & $-0.05$ to $+0.01$ & 4.33 / 0.130 \\
Per-path FT ($M=8$, equal weights) & 1.7839 & 1.7867 & 1.7760 & 0 & 4.37 / 0.129 \\
Mix-FT ($M=8$) & 1.7518 & 1.7552 & 1.7449 & $+0.14$, $+0.14$, $+0.14$ & 3.57 / 0.241 \\
\midrule
Mix-FT, $M=1$ & 1.7853 & & & $+0.01$ & 4.34 / 0.130 \\
Mix-FT, $M=2$ & 1.7645 & & & $+0.06$ & 4.01 / 0.168 \\
Mix-FT, $M=4$ & 1.7553 & & & $+0.14$ & 3.75 / 0.210 \\
\bottomrule
\end{tabular}
\end{table}

\begin{table}[h]
\centering
\caption{Text8 per seed: exact $K$-step NLL (bpc, block 8) and IWAE-64 (block 16) at the best temperature. $K=8$ rows: exact NLL at one token per step.}
\label{tab:perseed}
\small
\begin{tabular}{lccc}
\toprule
arm & seed 0 & seed 1 & seed 2 \\
\midrule
\multicolumn{4}{l}{\emph{block 8, $K=4$, exact}}\\
C0 (masked CE pretraining) & 1.9595 & 1.9540 & 1.9339 \\
Base 1$\times$ / Mix 1$\times$ (from C0) & 1.8988 / 1.9093 & 1.8862 / 1.9031 & 1.8768 / 1.8830 \\
Base 4$\times$ / Mix 4$\times$ (from C0) & 1.8076 / 1.8735 & 1.8054 / 1.8687 & 1.8005 / 1.8545 \\
Base 5$\times$ & 1.7944 & 1.7946 & 1.7878 \\
SDTT from Base 4$\times$ & 1.8041 & 1.8033 & 1.7974 \\
Hybrid from Base 4$\times$ & 1.7653 & 1.7658 & 1.7611 \\
Mix-FT from Base 4$\times$ & 1.7518 & 1.7552 & 1.7449 \\
\multicolumn{4}{l}{\emph{block 8, $K=8$, exact}}\\
Base 4$\times$ & 1.7112 & 1.7142 & 1.7036 \\
CE at $K=8$ states / Mix-FT trained at $K=8$ & 1.7011 / 1.6766 & 1.7042 / 1.6765 & 1.6896 / 1.6671 \\
Mix-FT trained at $K=4$ & 1.6742 & 1.6777 & 1.6673 \\
\multicolumn{4}{l}{\emph{block 16, $K=4$, IWAE-64}}\\
Base 5$\times$ / SDTT & 1.9608 / 1.9714 & 1.9631 / 1.9704 & 1.9636 / 1.9721 \\
Hybrid / Mix-FT & 1.9300 / 1.9220 & 1.9316 / 1.9234 & 1.9342 / 1.9245 \\
IWAE-512 (first 256 sequences): Base 5$\times$ / Mix-FT & 2.0080 / 1.9685 & 2.0089 / 1.9649 & 2.0094 / 1.9646 \\
\bottomrule
\end{tabular}
\end{table}

\paragraph{Decomposition arithmetic.} Means over seeds:
\begin{align*}
\text{$K=4$ gain:}&\quad 1.7923-1.7506=0.042,\\
\text{part that persists at $K=8$:}&\quad 1.6983-1.6731=0.025,\\
\text{gap narrowing:}&\quad (1.7923-1.6983)-(1.7506-1.6731)=0.094-0.078=0.016.
\end{align*}
The second line compares the $K=4$-trained Mix-FT with CE trained at the $K=8$ states, both sampled at $K=8$. The two parts sum to 0.041 because of rounding.

\paragraph{Width.} \Cref{tab:t8abl}a reruns the whole protocol, pretraining included, at widths 256, 512 and 768 (4.8M, 19M and 43M parameters) and splits the $K=4$ gain at each width. The wider networks are much better models: continued CE is 0.13\bpc{} lower at width 768 than at 256. The part of Mix-FT's gain that persists at one token per step shrinks with width, by 0.010 to 0.011\bpc{} in every seed against continued CE, while the narrowing of the gap between $K=4$ and $K=8$ stays level, as a capacity account of effect (ii) predicts. Against CE trained at the $K=8$ states the split is the same, and at width 256 this rerun reproduces the gains of \cref{tab:perseed} within 0.001\bpc{} per seed. %

\paragraph{Two-step training plans.} \Cref{tab:t8abl}b repeats the 110M comparison of training plans where the likelihood is exact. At $B=8$ a two-step plan reveals half a block per step, so we did not know whether the 110M result would carry over. At equal compute it does: two-step plans beat the four-step recipe on the exact $K=4$ likelihood, by 0.002 to 0.005\bpc{}, and on samples, in every seed. Unlike at 110M, the extra updates matter for the likelihood: with the four-step run's number of updates, the two-step model is 0.005\bpc{} worse at $K=4$ in every seed, though its samples are still better. Each model is best near the step count it trained for: at $K=8$ the two-step model is 0.024 to 0.028\bpc{} worse than the four-step one, at $K=2$ about 0.04 better. As at 110M, the start's plan weights are more concentrated at $K=2$ than at $K=4$ (median effective sample size of eight plans 3.60 to 3.65 against 4.21 to 4.29). %

\paragraph{More plans, and more CE.} \Cref{tab:t8abl}c trains Mix-FT with $M=8$, 32, 128 and 256 plans per sequence for the same number of updates, so its compute grows with $M$. The exact $K=4$ NLL improves in every seed up to $M=128$ ($M=32$ by 0.011 to 0.014\bpc{} over $M=8$, $M=128$ by a further 0.005 to 0.007) and is level from 128 to 256 plans (mean $-0.002$). Over the same ladder the gap between the eight-plan IWAE bound and the exact NLL widens by 0.03\bpc{}, so a bound with few evaluation plans understates the gain, as it may for the 110M 8$\times$ model (\cref{tab:bounds8x}). The samples improve along the whole ladder. Continued CE given the compute of $M=256$, as 769 times the updates, reaches a much lower exact NLL, 1.651 against 1.732\bpc{} (seed 0), but worse samples: its $G$ against Base 5$\times$ is below that of $M=8$ at one unit. More plans buy samples that more CE does not, and the exact likelihood alone does not rank these models by their samples. %

\begin{table}[h]
\centering
\caption{Text8 ablations (block 8, exact NLL in bpc at the best temperature). (a) Width: the protocol rerun at each width (means of 3 seeds). Gain: continued CE (Base 5$\times$) minus Mix-FT at $K=4$. It splits into the part that persists at one token per step (per step: both models sampled at $K=8$) and the narrowing of the gap between $K=4$ and $K=8$ (gap), against continued CE and against CE trained at the $K=8$ states. (b) Training plans: Mix-FT on two-step plans at the four-step recipe's compute (more updates) and at its number of updates (0.36 of its compute). $G$: judge gain over the four-step model at $K=4$ and matched distinct-4, per seed. (c) Plans per sequence at a fixed number of updates, with compute in units of 691,200 passes; IWAE-8 minus exact: how far the eight-plan bound lies above the exact NLL; ESS: median effective sample size of eight plan weights at $\beta=1$; $G$ against Base 5$\times$, mean of 3 seeds. The last row is continued CE given the compute of $M=256$ (seed 0).}
\label{tab:t8abl}
\footnotesize
\setlength{\tabcolsep}{4pt}
\begin{tabular}{lccccccc}
\toprule
\multicolumn{8}{l}{\emph{(a) width}} \\
 & & continued & gain & \multicolumn{2}{c}{vs continued CE} & \multicolumn{2}{c}{vs CE at $K=8$ states} \\
\cmidrule(lr){5-6}\cmidrule(lr){7-8}
width & parameters & CE, $K=4$ & at $K=4$ & per step & gap & per step & gap \\
\midrule
256 & 4.8M & 1.794 & 0.042 & 0.024 & 0.018 & 0.026 & 0.016 \\
512 & 19M & 1.683 & 0.036 & 0.018 & 0.018 & 0.019 & 0.017 \\
768 & 43M & 1.666 & 0.032 & 0.013 & 0.019 & 0.015 & 0.017 \\
\midrule
\multicolumn{8}{l}{\emph{(b) training plans}} \\
 & \multicolumn{4}{c}{exact $K=4$} & $K=8$ & $K=2$ & $G$ vs four-step, \\
\cmidrule(lr){2-5}
arm & seed 0 & seed 1 & seed 2 & mean & mean & mean & seeds 0, 1, 2 \\
\midrule
Start (Base 4$\times$) & 1.8076 & 1.8054 & 1.8005 & 1.8045 & 1.7097 & 2.1033 & -- \\
Mix-FT, four-step plans & 1.7518 & 1.7552 & 1.7449 & 1.7506 & 1.6731 & 2.0410 & reference \\
Two-step plans, equal compute & 1.7485 & 1.7500 & 1.7432 & 1.7472 & 1.6993 & 1.9984 & $+0.09$, $+0.11$, $+0.06$ \\
Two-step plans, equal updates & 1.7567 & 1.7600 & 1.7499 & 1.7555 & 1.7031 & 2.0125 & $+0.06$, $+0.10$, $+0.02$ \\
\midrule
\multicolumn{8}{l}{\emph{(c) plans per sequence at fixed updates}} \\
 & \multicolumn{4}{c}{exact $K=4$} & IWAE-8 & ESS & $G$ vs \\
\cmidrule(lr){2-5}
arm (compute) & seed 0 & seed 1 & seed 2 & mean & $-$ exact & of 8 & Base 5$\times$ \\
\midrule
Mix-FT, $M=8$ (1) & 1.7518 & 1.7552 & 1.7449 & 1.7506 & 0.063 & 3.39 & $+0.16$ \\
$M=32$ (3.9) & 1.7409 & 1.7411 & 1.7324 & 1.7382 & 0.079 & 2.95 & $+0.20$ \\
$M=128$ (15.4) & 1.7358 & 1.7346 & 1.7275 & 1.7327 & 0.093 & 2.65 & $+0.25$ \\
$M=256$ (30.8) & 1.7315 & 1.7350 & 1.7267 & 1.7311 & 0.096 & 2.56 & $+0.27$ \\
Continued CE (30.8) & 1.6509 & & & & 0.040 & 4.51 & $+0.12$ \\
\bottomrule
\end{tabular}
\end{table}

\paragraph{SDTT on text8.} SDTT \citep{deschenaux2024sdtt} from the same weights, following the paper's algorithm and optimizer with teacher passes counted, does learn its own objective: its held-out distillation KL falls. But it improves on its start by only 0.002 to 0.004\bpc{}, less than a fifth unit of CE gives, and its samples are worse than continued CE's in every seed. At the token level the same implementation removes the newline excess and is the best model at two steps per block (\cref{tab:main110,sec:limits}).

\paragraph{Why SDTT does poorly on text8.} The data point to two features of this setting, though we have not isolated either. First, at $B=8$, $K=4$ SDTT runs one round, and its teacher is the same network at one character per step, which is 0.095\bpc{} better than the student under the exact likelihood (Base 4$\times$: 1.7096 at $K=8$, 1.8045 at $K=4$). Second, its target is the teacher's per-position marginals, which by \cref{prop:fixed,prop:gap} carry no within-step dependence, and in character-level text the two characters filled in one step of an 8-character block can belong to the same word.

\paragraph{From scratch.} A gradient probe on a small CPU proxy shows that near CE-trained weights the IWAE gradient carries about as much signal per pass as CE's, but each step costs 25 passes. Once Mix has moved the weights, its plan weights become more unequal (effective sample size 3.8 to 4.1 of 8, against 4.7 to 4.9 at CE weights) and its gradient becomes 5 to 7 times noisier per pass. Raising the learning rate made it worse.

\begin{figure}[h]
\centering
\includegraphics[width=0.8\linewidth]{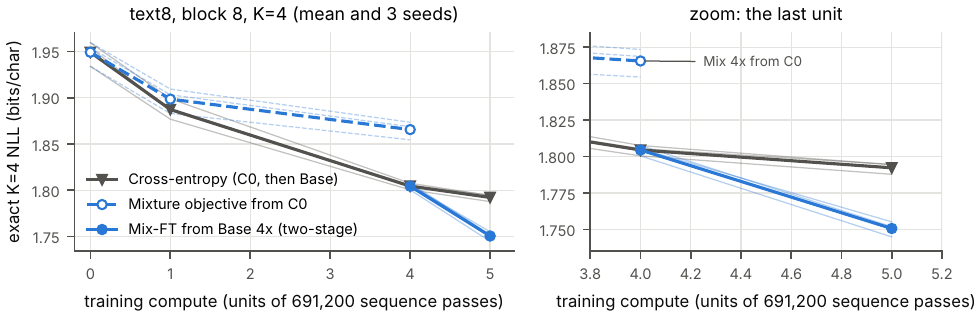}
\caption{Exact $K=4$ sampler NLL along the training budget (text8, block 8; thick: mean, thin: seeds). Trained from C0, the mixture objective falls behind CE, and the gap grows with compute. Started from the CE model at 4 units, one more unit of mixture fine-tuning beats one more unit of CE in every seed.}
\label{fig:scratch}
\end{figure}

\begin{figure}[h]
\centering
\includegraphics[width=0.8\linewidth]{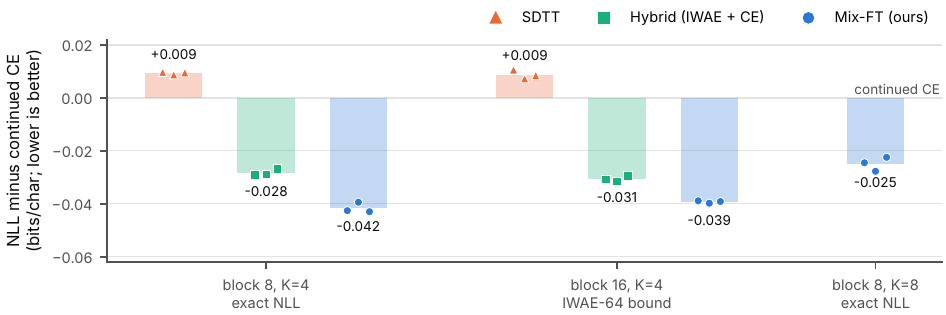}
\caption{Text8 sampler likelihood relative to continued CE from the same Base 4$\times$ weights and extra compute (bars: mean of 3 seeds; markers: seeds). Block 8 uses the exact NLL; block 16 uses IWAE-64.}
\label{fig:t8lik}
\end{figure}

\begin{figure}[h]
\centering
\includegraphics[width=0.8\linewidth]{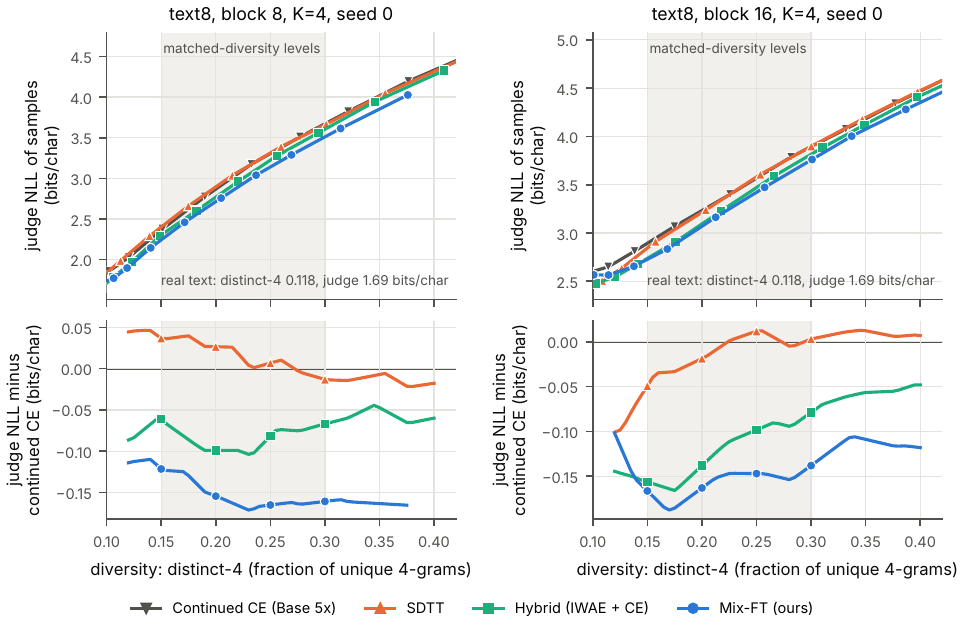}
\caption{Text8 quality against diversity, seed 0. Top: judge NLL against distinct-4, one point per temperature. Bottom: judge NLL minus that of continued CE at matched distinct-4; markers at the four levels that define $G$, whose mean is $-G$. Lower is better.}
\label{fig:t8frontier}
\end{figure}

\section{Exact toys}
\label{app:toys}

\begin{table}[h]
\centering
\caption{Exact toy ($L=8$, $V=4$, HMM mixture, entropy 8.95 nats). $\KL(p\,\|\,q)$ in nats for the $K$-step sampler with the true conditionals, after tuning a temperature on them, and after re-optimizing all conditionals for the sampler's exact likelihood. ``Fixed'' is a left-to-right plan.}
\label{tab:toy}
\small
\begin{tabular}{lcccc}
\toprule
sampler & true conditionals & best global temp. & per-step temp. & re-optimized table \\
\midrule
$K=2$, fixed plan & 1.2049 & -- & -- & 1.2049 (no change) \\
$K=2$, random order & 0.3899 & 0.2141 & 0.1138 & 0.0299 \\
$K=4$, fixed plan & 0.5158 & -- & -- & 0.5158 (no change) \\
$K=4$, random order & 0.0556 & 0.0200 & 0.0111 & 0.0000 \\
\bottomrule
\end{tabular}
\end{table}

\begin{table}[h]
\centering
\caption{Exact toy ($L=6$, $V=3$, entropy 5.79 nats): $\KL(p\,\|\,q)$ after training a free table on $\cL_M$ with fresh plans every step (1,500 Adam steps). $M=1$ is the per-path loss.}
\label{tab:iwaeM}
\small
\begin{tabular}{lcccccc}
\toprule
sampler & true cond. & $M=1$ & $M=4$ & $M=16$ & exact mixture & best temperature \\
\midrule
$K=3$, random order & 0.0440 & 0.0436 & 0.0103 & 0.0025 & 0.0000 & 0.0096 \\
$K=2$, random order & 0.1481 & 0.1462 & 0.0380 & 0.0085 & 0.0008 & 0.0319 \\
\bottomrule
\end{tabular}
\end{table}

\begin{figure}[h]
\centering
\includegraphics[width=0.6\linewidth]{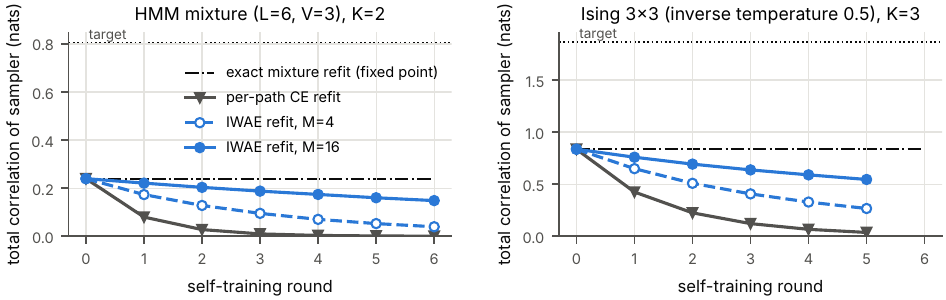}
\caption{Self-training on exact toys. Each round refits the sampler's conditionals to its own law. The per-path refit removes dependence round after round; the IWAE refit drifts more slowly as $M$ grows; the exact mixture refit is a fixed point. Dotted: total correlation of the target.}
\label{fig:toy}
\end{figure}

\paragraph{Single-trajectory objectives.} When fitting a target $\pi^\star$ from rewards or energies,
$\KL\big(q_\theta(x,d)\,\|\,\pi^\star(x)\mu(d)\big) = \KL\big(q_\theta\,\|\,\pi^\star\big) + \E_{x\sim q_\theta}\KL\big(\pi_\theta(\cdot\mid x)\,\|\,\mu\big)$.
The extra term vanishes only when every plan yields the same law. Plan invariance does not imply independence, but in the Ising toy the factorized sampler reached it by filling positions nearly independently. On a $3\times3$ Ising model at inverse temperature 0.9 ($K=3$), the path objective collapses the sampler's total correlation from 1.72 to 0.17 nats, while the marginal objective keeps 1.42 (\cref{tab:orderinv}). Self-training is a second case: per-token CE on the sampler's own samples converges to the true conditionals of $q_\theta$, and iterating it re-factorizes the law each round. The per-path refit turns the $K=2$ HMM sampler into a near-product law in five rounds; the IWAE refit drifts at a rate that falls with $M$, and the exact mixture refit is a fixed point (\cref{fig:toy}). A test of reward-filtered self-training on Sudoku with a 5M-parameter transformer (two seeds) found no advantage for the IWAE refit.

\begin{table}[h]
\centering
\caption{Fitting a target from its energy alone: $\KL(q\,\|\,\pi^\star)$ and total correlation (TC; 0 is a product law), nats, after 600 (HMM) or 300 (Ising) Adam steps from the true conditionals. The path objective had converged; the marginal one was still improving slowly.}
\label{tab:orderinv}
\small
\begin{tabular}{llccc}
\toprule
target, $K$ & objective & $\KL(q\,\|\,\pi^\star)$ & TC & TC of $\pi^\star$ \\
\midrule
HMM ($L=6$), $K=2$ & true conditionals & 0.1375 & 0.240 & \multirow{3}{*}{0.806} \\
 & path reverse KL & 0.1197 & 0.268 & \\
 & marginal reverse KL & 0.0008 & 0.769 & \\
\midrule
Ising $3\times3$, inv.\ temp.\ 0.9, $K=3$ & true conditionals & 1.144 & 1.720 & \multirow{3}{*}{4.546} \\
 & path reverse KL & 0.680 & 0.171 & \\
 & marginal reverse KL & 0.573 & 1.419 & \\
\bottomrule
\end{tabular}
\end{table}

The HMM targets are mixtures of sticky hidden Markov chains ($L=6$, $V=3$: 729 sequences; $L=8$, $V=4$: 65,536 sequences). The model is a free logit table over partial states, the same conditioning a network has; Adam minimizes the exact objective through a differentiable dynamic program from the true-conditional start. In \cref{tab:toy} the $K=2$ random-order value is still falling after 150 Adam steps (0.0299); at $L=6$ it reaches 0.0008 after 2,000 steps.

\section{Exact dynamic program}
\label{app:dp}
For one block with $B$ positions, $K$ steps and $g=B/K$, let $\mathcal{T}_k$ be the subsets of size $kg$. For $T\in\mathcal{T}_k$, $f(T)$ is the total probability that the sampler has revealed exactly $T$ after $k$ steps with tokens $x_T$, summed over ordered partitions of $T$:
\[
\log f(T) = \operatorname{logsumexp}_{S\subset T,\,|S|=(k-1)g}\Big[\log f(S) + \sum_{i\in T\setminus S}\log r^{\beta}_\theta(x_i\mid x_S)\Big],\qquad q_\theta(x)=\frac{f(\{1,\dots,B\})}{B!/(g!)^K}.
\]
One forward pass per reveal set gives $\log r^\beta_\theta$ at all positions and temperatures; blocks share passes because they are conditionally independent given the clean prefix. Reveal sets per sequence: 127 ($B=8$, $K=4$), 255 ($B=8$, $K=8$), 16,511 ($B=16$, $K=4$, not affordable; we report IWAE-64 and IWAE-512 there). A self-test against enumeration of all plans for several small $(B,K)$ gives errors below $1.6\times10^{-6}$ nats.

\section{Extended related work: PUMBA}
\label{app:related}
PUMBA \citep{madeira2026pumba} trains the denoiser on consecutive steps of the few-step policy's own trajectories, with teacher-forced commits, a carry state and truncated backpropagation through time, scales this to fine-tuning LLaDA-8B, and argues that training should follow the states the sampler visits, finding that a looser alignment helps where exact alignment overfits. Three points reconcile it with our argument. First, when the next group is a deterministic function of the revealed tokens, the per-trajectory likelihood is the sampler's likelihood, so trajectory training targets the right quantity for such samplers; our argument concerns random-order samplers, and PUMBA's planner reads the model's own confidence, which \cref{prop:fixed} does not cover. Second, the carry state makes the conditionals depend on the trajectory, not only on the revealed set; \cref{prop:gap} still holds for any mixture over plans, but \cref{eq:sampler} and our dynamic program do not. Third, PUMBA's critique targets random-masking CE, whose states do not match the sampler's; our CE baseline trains at the sampler's own states and is the closest analogue of trajectory training without a carry, and the mixture fine-tune beats it. A comparison under a policy-driven sampler remains open.

\FloatBarrier
\section{Hyperparameters}
\label{app:hparams}

\begin{center}
\small
\begin{tabularx}{\linewidth}{@{}l>{\raggedright\arraybackslash}X@{}}
\toprule
\multicolumn{2}{l}{\emph{110M OpenWebText}}\\
start & BD3-LM, OpenWebText, block 16 (\texttt{kuleshov-group/bd3lm-owt-block\_size16}); 12 layers, width 768, 12 heads, 110M parameters as reported by \citet{arriola2025bd3lm} (169.6M stored values, 92.4M outside the input embedding and output layer); $L=1024$, GPT-2 tokenizer; NELBO perplexity 23.0 on our held-out split \\
data & OpenWebText; training sequences from the first of 80 parquet shards; held-out: the last 3,000 documents of the last shard \\
equal compute & 276,480 forward-equivalents (FE) per arm; batch $8\times1024$ tokens \\
forward-equivalents & a 1,024-token pass counts 1, or 3 with a backward; BD3-LM's pass over the clean context counts too, so a CE sequence costs 6 FE and a Mix-FT sequence $3(2+M(K-1))$. The checkpoint was pretrained for 1M steps of 512 sequences (524B tokens; 850K steps at block 1,024 and 150K at block 16; \citealp{arriola2025bd3lm}), at least $1.76\times10^9$ FE by this count, so the 1$\times$ budget is under 0.02\% of it and the 8$\times$ budget under 0.13\% (0.18\% as the 8$\times$ Mix-FT runs executed it; next row) \\
chunked runs & runs with $M\ge128$ do not fit a T4's memory in one pass. The code first computes the plan weights without gradient, then repeats the forward pass with gradient in chunks of 32 plans (\cref{app:repro}), which executes about 1.4 times the counted FE (1,088 instead of 774 per sequence at $M=256$) and gives the same loss and gradient as one pass (largest relative gradient difference $7\times10^{-15}$ in fp64 at $M=256$). We count the single gradient pass that the method needs. A block-chunked backward, which puts all plans of a block in one chunk instead of chunking by plans, gives each block's weights inside its chunk and removes the pass without gradient: our implementation of it executes exactly the counted 774 FE per sequence at $M=256$, holds 32 rather than 34 sequence-equivalents per forward pass, and gives the single-pass loss and gradient (largest relative gradient difference $1.1\times10^{-14}$ in fp64). At 1.1B every Mix-FT run takes that pass, and there we count it (4 FE per row) \\
Mix-FT (main) & $K=2$ plans, $M=8$, 1,152 steps; ablations: $K=4$/$M=8$ 443 steps, $K=4$/$M=3$ 1,047, $K=4$/$M=2$ 1,440, $K=2$/$M=4$ 1,920 \\
continued CE, Alpha & 5,760 steps, CE (Alpha: its reweighted CE) at the $K=4$ sampler's states, same batches; Alpha block weights centered by a constant measured once on the start model; Alpha $\kappa\in\{0.1,0.25,0.5,2,8\}$, best $\kappa=0.5$ \\
optimizer (CE, Mix-FT, hybrid, Alpha) & AdamW (0.9, 0.999), no weight decay, peak lr $5\times10^{-5}$, warmup $\min(100,\lfloor\text{steps}/8\rfloor)$, cosine to 10\%, clip 1, fp16 autocast \\
SDTT-4 / SDTT-8 & 2 rounds of 1,920 steps (16 to 8 to 4 steps per block) / 1 round of 3,840 (16 to 8); reverse KL, lr $6\times10^{-5}$ constant after warmup, EMA 0.999, teacher passes counted \\
likelihood & IWAE-32 of the $K$-step sampler, $K\in\{4,2\}$, 256 held-out sequences, $\beta\in\{1,1.1,1.25\}$, best reported; NELBO by the order-averaged estimator \\
generation & 512 samples of 1,024 tokens per temperature, $\beta\in\{0.95,1,1.05,1.1,1.15,1.25\}$, $K\in\{2,4,8,16\}$; exact categorical sampling in fp32; one T4, batch 64 \\
judges & GPT-2-large (fp32) and Pythia-1.4B (fp16, checked against fp32); content tokens exclude newline and end-of-text targets; matched per-sample entropy at real $-0.3$, $-0.2$, $-0.1$, $0$ nats \\
\bottomrule
\end{tabularx}
\end{center}

\begin{center}
\small
\begin{tabularx}{\linewidth}{@{}l>{\raggedright\arraybackslash}X@{}}
\toprule
\multicolumn{2}{l}{\emph{text8}}\\
model & 6 layers, $d=256$, 4 heads, 4.82M parameters; sequences of 256 characters; block-causal attention over [clean $|$ noisy] \\
C0 & 20,000 steps, lr $10^{-3}$, warmup 200, cosine to $10^{-4}$ \\
one unit & 691,200 sequence passes: 21,600 steps of Base CE (lr $3\times10^{-4}$) or 864 of Mix-FT ($M=8$, 25 passes per sequence) \\
optimizer & AdamW (0.9, 0.95), weight decay 0.01, clip 1.0, batch 32, fresh optimizer state at every restart \\
SDTT & reverse KL, dense Algorithm-1 targets, lr $6\times10^{-5}$, no weight decay, EMA 0.999, $\log_2(B/K)$ rounds, teacher passes counted \\
evaluation & exact NLL ($B=8$) or IWAE-64 ($B=16$) at 10 temperatures, IWAE-512 ($B=16$) at 4; 512 samples at $\beta\in\{0.8,\dots,4\}$; judge: causal character transformer (4 layers, $d=256$) \\
\bottomrule
\end{tabularx}
\end{center}

\section{Reproducing the half-step, baseline, 1.1B and rater experiments}
\label{app:repro}

All training, sampling and judging with open models ran on Kaggle T4 GPUs (16 GB), one job per GPU and usually two jobs per T4$\times$2 kernel; the 8$\times$ Mix-FT runs (stage 20) and the schedule jobs (stage 21) had a kernel each. A job longer than one kernel (one Kaggle session, 11 to 11.8 h) checkpointed and resumed in the next. Stage numbers are those of \cref{app:rules}. %
Each stage's scripts check the md5 of every imported earlier script and refuse to run otherwise; the md5s are listed in the stage's handoff file.

\paragraph{The 8$\times$ half-step model (110M).} \Cref{tab:repro8x} gives the recipe. It is the 1$\times$ two-step recipe of \cref{app:hparams} with $M=256$ plans per sequence and 8 times the budget. Each update evaluates the 256 plans in 8 chunks of 32: a pass without gradient over all chunks gives the self-normalized plan weights, then a pass with gradient backpropagates $-\sum_m \tilde w_m \log q_{\theta,d_m}(x)$ chunk by chunk. This is the exact IWAE gradient (relative difference to the unchunked one $7\times10^{-15}$ in fp64) and keeps every forward at 34 rows. %

\begin{table}[h]
\centering
\caption{Recipe of the 8$\times$ model (\texttt{mixft\_k2m256x8}, seeds 0, 1, 2).}
\label{tab:repro8x}
\footnotesize
\begin{tabularx}{\linewidth}{@{}l>{\raggedright\arraybackslash}X@{}}
\toprule
start & the public BD3-LM checkpoint of \cref{app:hparams} (no earlier fine-tune) \\
plans & $K=2$ steps per block in training, $M=256$ plans per sequence, batch 8 sequences of 1,024 tokens, micro-batch 1 \\ %
budget & $3(2+M(K-1))=774$ FE per sequence; 357 updates; 2,210,544 FE $=7.9953\times$ the 1$\times$ budget of 276,480 (under 0.13\% of pretraining); the chunked backward executes 1,088 FE per sequence, 3,107,328 in all, 0.18\% of pretraining (\cref{app:hparams}) \\ %
sequences seen & 2,856, a prefix of the 1$\times$ run's data order (no sequence repeats); same plan generator (RNG code 52, seed $1000s+52$) \\ %
optimizer & AdamW (0.9, 0.999), no weight decay, peak lr $5\times10^{-5}$, warmup $\min(100,\lfloor357/8\rfloor)=44$ steps, cosine to 10\% over the 357 steps, clip 1, fp16 autocast, gradient checkpointing \\ %
result & IWAE-32 at $K=4$: 3.2141, 3.2128, 3.2145; at $K=2$: 3.7293, 3.7289, 3.7300; NELBO PPL 28.82, 28.68, 28.85 (seeds 0, 1, 2) \\ %
\bottomrule
\end{tabularx}
\end{table}

\paragraph{Global temperature (stage 20).} Every model is sampled at $\beta\in\{0.95,1,1.05,1.1,1.15,1.25\}$, 512 samples of 1,024 tokens per $\beta$, with the same sampling seeds for every model. %
For each judge we compute the content-token NLL and the per-sample entropy (the entropy of each sample's own token frequencies, averaged over samples) at every $\beta$. We then interpolate the NLL along the model's $\beta$ sweep at four entropy levels, real text's entropy minus 0.3, 0.2, 0.1 and 0 nats, and average the four values. Real held-out text has 5.397 nats, so the levels are 5.097 to 5.397, and at least 3 of the 4 must be bracketed by the sweep. %
$H_8$ is this value for the start at $K=16$ minus that of the model at $K=8$ (positive: $K=8$ better).

\paragraph{Per-step schedules (stage 21).} Step $k\in\{0,\dots,K-1\}$ uses $\beta_k=b\exp\!\big(\alpha u_k+\gamma(u_k^2-\overline{u^2})\big)$ with $u_k=2k/(K-1)-1$, so $b$ is the geometric mean. There are nine shapes, $\alpha\in\{-0.15,0,0.15\}\times\gamma\in\{-0.3,0,0.3\}$, each swept over $b\in\{0.95,1,1.05,1.1,1.15,1.25,1.35,1.5\}$. %
The shape is chosen per model on a development split of 256 samples per (shape, $b$), drawn with the evaluation seeds plus 15,000,000 and so disjoint from every evaluation seed. We pick the shape with the largest mean over the two judges of its matched-entropy gain over the constant schedule if that mean is positive, and the constant schedule otherwise. %
The chosen shape is then sampled afresh (512 samples per $b$, evaluation seeds) and only these fresh rows enter the reading. All three 8$\times$ seeds and the start chose $\alpha=0.15$, $\gamma=-0.3$. %

\paragraph{Bootstrap.} The bootstrap uses $B=2{,}000$ replicates. In each, every (model, $K$, $\beta$) file of 512 samples is resampled with replacement, independently across files, and the same indices are applied to the per-sample entropy and to every judge's per-sample content NLL sums and counts; the whole interpolation is then recomputed. The start's files are resampled once per replicate and shared by the three seeds, so the seed mean carries that dependence; the real-text entropy is held fixed; a replicate with fewer than 3 bracketed levels counts as $-\infty$. Intervals are 95\% percentiles. %

\paragraph{Third judge and margin.} The third judge is Qwen2.5-1.5B base (commit \texttt{8faed761}, fp16 after an fp32 check), scoring content tokens by the Pythia judge's rule applied to its own tokenizer. %
The equivalence (noninferiority) margin is $-0.03$ nats/token on the seed-mean $H_8$: the reading is ``within the margin'' if the lower 95\% bound exceeds $-0.03$ under both GPT-2-large and Pythia, and ``better'' if it exceeds 0 under both. The margin was fixed before any bootstrap replicate or third-judge score was computed; the point estimates of $H_8$ were known by then. %
\Cref{tab:h8} gives $H_8$ per seed and the seed mean with its interval, under each judge and in both temperature settings. %
The one-sided bootstrap $p$ of $H_8\le-0.03$ is at most 0.002 for every seed mean and judge. %
These readings are for samples whose first token is drawn. With the first token fixed for both models, the same bootstrap puts the seed-mean lower bound below $-0.03$ under both judges, so there $H_8$ is neither within the margin nor better (\cref{tab:h8}). %

\paragraph{Share of the gap, and all tokens.} The same bootstrap gives an interval for the share of the start's own gap between $K=8$ and 16 that the 8$\times$ model closes, $1+H_8/(\text{start at }K=8\text{ minus start at }K=16)$, with both terms recomputed in every replicate. With the first token fixed for both models it is 95\% $[92, 99]$ under GPT-2-large and 97\% $[94, 102]$ under Pythia; with it drawn, 102\% $[99, 106]$ and 105\% $[101, 110]$. Scoring all tokens, newline and end-of-text targets included, moves $H_8$ toward the 8$\times$ model, whose $K=8$ samples carry more newlines than the start's at $K=16$ (5.5\% against 4.2\% at matched entropy, \cref{tab:newline}): with the first token fixed, the all-token seed-mean $H_8$ is $-0.023$ $[-0.051, +0.009]$ under GPT-2-large and $-0.011$ $[-0.046, +0.026]$ under Pythia, and with it drawn $+0.049$ $[+0.018, +0.080]$ and $+0.065$ $[+0.031, +0.100]$. Content tokens stay our primary measure because they do not reward the newline excess. %

\paragraph{The 16$\times$ model (110M).} It doubles the 8$\times$ recipe once more: $M=512$ plans per sequence, 358 updates of eight sequences, $3(2+M(K-1))=1{,}542$ FE per sequence and 4,416,288 in all, 16 times the 1$\times$ budget and 0.25\% of pretraining. It ran on two T4s with the block-chunked backward of \cref{app:hparams}, which executes exactly the counted FE; everything else is the 8$\times$ recipe. \Cref{tab:h8} gives its $H_8$ under both first-token conventions, with the bootstrap above. Every seed gains 0.03 to 0.08 over the 8$\times$ model under either convention. With the first token drawn its 8 steps beat the start's 16 in every seed under both judges. With it fixed, the start's convention, the seed means are $-0.005$ under GPT-2-large and $+0.016$ under Pythia, and the GPT-2-large interval misses the $-0.03$ margin by 0.006. At equal steps it beats the start by 0.58 to 0.62 nats/token with the first token fixed. Its IWAE-32 bound is level with the start's at $K=4$ (0.009 above the 8$\times$ model's) and 0.04 nats/token below it at $K=2$, and its NELBO perplexity rises to 29.8 (8$\times$: 28.8; start: 23.0). %

\begin{table}[h]
\centering
\caption{$H_8$ of the 8$\times$ and 16$\times$ models at 110M: the content-token judge NLL of the start at $K=16$ minus that of the fine-tuned model at $K=8$, at matched per-sample entropy (nats/token; positive: 8 steps better), per seed and as the seed mean with its 95\% bootstrap interval over samples. Global: one temperature per model. Schedule: each model's own per-step temperature schedule, chosen on a development split. The first two blocks draw the first token of each sequence, as every table does; the next two fix it to end-of-text for both models (\cref{app:conv}) and were not scored by the third judge. The last two give the 16$\times$ model with one temperature under each convention.}
\label{tab:h8}
\footnotesize
\setlength{\tabcolsep}{3pt}
\begin{tabular}{llccccc}
\toprule
setting & judge & seed 0 & seed 1 & seed 2 & seed mean & 95\% interval \\
\midrule
global & GPT-2-large & $+0.053$ & $+0.011$ & $+0.002$ & $+0.022$ & $[-0.011, +0.054]$ \\
 & Pythia-1.4B & $+0.076$ & $+0.040$ & $+0.025$ & $+0.047$ & $[+0.011, +0.083]$ \\
 & Qwen2.5-1.5B & $+0.079$ & $+0.063$ & $+0.050$ & $+0.064$ & $[+0.030, +0.098]$ \\
 & MAUVE at $K=8$ (start at $K=16$: 0.948) & 0.940 & 0.934 & 0.926 & 0.933 & -- \\
\midrule
schedule & GPT-2-large & $+0.025$ & $+0.024$ & $+0.016$ & $+0.022$ & $[-0.004, +0.050]$ \\
 & Pythia-1.4B & $+0.041$ & $+0.045$ & $+0.032$ & $+0.039$ & $[+0.012, +0.069]$ \\
 & Qwen2.5-1.5B & $+0.035$ & $+0.050$ & $+0.037$ & $+0.041$ & $[+0.012, +0.071]$ \\
 & MAUVE at $K=8$ (start at $K=16$: 0.952) & 0.951 & 0.949 & 0.945 & 0.948 & -- \\
\midrule
global, & GPT-2-large & $-0.032$ & $-0.058$ & $-0.038$ & $-0.043$ & $[-0.073, -0.009]$ \\
first token fixed & Pythia-1.4B & $-0.012$ & $-0.038$ & $-0.018$ & $-0.023$ & $[-0.059, +0.017]$ \\
 & MAUVE at $K=8$ (start at $K=16$: 0.942) & 0.943 & 0.940 & 0.940 & 0.941 & -- \\
\midrule
schedule, & GPT-2-large & $-0.060$ & $-0.066$ & $-0.053$ & $-0.060$ & $[-0.085, -0.035]$ \\
first token fixed & Pythia-1.4B & $-0.057$ & $-0.064$ & $-0.051$ & $-0.057$ & $[-0.082, -0.030]$ \\
 & MAUVE at $K=8$ (start at $K=16$: 0.947) & 0.946 & 0.952 & 0.958 & 0.952 & -- \\
\midrule
16$\times$, global, & GPT-2-large & $+0.100$ & $+0.058$ & $+0.078$ & $+0.079$ & $[+0.047, +0.113]$ \\
first token drawn & Pythia-1.4B & $+0.126$ & $+0.087$ & $+0.105$ & $+0.106$ & $[+0.071, +0.144]$ \\
 & MAUVE at $K=8$ (start at $K=16$: 0.948) & 0.945 & 0.933 & 0.954 & 0.944 & -- \\
\midrule
16$\times$, global, & GPT-2-large & $+0.012$ & $-0.024$ & $-0.004$ & $-0.005$ & $[-0.036, +0.028]$ \\
first token fixed & Pythia-1.4B & $+0.034$ & $-0.005$ & $+0.019$ & $+0.016$ & $[-0.020, +0.056]$ \\
 & MAUVE at $K=8$ (start at $K=16$: 0.942) & 0.951 & 0.942 & 0.949 & 0.947 & -- \\
\bottomrule
\end{tabular}
\end{table}

\paragraph{The 8$\times$ baselines (110M).} \Cref{tab:repro8xbase} lists them. All use the 8$\times$ budget, the same token cache and peak learning rate as their 1$\times$ versions, and sequences beyond the 110,721 of the cache from a fixed permutation of OpenWebText shards 1 to 4, so no sequence repeats. %
Stage 22 evaluates the final weights. Stage 25 instead saves the weights at 1/4, 1/2, 3/4 and all of the steps (EMA weights for SDTT) and samples each checkpoint at $K=8$ on a development split (256 samples at each of the six global temperatures, evaluation seeds plus 15,000,000). It then evaluates one checkpoint only: the one with the largest mean matched-entropy gain over the final checkpoint under the two judges if that gain is positive, and otherwise the final one. %
In seed 0 SDTT-8 kept its final checkpoint and Alpha-2 its half-budget one, in seed 1 both kept the quarter-budget one, and in seed 2 SDTT-8 kept the half-budget one and Alpha-2 the quarter-budget one. %

\begin{table}[h]
\centering
\caption{The 8$\times$ baselines. FE as in \cref{app:hparams}; SDTT counts its teacher's passes (9 FE per sequence). $H_8$: GPT-2-large / Pythia content, against the start at $K=16$. Gain: the same comparison at $K=8$ against the arm's own 1$\times$ run in the same seed (Mix-FT: the main 1$\times$ model; Alpha-2 in seeds 1 and 2: the seed-0 1$\times$ run, the only one). Rows without a seed are seed 0.}
\label{tab:repro8xbase}
\footnotesize
\setlength{\tabcolsep}{3pt}
\begin{tabularx}{\linewidth}{@{}l>{\raggedright\arraybackslash}Xrrrrrr@{}}
\toprule
arm & recipe & steps & FE & warmup & sequences & $H_8$ & gain \\
\midrule
continued CE & CE at the $K=4$ sampler's states, lr $5\times10^{-5}$ cosine to 10\% over all steps & 46,080 & 2,211,840 & 100 & 368,640 & $-1.24$ / $-1.25$ & $-0.18$ / $-0.20$ \\ %
\quad seed 1 & as seed 0 & 46,080 & 2,211,840 & 100 & 368,640 & $-1.21$ / $-1.21$ & $-0.09$ / $-0.09$ \\ %
\quad seed 2 & as seed 0 & 46,080 & 2,211,840 & 100 & 368,640 & $-1.22$ / $-1.22$ & $-0.08$ / $-0.10$ \\ %
Alpha & as CE, Alpha loss $\kappa=0.5$, centering constant from the start model ($c=-3.608$) & 46,080 & 2,211,840 & 100 & 368,640 & $-1.04$ / $-1.05$ & $-0.24$ / $-0.25$ \\ %
\quad seed 1 & as seed 0, with its own constant ($c=-3.373$) & 46,080 & 2,211,840 & 100 & 368,640 & $-0.93$ / $-0.94$ & $-0.13$ / $-0.14$ \\ %
\quad seed 2 & as seed 0, with its own constant ($c=-3.979$) & 46,080 & 2,211,840 & 100 & 368,640 & $-0.90$ / $-0.91$ & $-0.02$ / $-0.04$ \\ %
SDTT-8 & one round, teacher the start at 16 steps per block, student 8; reverse KL, lr $6\times10^{-5}$ constant after warmup, EMA 0.999 & 30,720 & 2,211,840 & 500 & 245,760 & $-0.86$ / $-0.86$ & $0.00$ / $0.00$ \\ %
\quad seed 1 & as seed 0; evaluated at step 7,680 & 30,720 & 2,211,840 & 500 & 245,760 & $-0.88$ / $-0.87$ & $-0.03$ / $-0.03$ \\ %
\quad seed 2 & as seed 0; evaluated at step 15,360 & 30,720 & 2,211,840 & 500 & 245,760 & $-0.90$ / $-0.89$ & $-0.01$ / $-0.02$ \\ %
Alpha-2 & as Alpha, at the $K=2$ sampler's states, with the constant measured there ($c=-4.231$); evaluated at step 23,040 & 46,080 & 2,211,840 & 100 & 368,640 & $-0.93$ / $-0.93$ & $-0.07$ / $-0.09$ \\ %
\quad seed 1 & as seed 0, with its own constant ($c=-4.093$); evaluated at step 11,520 & 46,080 & 2,211,840 & 100 & 368,640 & $-0.91$ / $-0.90$ & $-0.05$ / $-0.06$ \\ %
\quad seed 2 & as seed 0, with its own constant ($c=-4.454$); evaluated at step 11,520 & 46,080 & 2,211,840 & 100 & 368,640 & $-0.72$ / $-0.71$ & $+0.13$ / $+0.13$ \\ %
Mix-FT (\cref{tab:repro8x}) & $K=2$, $M=256$ & 357 & 2,210,544 & 44 & 2,856 & $+0.05$ / $+0.08$ & $+0.42$ / $+0.43$ \\ %
\bottomrule
\end{tabularx}
\end{table}

\paragraph{The 1.1B experiments.} \Cref{tab:repro11b} gives the setting. The sampler fills a 256-position canvas in blocks of 16 from left to right; positions after the current block hold the mask token, each step is one full bidirectional forward, and there is no KV cache. %
The bound is IWAE-32 per block on 512 held-out (window, block) pairs. Stage 14 sampled the start and every 1$\times$ model at $K\in\{2,4,8\}$; stage 23 sampled the same models at $K=16$ with the same seeds and no further training; stage 27 trained the 8$\times$ model and sampled it at $K\in\{4,8\}$. %
In stage 27 each job ran on both GPUs of a T4$\times$2. The 2,827 rows of an update are split in two halves. The per-example plan log-likelihoods are summed across the two processes before the plan weights are formed, so the weights are exact, and the LoRA gradients are summed by one all-reduce before clipping and AdamW. The two replicas' parameters were compared every 10 steps and stayed identical. %
Intervals at 1.1B are the bootstrap above with 256 samples per file. %

\begin{table}[h]
\centering
\caption{1.1B setting (SMDM base, LoRA, OpenWebText).}
\label{tab:repro11b}
\footnotesize
\begin{tabularx}{\linewidth}{@{}l>{\raggedright\arraybackslash}X@{}}
\toprule
start & SMDM 1.1B base \citep{nie2025smdm}, \texttt{nieshen/SMDM}, file \texttt{mdm\_safetensors/mdm-1028M-3300e18-rsl-0.01-bs-1024.safetensors}, revision \texttt{f28f1dfe}, sha256 \texttt{1b7378eb}; 1,028M non-embedding parameters, pretrained on SlimPajama with $3.3\times10^{21}$ FLOPs; TinyLlama (Llama-2) tokenizer \\ %
data & OpenWebText: 65,536 training windows of 256 tokens from shard 0 (documents packed with EOS); 2,048 held-out windows from the last 3,000 documents of shard 79; real reference: held-out windows 0 to 1,023 (per-sample entropy 4.7235 nats) \\ %
LoRA & rank 32, alpha 32, all attention and MLP linears (24,084,480 parameters); base frozen in fp16 \\ %
sampler & $L=256$, block 16; FE: a 256-position pass counts 1, or 3 with a backward \\ %
1$\times$ (stage 14) & Mix-FT: $K=2$, $M=8$, 22 examples (window, block) per update, 600 updates, 475,200 FE; continued CE and Alpha ($\kappa=0.5$): 267 rows per update at the $K=4$ states, 600 updates, 480,600 FE; under 0.01\% of pretraining \\ %
8$\times$ (stage 27) & Mix-FT: $K=2$, $M=256$, 11 examples per update, 339 updates, 3,833,412 FE $=7.99\times480{,}000$ (about 0.06\% of pretraining) \\ %
optimizer & AdamW (0.9, 0.999), no weight decay, lr $2\times10^{-4}$, 5\% warmup (30 and 16 steps), cosine to 0, clip 1, fp16 loss scaling, micro-batch 16 rows; window order seed $1000+s$, LoRA init seed $7000+s$ \\ %
sampling & $\beta\in\{0.9,1,1.1,1.2,1.35,1.6\}$, 256 samples per file, batch 64, seeds $4242$ / $5353+100{,}000K+1{,}000\beta$ (identical across models) \\ %
judges & GPT-2-large (fp32, \texttt{openai-community/gpt2-large@212095d5}) and Pythia-1.4B (fp16 after an fp32 check, \texttt{EleutherAI/pythia-1.4b@fedc38a1}) on the decoded text, re-tokenized; content tokens only; matched-entropy levels 4.423 to 4.723 nats \\ %
\bottomrule
\end{tabularx}
\end{table}

\begin{table}[h]
\centering
\caption{The 1.1B SMDM base model fine-tuned with LoRA on OpenWebText (setting in \cref{tab:repro11b}). (a) Means of 3 seeds (start: one model) of the IWAE-32 bound of the $K$-step sampler (nats/token, best of three temperatures), the NELBO perplexity, and the content-token judge NLL at matched per-sample entropy at $K$ steps per block (nats/token, lower is better, 256 samples per temperature). Bold: best entry in the column; --: not measured. (b) $H_8$ is the start's judge NLL minus the model's (positive: 8 steps better), with the seed mean and its 95\% bootstrap interval over samples. MAUVE at matched entropy uses GPT-2-large features of each sample's first 200 tokens; the start at $K=16$ scores 0.765, and real text scored against real text gives 0.829 at this sample size, so MAUVE differences near the top are noisy. Per-seed gains at equal steps are in \cref{tab:perseed11b}.}
\label{tab:results11b}
\footnotesize
\setlength{\tabcolsep}{3pt}
\begin{tabular}{lcccccccccccc}
\toprule
\multicolumn{13}{l}{\emph{(a) seed means}} \\
 & & \multicolumn{2}{c}{sampler bound} & & \multicolumn{4}{c}{GPT-2-large} & \multicolumn{4}{c}{Pythia-1.4B} \\
\cmidrule(lr){3-4}\cmidrule(lr){6-9}\cmidrule(lr){10-13}
model & opt.\ steps & $K=4$ & $K=2$ & NELBO PPL & $K=2$ & 4 & 8 & 16 & $K=2$ & 4 & 8 & 16 \\
\midrule
Start (SMDM base) & -- & 3.279 & 3.868 & 23.47 & \textbf{7.34} & 6.68 & 5.42 & 4.37 & 7.23 & 6.58 & 5.34 & 4.31 \\
Continued CE & 600 & 2.997 & 3.605 & \textbf{17.88} & 7.71 & 6.63 & 5.24 & 4.29 & 7.36 & 6.43 & 5.07 & 4.13 \\
Alpha ($\kappa=0.5$) & 600 & 2.999 & 3.613 & 18.43 & 7.59 & 6.48 & 5.16 & 4.37 & 7.36 & 6.35 & 5.04 & 4.27 \\
Mix-FT 1$\times$ ($M=8$) & 600 & \textbf{2.990} & \textbf{3.598} & 19.37 & 7.45 & 5.99 & 4.46 & \textbf{3.75} & \textbf{7.16} & 5.80 & 4.31 & \textbf{3.61} \\
Mix-FT 8$\times$ ($M=256$) & 339 & 3.024 & 3.629 & 23.00 & -- & \textbf{5.65} & \textbf{4.34} & -- & -- & \textbf{5.48} & \textbf{4.18} & -- \\
\bottomrule
\end{tabular}

\vspace{6pt}
\begin{tabular}{llccccc}
\toprule
\multicolumn{7}{l}{\emph{(b) half the steps: each model at $K=8$ against the start at $K=16$}} \\
model & measure & seed 0 & seed 1 & seed 2 & seed mean & 95\% interval \\
\midrule
Mix-FT 8$\times$ ($M=256$) & $H_8$, GPT-2-large & $+0.090$ & $+0.027$ & $-0.037$ & $+0.027$ & $[-0.107, +0.155]$ \\
 & $H_8$, Pythia-1.4B & $+0.151$ & $+0.149$ & $+0.098$ & $+0.133$ & $[+0.013, +0.242]$ \\
 & MAUVE at $K=8$ & 0.932 & 0.930 & 0.874 & 0.912 & -- \\
\addlinespace[2pt]
Mix-FT 1$\times$ ($M=8$) & $H_8$, GPT-2-large & $-0.013$ & $-0.115$ & $-0.151$ & $-0.093$ & $[-0.213, +0.026]$ \\
 & $H_8$, Pythia-1.4B & $+0.073$ & $-0.023$ & $-0.043$ & $+0.002$ & $[-0.111, +0.111]$ \\
 & MAUVE at $K=8$ & 0.885 & 0.865 & 0.898 & 0.883 & -- \\
\addlinespace[2pt]
Continued CE & $H_8$, GPT-2-large & $-0.760$ & $-0.890$ & $-0.962$ & $-0.871$ & -- \\
 & $H_8$, Pythia-1.4B & $-0.666$ & $-0.769$ & $-0.817$ & $-0.751$ & -- \\
 & MAUVE at $K=8$ & 0.526 & 0.539 & 0.498 & 0.521 & -- \\
\addlinespace[2pt]
Alpha ($\kappa=0.5$) & $H_8$, GPT-2-large & $-0.805$ & $-0.668$ & $-0.896$ & $-0.790$ & -- \\
 & $H_8$, Pythia-1.4B & $-0.754$ & $-0.645$ & $-0.765$ & $-0.721$ & -- \\
 & MAUVE at $K=8$ & 0.606 & 0.581 & 0.514 & 0.567 & -- \\
\bottomrule
\end{tabular}
\end{table}

\begin{table}[h]
\centering
\caption{Per-seed comparisons at 1.1B. Upper part: each arm's content-token judge NLL minus Mix-FT's at the same number of steps per block (nats/token, matched per-sample entropy, 256 samples per temperature; positive: Mix-FT better). The start is one model, compared with each Mix-FT seed; every other arm is paired by seed. Lower part: the IWAE-32 bound of the four-step sampler, each arm minus Mix-FT 1$\times$ (nats/token; positive: Mix-FT 1$\times$ better).}
\label{tab:perseed11b}
\footnotesize
\setlength{\tabcolsep}{3pt}
\begin{tabular}{llcccccccc}
\toprule
 & & \multicolumn{4}{c}{GPT-2-large} & \multicolumn{4}{c}{Pythia-1.4B} \\
\cmidrule(lr){3-6}\cmidrule(lr){7-10}
arm & $K$ & seed 0 & seed 1 & seed 2 & mean & seed 0 & seed 1 & seed 2 & mean \\
\midrule
\multicolumn{10}{l}{\emph{Mix-FT 1$\times$ ($M=8$) against}} \\
Start (SMDM base) & 2 & $-0.008$ & $-0.158$ & $-0.153$ & $-0.106$ & $+0.181$ & $-0.011$ & $+0.042$ & $+0.071$ \\
 & 4 & $+0.883$ & $+0.516$ & $+0.660$ & $+0.686$ & $+0.957$ & $+0.625$ & $+0.766$ & $+0.782$ \\
 & 8 & $+1.039$ & $+0.938$ & $+0.902$ & $+0.960$ & $+1.095$ & $+0.999$ & $+0.979$ & $+1.025$ \\
 & 16 & $+0.693$ & $+0.540$ & $+0.623$ & $+0.619$ & $+0.780$ & $+0.617$ & $+0.705$ & $+0.701$ \\
\addlinespace[2pt]
Continued CE & 2 & $+0.304$ & $+0.246$ & $+0.239$ & $+0.263$ & $+0.249$ & $+0.167$ & $+0.181$ & $+0.199$ \\
 & 4 & $+0.684$ & $+0.518$ & $+0.712$ & $+0.638$ & $+0.677$ & $+0.526$ & $+0.715$ & $+0.639$ \\
 & 8 & $+0.747$ & $+0.775$ & $+0.811$ & $+0.778$ & $+0.739$ & $+0.746$ & $+0.774$ & $+0.753$ \\
 & 16 & $+0.603$ & $+0.405$ & $+0.601$ & $+0.536$ & $+0.565$ & $+0.403$ & $+0.579$ & $+0.516$ \\
\addlinespace[2pt]
Alpha ($\kappa=0.5$) & 2 & $+0.340$ & $+0.076$ & $+0.025$ & $+0.147$ & $+0.395$ & $+0.103$ & $+0.120$ & $+0.206$ \\
 & 4 & $+0.734$ & $+0.250$ & $+0.460$ & $+0.481$ & $+0.810$ & $+0.353$ & $+0.511$ & $+0.558$ \\
 & 8 & $+0.792$ & $+0.553$ & $+0.745$ & $+0.697$ & $+0.827$ & $+0.622$ & $+0.722$ & $+0.724$ \\
 & 16 & $+0.862$ & $+0.512$ & $+0.474$ & $+0.616$ & $+0.883$ & $+0.556$ & $+0.540$ & $+0.660$ \\
\midrule
\multicolumn{10}{l}{\emph{Mix-FT 8$\times$ ($M=256$) against}} \\
Start (SMDM base) & 4 & $+1.131$ & $+1.058$ & $+0.901$ & $+1.030$ & $+1.184$ & $+1.121$ & $+0.979$ & $+1.095$ \\
 & 8 & $+1.143$ & $+1.080$ & $+1.016$ & $+1.080$ & $+1.173$ & $+1.172$ & $+1.120$ & $+1.155$ \\
\addlinespace[2pt]
Continued CE & 4 & $+0.932$ & $+1.059$ & $+0.953$ & $+0.981$ & $+0.904$ & $+1.022$ & $+0.927$ & $+0.951$ \\
 & 8 & $+0.850$ & $+0.917$ & $+0.925$ & $+0.898$ & $+0.817$ & $+0.919$ & $+0.915$ & $+0.883$ \\
\addlinespace[2pt]
Mix-FT 1$\times$ ($M=8$) & 4 & $+0.248$ & $+0.542$ & $+0.241$ & $+0.344$ & $+0.227$ & $+0.496$ & $+0.213$ & $+0.312$ \\
 & 8 & $+0.104$ & $+0.142$ & $+0.114$ & $+0.120$ & $+0.078$ & $+0.173$ & $+0.141$ & $+0.130$ \\
\bottomrule
\end{tabular}

\vspace{6pt}
\begin{tabular}{lcccc}
\toprule
\multicolumn{5}{l}{\emph{sampler bound at $K=4$, arm minus Mix-FT 1$\times$}} \\
arm & seed 0 & seed 1 & seed 2 & mean \\
\midrule
Start (SMDM base) & $+0.2882$ & $+0.2903$ & $+0.2868$ & $+0.2884$ \\
Continued CE & $+0.0075$ & $+0.0089$ & $+0.0038$ & $+0.0068$ \\
Alpha ($\kappa=0.5$) & $+0.0086$ & $+0.0104$ & $+0.0082$ & $+0.0091$ \\
Mix-FT 8$\times$ ($M=256$) & $+0.0269$ & $+0.0379$ & $+0.0368$ & $+0.0339$ \\
\bottomrule
\end{tabular}
\end{table}

\begin{table}[h]
\centering
\caption{The three rater contrasts. Each pair shows one complete sample of 1,024 tokens from each side, and every pair is judged in both display orders. For each model the temperature is the grid point whose per-sample entropy is closest to that of real text, 5.397 nats. A control pairs a model's text at that temperature with the same model's text at its hottest grid point, $\beta=0.95$. In c3 the controls use the start, because at $K=16$ the 8$\times$ model's hottest text was too close in quality to its matched text to test a rater.}
\label{tab:contrasts}
\footnotesize
\setlength{\tabcolsep}{3pt}
\begin{tabularx}{\linewidth}{@{}l>{\raggedright\arraybackslash}X>{\raggedright\arraybackslash}X>{\raggedright\arraybackslash}p{3.3cm}>{\raggedright\arraybackslash}p{2.4cm}>{\raggedright\arraybackslash}p{2.3cm}>{\raggedright\arraybackslash}p{2.4cm}r@{}}
\toprule
contrast & test side & reference side & temperatures & per-sample entropy (nats) & main pairs & control pairs & judgments \\
\midrule
c1 & 8$\times$ model at $K=8$ & start at $K=16$ & each model's own per-step schedule (the same shape for all four models), $\beta=1.15$ & test 5.431 to 5.444, reference 5.434 & 512, drawn at random from 1,536 (171, 171 and 170 per seed) & 192 on the 8$\times$ model at $K=8$ (64 per seed) & 1,408 \\
c0 & start at $K=8$ & start at $K=16$ & the same schedule for both, $\beta=1.15$ & test 5.403, reference 5.434 & 256, drawn at random from 512 & 64 on the start at $K=8$ & 640 \\
c3 & 8$\times$ model at $K=16$ & start at $K=16$ & one global temperature each: $\beta=1$ for the 8$\times$ model (every seed), $\beta=1.05$ for the start & test 5.385 to 5.414, reference 5.457 & 192 (64 per seed) & 96 on the start at $K=16$ & 576 \\
\bottomrule
\end{tabularx}
\end{table}

\paragraph{Blinded pairwise raters.} Every pair shows two complete unconditional samples of 1,024 GPT-2 tokens, decoded exactly, each end-of-text token shown as \texttt{[document break]} on its own line; nothing is excerpted. %
\Cref{tab:contrasts} defines the three contrasts, c1 (the 8$\times$ model at $K=8$ against the start at $K=16$), c0 (the start against itself at $K=8$ and 16) and c3 (the 8$\times$ model against the start, both at $K=16$), with their temperature settings and controls. In c3 the start's text is 0.04 to 0.07 nats hotter in per-sample entropy on this coarse grid, which leans toward Mix-FT, but at exactly matched entropy the judges still put Mix-FT 0.62 to 0.67 nats/token ahead. The rerun with the first token fixed on both sides uses the same pairs, orders and files with new opaque ids, each side at the grid point the same rule picks on the fixed-convention samples: the temperatures of \cref{tab:contrasts} except the start's in c3 ($\beta=1.1$). There the start's text is 0.02 to 0.04 nats hotter in c1 (5.471 against 5.427 to 5.451), which leans slightly toward Mix-FT, and 0.065 to 0.095 nats colder in c3 (5.322 against 5.387 to 5.417), which leans toward the start. %
A fixed derangement pairs sample $i$ of one side with sample $\pi(i)\neq i$ of the other, so no pair shares a sampling seed; the three seeds' pairs that share a start sample form one cluster. Each pair is judged in both display orders by separate rater sessions, and the stored order is a fair coin per pair. Controls pair a model's matched-temperature text with the same model's hottest grid point, $\beta=0.95$. %
A verdict scores 1 if the test side (the fine-tuned model in c1 and c3, the start at $K=8$ in c0) is preferred, $\frac12$ for a tie and 0 otherwise; a pair's score is the mean of its two orders, and $W$ is the mean over pairs. %
Intervals are 95\% percentile bootstraps with 10,000 draws: over clusters for the pooled $W$ (ratio estimator), over pairs within a seed, and over control pairs for the control rate $W_c$. %
A rater counts only if the lower bound of $W_c$ exceeds 0.60. On c1 the reading is ``noninferior'' if the lower bound of $W$ exceeds 0.45, ``inconclusive'' if it lies in $(0.40,0.45]$ with the upper bound at least 0.45, and ``worse'' otherwise; on c3 it is ``preferred'' if the lower bound exceeds 0.50. Each rule was fixed before any text of its contrast was rated. %
The share of the halving gap recovered is
\[
\frac{W_{\mathrm{c1}}-W_{\mathrm{c0}}}{0.5-W_{\mathrm{c0}}},
\]
with an interval from the two bootstrap standard errors (normal approximation, c0 and c1 treated as independent). %
\Cref{tab:reprorater} lists the runs. The ChatGPT runs gave each file of 32 to 44 judgments to a fresh agent with no shared memory; the Claude Sonnet runs gave each batch of judgments (44 in c1, 40 in c0, 36 per file in c3) to a separate helper; order-1 and order-2 batches were disjoint. %
The rater instructions below head every ChatGPT file and every file of the Claude runs we use (here for a 44-judgment file); in c1 and c0 the Claude helpers rated the very files ChatGPT rated. An earlier Claude run of c1 and c0 worked from the pilot's version of this prompt, which we did not archive word for word, and also asked for a short reason. We decided before the re-run that its numbers would replace that run's in the text whatever they showed, and we list the earlier run in \cref{tab:reprorater} for reference. %

{\scriptsize\ttfamily\raggedright
\begin{quote}
You are a careful, impartial judge of English writing quality. This is a blinded rating task for a research study.\par\smallskip
This file holds 44 judgments. Each judgment shows Text A and Text B. Both were produced by language models trained on English web text. Each excerpt is about the same length (1,024 tokens), may begin or end in the middle of a sentence, and may contain the marker [document break] where one document ends and another begins. Do not reward or penalize an excerpt for its topic, for starting or ending abruptly, or for containing a document break. Judge which excerpt is better written, using these four criteria, weighted equally:\newline
1. Coherence: the text stays on topic and makes sense from sentence to sentence and from paragraph to paragraph.\newline
2. Readability: the sentences are grammatical, fluent, and easy to follow.\newline
3. Repetition: the text is free of loops, repeated phrases, and degenerate lists (less repetition is better).\newline
4. Substance: the text conveys concrete, specific, and consistent content rather than vague, contradictory, or empty filler.\newline
The order in which the excerpts are shown is random; do not let it influence you. If the two excerpts are of essentially equal quality overall, answer "tie".\par\smallskip
Read both texts of every judgment in full and judge each judgment on its own merits. Do not use code, scripts or statistics to decide; the verdicts must be your own reading.\par\smallskip
Answer with exactly 44 lines of JSON, one per judgment, in order, and nothing else:\newline
\{"jid": "...", "coherence": "A|B|tie", "readability": "A|B|tie", "repetition": "A|B|tie", "substance": "A|B|tie", "overall": "A|B|tie"\}
\end{quote}}

\noindent The open-weights Qwen2.5-7B-Instruct run (greedy, fp16) used the same rubric as a single-pair chat prompt that also asked for a reason of one or two sentences. %

\begin{table}[h]
\centering
\caption{Rater runs, 21 to 31 October 2026 (contrasts in \cref{tab:contrasts}). ChatGPT names are the product names shown in its interface; the Claude helpers ran \texttt{claude-sonnet-5-5}, at maximum effort in every run we use. The upper block rates samples with the first token fixed on both sides, the start's convention; the lower blocks rate the earlier samples with it drawn. (a) $W$: test side preferred, ties half, with its 95\% interval; per seed: $W$ for fine-tune seeds 0, 1 and 2 (c0 compares the start with itself and has no seeds); $W_c$: control rate; cons.: share of pairs with the same outcome in both display orders; agr.: share of the contrast's judgments on which the Claude and ChatGPT runs give the same overall verdict; share: the share of the halving gap recovered, $(W_{\mathrm{c1}}-W_{\mathrm{c0}})/(0.5-W_{\mathrm{c0}})$, with its 95\% interval. (b) $W$ in c3 on each criterion of the rubric, with its 95\% interval.}
\label{tab:reprorater}
\footnotesize
\setlength{\tabcolsep}{3pt}
\begin{tabularx}{\linewidth}{@{}>{\raggedright\arraybackslash}p{2.5cm}llllcc>{\raggedright\arraybackslash}X@{}}
\toprule
\multicolumn{8}{@{}l}{\emph{(a) all runs}} \\
rater & contrast & $W$ & per seed & $W_c$ & cons. & agr. & reading \\
\midrule
\multicolumn{8}{@{}l}{\emph{first token fixed (27 to 31 October)}} \\
Claude Sonnet 5.5 helpers & c1 & 0.428 [0.385, 0.472] & 0.485, 0.415, 0.384 & 0.831 [0.783, 0.878] & 0.78 & 76\% & worse; share 0.84 [0.74, 0.94] \\
 & c0 & 0.055 [0.032, 0.081] & -- & 0.984 [0.953, 1.000] & 0.95 & 95\% & \\
 & c3 & 0.786 [0.710, 0.855] & 0.848, 0.781, 0.730 & 0.987 [0.971, 1.000] & 0.88 & 89\% & preferred \\
\addlinespace[2pt]
ChatGPT ``6.1 Sol Ultra'' & c1 & 0.468 [0.425, 0.513] & 0.515, 0.455, 0.435 & 0.850 [0.802, 0.896] & 0.80 & 76\% & inconclusive; share 0.93 [0.82, 1.03] \\
 & c0 & 0.072 [0.046, 0.102] & -- & 0.984 [0.953, 1.000] & 0.93 & 95\% & \\
 & c3 & 0.758 [0.680, 0.829] & 0.805, 0.770, 0.699 & 0.964 [0.927, 0.990] & 0.87 & 89\% & preferred \\
\addlinespace[2pt]
both raters pooled & c1 & 0.448 [0.409, 0.490] & & 0.841 & & & share 0.88 [0.79, 0.97] \\
 & c0 & 0.064 [0.041, 0.089] & & 0.984 & & & \\
\midrule
\multicolumn{8}{@{}l}{\emph{first token drawn (21 to 26 October)}} \\
Qwen2.5-7B-Instruct (\texttt{a09a3545}) & c1 & 0.500 [0.498, 0.503] & 0.499, 0.501, 0.502 & 0.503 [0.495, 0.510] & 0.003 & -- & fails control (second text chosen in 99.2\%); rated all 1,536 main pairs \\
\addlinespace[2pt]
Claude Sonnet 5.5 helpers & c1 & 0.416 [0.371, 0.462] & 0.450, 0.401, 0.396 & 0.888 [0.848, 0.926] & 0.82 & 79\% & worse; share 0.79 [0.68, 0.91] \\
 & c0 & 0.091 [0.059, 0.127] & -- & 0.902 [0.832, 0.961] & 0.96 & 95\% & \\
 & c3 & 0.882 [0.820, 0.934] & 0.879, 0.906, 0.859 & 0.898 [0.844, 0.948] & 0.94 & 90\% & preferred \\
\addlinespace[2pt]
ChatGPT ``Astra 6 Ultra'' & c1 & 0.502 [0.455, 0.547] & 0.550, 0.484, 0.472 & 0.893 [0.849, 0.932] & 0.81 & 79\% & noninferior; share 1.00 [0.89, 1.12] \\
 & c0 & 0.089 [0.057, 0.122] & -- & 0.930 [0.859, 0.984] & 0.97 & 95\% & \\
\addlinespace[2pt]
ChatGPT ``6.1 Sol Ultra'' & c3 & 0.878 [0.823, 0.927] & 0.898, 0.871, 0.863 & 0.909 [0.854, 0.956] & 0.91 & 90\% & preferred \\
\addlinespace[2pt]
both raters pooled (mean of the four verdicts of a pair) & c1 & 0.459 [0.416, 0.501] & & 0.891 & & & share 0.90 [0.80, 1.00] \\
 & c0 & 0.090 [0.059, 0.123] & & 0.916 & & & \\
\midrule
Claude Sonnet 5.5, pilot's prompt (superseded) & c1 & 0.453 [0.408, 0.498] & 0.491, 0.436, 0.431 & 0.876 [0.835, 0.914] & 0.77 & 78\% & inconclusive; share 0.88 [0.76, 0.99] \\
 & c0 & 0.106 [0.073, 0.143] & -- & 0.895 [0.820, 0.957] & 0.93 & 93\% & \\
\midrule
\multicolumn{8}{@{}l}{ChatGPT re-rating 192 of its own earlier judgments gave the same verdict on 90\%.} \\
\bottomrule
\end{tabularx}

\vspace{6pt}
\begin{tabular}{lcccc}
\toprule
\multicolumn{5}{l}{\emph{(b) c3 by criterion}} \\
rater & coherence & readability & substance & repetition \\
\midrule
\multicolumn{5}{l}{\emph{first token fixed}} \\
Claude Sonnet 5.5 helpers & 0.772 [0.699, 0.837] & 0.824 [0.767, 0.876] & 0.728 [0.659, 0.792] & 0.483 [0.422, 0.546] \\
ChatGPT ``6.1 Sol Ultra'' & 0.738 [0.656, 0.815] & 0.876 [0.816, 0.928] & 0.738 [0.656, 0.815] & 0.514 [0.427, 0.602] \\
\multicolumn{5}{l}{\emph{first token drawn}} \\
Claude Sonnet 5.5 helpers & 0.867 [0.809, 0.918] & 0.908 [0.855, 0.951] & 0.831 [0.772, 0.882] & 0.474 [0.410, 0.540] \\
ChatGPT ``6.1 Sol Ultra'' & 0.872 [0.816, 0.922] & 0.928 [0.878, 0.969] & 0.859 [0.803, 0.909] & 0.457 [0.375, 0.542] \\
\bottomrule
\end{tabular}
\end{table}

\noindent Pooled over the two raters, the share of the halving gap recovered is 0.88 $[0.79,0.97]$ with the first token fixed and 0.90 $[0.80,1.00]$ with it drawn, and the two raters give the same verdict on 76 to 95\% of the judgments of each contrast (\cref{tab:reprorater}). Fixing the first token lowers both raters' equal-step preference from 88 to 76--79\%, as it lowers the judges' margin: it helps the start more than the fine-tuned model (\cref{app:conv}). For Claude the model, prompt and effort of the two runs are the same, so its change comes from the samples alone. With the archived prompt Claude rates c1 0.04 lower than with the pilot's prompt; its two runs give the same verdict on 83\% of the c1 judgments, little more than either gives with ChatGPT, and on 96\% of the c0 judgments, so the c1 pairs are close calls for every rater. %

\paragraph{Fine-tuning time.} \Cref{tab:reprotime} gives measured training hours on T4 GPUs, summed over kernels from each kernel's trainer start line to its budget stop or end of training (model loading and evaluation excluded). The 8$\times$ Mix-FT run took 1.45 times as long as 8$\times$ continued CE, mostly because its chunked backward executes about 1.4 times the counted FE. The block-chunked backward of \cref{app:hparams} executes only the counted FE: on a T4, one 8$\times$ update of eight sequences takes 272 seconds with it instead of 397, which would bring the 8$\times$ run to about 27 hours, the time of 8$\times$ continued CE. %

\begin{table}[h]
\centering
\caption{Measured fine-tuning time (T4 GPU-hours, training only; seeds 0, 1, 2 where run). At 110M every job used one T4 except the 16$\times$ jobs, which used two. At 1.1B the 1$\times$ jobs used one T4 and the 8$\times$ jobs two; at 8B every job used two, with the model split by layers. Jobs on two T4s count wall-clock hours $\times$ 2.}
\label{tab:reprotime}
\footnotesize
\begin{tabular}{lllr}
\toprule
model & run & budget & T4-hours \\
\midrule
110M & Mix-FT, $K=2$, $M=8$ (1,152 steps) & 1$\times$ & 3.85, 3.38, 3.76 \\ %
110M & continued CE (5,760 steps) & 1$\times$ & 3.10, 3.58, 3.37 \\ %
110M & SDTT-8 (3,840 steps) & 1$\times$ & 3.35, 3.30, 3.28 \\ %
110M & Mix-FT, $K=2$, $M=256$ (357 steps) & 8$\times$ & 38.9, 37.7, 38.0 \\ %
110M & Mix-FT, $K=2$, $M=512$ (358 steps) & 16$\times$ & 49.7, 48.5, 49.2 \\ %
110M & continued CE (46,080 steps) & 8$\times$ & 26.8, 26.0, 25.4 \\ %
110M & Alpha (46,080 steps) & 8$\times$ & 26.5, 25.5, 26.1 \\ %
110M & SDTT-8 (30,720 steps) & 8$\times$ & 25.2, 25.0, 25.2 \\ %
1.1B & Mix-FT, $K=2$, $M=8$ (600 steps) & 1$\times$ & 6.47, 7.04, 6.93 \\ %
1.1B & continued CE (600 steps) & 1$\times$ & 8.49, 7.94, 7.61 \\ %
1.1B & Mix-FT, $K=2$, $M=256$ (339 steps) & 8$\times$ & 59.6, 56.1, 59.8 \\ %
8B & Mix-FT, $K=4$, $M=8$ (600 steps) & 1$\times$ & 73.3, 74.5, 75.4 \\ %
8B & continued CE (600 steps) & 1$\times$ & 78.6, 80.0, 78.9 \\ %
8B & Alpha (600 steps) & 1$\times$ & 77.7, 83.5, 79.4 \\ %
\bottomrule
\end{tabular}
\end{table}

\FloatBarrier
\section{Experimental decisions and outcomes}
\label{app:rules}

We wrote each expectation in a version-controlled log before the experiment ran, together with its code, arms and compute. The table gives, in order, what we expected, what happened, and where the paper reports the evidence; stage numbers are the log's. $G$ is a matched-diversity judge gap (positive: arm better); $G_c$ is its content-token version at matched per-sample entropy; $\Delta$ is a likelihood difference, arm minus reference (negative: arm better); $H_8$ is $G_c$ of a model at $K=8$ against the start at $K=16$. Unless noted, three seeds.

In the table, C0 is the model after masked-CE pretraining only, Base~$n\times$ is C0 followed by $n$ units of CE at the sampler's states (stage~1 continued plain masked CE instead), Mix is the mixture objective trained from C0, and the hybrid adds the CE loss on Mix-FT's forward passes. Q-numbers name the questions within a stage, and a kill condition is the outcome, fixed in advance, that rejects the expectation it names. The coverage guard asks whether, where Mix wins at matched distinct-4, its samples still cover at least as much of real held-out text's 4-gram occurrences as the baseline's, within half a percentage point (stage~2b also accepted a win at matched coverage), so that the gain is not bought with lost real 4-grams. The share of a per-path control is the part of Mix-FT's gain over continued CE that the control recovers, on the $K=4$ likelihood in rows 10 and 13A and on the judge gain in row 30.

{\small
\begin{xltabular}{\linewidth}{@{}l>{\raggedright\arraybackslash}X>{\raggedright\arraybackslash}X>{\raggedright\arraybackslash}p{2.1cm}@{}}
\toprule
stage & question and what we expected & what happened & evidence \\
\midrule
\endfirsthead
\toprule
stage & question and what we expected & what happened & evidence \\
\midrule
\endhead
1 & Text8 pilot ($B=8$, $K=2$, 2 seeds, weaker Base): Mix beats Base on IWAE-64 by $\ge0.02$\bpc{} without losing $>10\%$ distinct-4. & Pass (0.055, 0.063\bpc{}). Later stages used a Base trained at the sampler's states. & -- \\
2 & Mix from C0 against C0, Base, SDTT at $B\in\{8,16\}$: $G>0$ in all seeds and mean $G\ge0.05$; every baseline must learn. & Invalid: SDTT did not learn. Without SDTT the pass condition held. & -- \\
2b & Fixed SDTT; kill ``Mix is under-trained'' if the Mix minus Base bound gap at 4$\times$ is not smaller than at 1$\times$; coverage guard. & Pass; explanation rejected: on the $K=4$ IWAE-64 bound, averaged over blocks 8 and 16, the gap is $+0.020$\bpc{} at 1$\times$ and $+0.076$ at 4$\times$ (exact values at block 8 in \cref{tab:perseed}); guard holds. & \cref{tab:perseed} \\
2c & Two-stage recipe ($B=8$, $K=4$, exact): some arm with $\Delta\le+0.005$ and $G>0$ against Base 5$\times$ in all seeds. & Pass for Mix-FT and the hybrid. & \cref{tab:perseed} \\
3 & Q1 Mix-FT vs SDTT (exact, $B=8$); Q2a $B=16$; Q2b $K=8$; pass if $\Delta<0$ (Q2: $\le+0.005$) and $G>0$ in all seeds. & Pass, pass, pass; coverage guard: no caveat. & \cref{tab:perseed} \\
3, Q3 & 34M FineWeb-Edu from Base 1$\times$ (one seed, reported only). & No verdict ($\Delta=+0.022$ for both arms). & \cref{fig:scale} \\
4 & 34M from Base 2$\times$: Mix-FT ($M=4$) and hybrid vs Base 3$\times$, $\Delta\le+0.005$ nats/token and $G>0$ in all seeds. & Pass for both. & \cref{fig:scale} \\
5 & 110M BD3-LM: Mix-FT and hybrid vs continued CE and SDTT; $\Delta_{K=4}\le+0.005$ and all-token $G>0$ (mean over $K=2,4,8$) in all seeds. & Mix-FT passes against both; hybrid passes against CE. The all-token $G$ against SDTT depends on newline targets, so only the likelihood part of that comparison is claimed. & \cref{tab:scale} \\
6, Q6B & Mix-FT on two-step plans: $G_{K=2}>0$ against SDTT (all tokens) and $\Delta_{K=2}\le+0.005$ against CE in all seeds; kill if mean $G_{K=2}\le0$. & Kill (mean $G_{K=2}=-0.093$), although $\Delta_{K=2}=-0.041$ in every seed. & \cref{tab:main110} \\
7, Q7a & Four-step Mix-FT vs SDTT and CE at $K=8$, 16 (512 samples): $G_c>0$ under both judges in every seed. & Mix-FT better in all four comparisons. & \cref{app:110m} \\
7, Q7b & Mix-FT on top of SDTT vs more SDTT: $\Delta_{K=4}\le-0.005$ and $G_c>0$ at $K=4$, 8 in every seed. & No verdict ($\Delta=-0.063$ every seed; one seed slightly negative at $K=4$). & \cref{app:110m} \\
8 & Alpha diffusion, $\kappa\in\{0.5,2,8\}$: likelihood, and samples per $K$ against the best $\kappa$. & Four-step Mix-FT better on likelihood; samples mixed at $K=2$, Alpha better at $K=4$, Mix-FT better at $K=8$, 16. & \cref{tab:scale} \\
9c & 1.1B GSM8K, fine-tuning on the model's verified solutions: Mix-FT minus CE accuracy $>0$ in every seed with a bootstrap interval excluding 0. & No difference (\cref{app:gsm8k}). & \cref{app:gsm8k} \\
10 & Text8 mechanism control: weighting is ``the main cause'' if Mix-FT beats per-path FT in every seed and per-path FT recovers at most half of Mix-FT's likelihood gain over continued CE. & Main cause (per-path FT recovers a share of 0.24 on average); samples: Mix-FT better. & \cref{tab:control} \\
11 & 110M matched budgets (512 samples): four-step Mix-FT vs SDTT and Alpha at $K=4$, vs SDTT-8 at $K=8$, 16. & Mixed at $K=4$ against both; Mix-FT better than SDTT-8 at $K=8$, 16 and on likelihood. & \cref{app:110m} \\
12, Q12a & Two-step Mix-FT vs Alpha, SDTT-4 ($K=4$), SDTT-8 ($K=8$, 16) and four-step Mix-FT, at $K=4$, 8, 16: $G_c>0$ under both content judges in every seed. & Two-step Mix-FT better in all nine readings. & \cref{tab:main110} \\
12, Q12b & Seed 0: $d=G_c$(two-step, $M=8$) minus $G_c$(four-step, $M=3$, 91\% of the updates) at $K=4$; ``two steps matter'' if $d>0.05$ under both judges. & Two steps matter ($d=+0.48$ under both). & \cref{tab:controls110}a \\
12E & Seed 0, equal updates: two-step at 443 steps vs four-step Mix-FT, and two-step vs four-step at 1,152 steps; $d>0.05$ under both judges at $K=4$. & Two steps matter in both ($+0.25$, $+0.25$). & \cref{tab:controls110}a \\
13A & 110M per-path control: ``main cause'' if Mix-FT beats per-path FT in every seed and per-path FT recovers at most half of Mix-FT's gain over CE; samples at $K=4$, 8. & Main cause (share $-0.14$); samples: Mix-FT better at both. & \cref{tab:controls110}c \\
13B, C & IWAE-64 at $K=16$, Mix-FT minus CE per seed; 30 bound claims re-read at the validation temperature. & Mix-FT better than CE; no claim changes. & \cref{app:110m} \\
13E, 31 & Text8 width test, rule on three-seed means: the one-token-per-step part of the gain shrinks with width (by at least 0.008\bpc{}), the gap part does not. & Supported. Against continued CE, the rule's comparator, the first part falls from 0.024 to 0.013\bpc{} from width 256 to 768 (by 0.010 to 0.011 in each seed) and the second moves from 0.018 to 0.019; against CE trained at the $K=8$ states the split is the same. & \cref{tab:t8abl}a \\
14 & 1.1B SMDM base on OpenWebText: two-step Mix-FT vs continued CE, $\Delta_{K=4}\le+0.005$ and $G_c>0$ under both judges in every seed. & Pass ($\Delta$ from $-0.004$ to $-0.009$; $G_c$ from $+0.48$ to $+0.59$). & \cref{tab:results11b} \\
15B & 110M, two-step Mix-FT with $M=32$ against $M=8$ at equal compute: $G_c>0$ under both judges at $K=4$, 8, 16 in every seed. & $M=32$ better in all nine ($+0.18$, $+0.24$, $+0.16$). & \cref{tab:controls110}b \\
16, 16b & Seed 0: more plans against $M=32$ at $K=8$, at equal compute ($M=64$, 128) and with compute scaled to keep about 350 updates ($M=64$ at 2$\times$, $M=128$ at 4$\times$). & Equal compute saturates ($+0.03$, no gain); with compute more plans help (about $+0.06$ per doubling). & \cref{tab:controls110}b \\
17, 19 & Per-step temperature schedules, chosen on a development split for every model and the start: $K=8$ matches the start at $K=16$ ($H_8\ge0$ under both judges in every seed). & No match ($H_8=-0.135$, $-0.119$); the schedule helps every model, the start included, by $+0.23$ to $+0.28$ on fresh samples (eight-temperature grid). & \cref{tab:controls110}d \\
18 & Importance proposal over plans instead of uniform plans. & Worse ($-0.39$); dropped. & -- \\
20 & $M=256$ at 8$\times$ compute: $H_8\ge0$ under both judges in every seed (one global temperature). & Match in all three ($+0.002$ to $+0.053$; Pythia $+0.025$ to $+0.076$), with the first token drawn (row 36). & \cref{tab:h8} \\
21 & The same with each model's own schedule. & Match in all three ($+0.016$ to $+0.025$; Pythia $+0.032$ to $+0.045$), with the first token drawn (row 36). & \cref{tab:h8} \\
23 & 1.1B, the 1$\times$ two-step fine-tunes of stage 14 sampled at $K=16$: Mix-FT at $K=8$ against the start at $K=16$, $H_8\ge0$ under both judges in every seed. & No match in any seed ($-0.15$ to $+0.07$). At equal steps ($K=16$) Mix-FT beats the start by 0.54 to 0.78 and continued CE by 0.40 to 0.60, in every seed under both judges. & \cref{tab:results11b} \\
22 & Seed 0: continued CE and Alpha diffusion at 8$\times$ compute; compute alone closes the gap if their $H_8\ge0$ under both judges. & Does not close it: $H_8$ $-1.24$ (CE), $-1.04$ (Alpha); Mix-FT beats both at $K=4$, 8, 16. Seeds 1 and 2 (reported): $H_8$ $-1.21$, $-1.22$ for CE and $-0.93$, $-0.90$ for Alpha; Mix-FT beats both in the same seed by 0.78 to 1.25 at each $K$. & \cref{tab:repro8xbase} \\ %
25 & Seed 0: SDTT-8 (teacher passes counted, development-selected checkpoint) and Alpha diffusion on two-step states, at 8$\times$; same readings as 22. & Neither closes it ($H_8$ $-0.86$ SDTT-8, $-0.93$ Alpha-2); Mix-FT beats both at $K=4$, 8, 16. Seeds 1 and 2 (reported): $H_8$ $-0.88$, $-0.90$ for SDTT-8 and $-0.91$, $-0.72$ for Alpha-2; Mix-FT beats both in the same seed by 0.68 to 0.94 at each $K$. & \cref{tab:repro8xbase} \\
27 & 1.1B, $M=256$ at 8$\times$: $H_8\ge0$ under both judges in every seed; seed-mean $H_8$ within the $-0.03$ margin under both judges (\cref{app:repro}). & Not met in every seed: seeds 0 and 1 match, seed 2 is $-0.04$ under GPT-2-large and $+0.10$ under Pythia; seed means $+0.03$ and $+0.13$. Margin shown under Pythia, not under GPT-2-large (interval $[-0.11,+0.15]$). MAUVE higher in every seed. & \cref{tab:results11b} \\
29 & Bounds of the 8$\times$ model with up to 256 plans at $K=8$, 16 against continued CE (1$\times$, 8$\times$) and the start: sign per seed and whether the gap shrinks from 64 to 256 plans. & Above continued CE (by 0.03 at $K=16$; within 0.004 at $K=8$), below the start at $K=8$; the gap shrinks with more plans in every seed. & \cref{tab:bounds8x} \\
28 & 110M, LLaDA's low-confidence planner, matched entropy: the 8$\times$ model at $K=8$ against the start at $K=16$ ($\ge0$ under both judges in every seed), and Mix-FT against every baseline at $K=4$, 8, 16. & Match ($+0.15$ to $+0.19$; Pythia $+0.20$ to $+0.22$) with the first token drawn, and $+0.02$ to $+0.04$ (Pythia $+0.04$ to $+0.07$) when both fix it (row 38); Mix-FT better in all 24 readings. & \cref{tab:mauve110} \\
30 & Seed 0 controls: (a) Mix-FT on eight-step plans, placed against the two- and four-step recipes ($\pm0.05$); (b) two-step per-path control, ``main cause'' if it recovers under half of Mix-FT's gain over CE at $K=4$, 8, 16; (c) continued CE at $0.1\times$ the learning rate and with EMA weights: does Mix-FT still win widely? & (a) Behind two-step plans at every $K$; level with four-step plans at $K=8$, 16, behind at $K=2$, 4. (b) Main cause: per-path FT ends behind CE (share $-0.41$ to $-0.07$). (c) Mix-FT still wins by 0.51 to 0.81; both variants stay behind the start. & \cref{tab:controls110} \\
33 & Blocks 32 and 64, before training: is the plan-mixture bound still usable (median ESS of eight plans $\ge1.25$ and two further gates)? Block 32 then fine-tuned anyway as an exploratory run: Mix-FT at $K=4$, 1$\times$, better than continued CE if $G_c>0$ under both judges in every seed; at 8$\times$ (seed 0), $H\ge0$ under both judges at $K=16$ against the start at $K=32$, and Mix-FT better than continued CE at 8$\times$. & No-go at both blocks (at block 32 the few-step candidates $K=2$, 4 have ESS 1.06, 1.13). Fine-tune: Mix-FT better at $K=4$, 8 and 16 ($+0.11$ to $+0.48$), and in seed 0 better than the start at each $K$ under both conventions ($+0.14$ to $+0.30$), where continued CE is behind it at $K=8$ and 16. At 8$\times$: half-step match at $K=16$ with the first token drawn ($+0.02$, $+0.05$), not when both fix it, the start's convention ($-0.10$, $-0.09$), and not at $K=4$ or 8; better than continued CE 8$\times$ at $K=4$, 8, 16 ($+0.55$ to $+0.91$). & \cref{tab:block32} \\
35 & LLaDA-8B-Base, block 32, LoRA, 0.0018\% of pretraining per arm, three seeds, LLaDA's two planners (\cref{app:8b}). (R1) Mix-FT better than continued CE and Alpha diffusion at $K=4$, 8, 16 under GPT-2-large, Pythia-1.4B and Qwen2.5-7B in every seed. (R2, R2-32) Per planner, Mix-FT at $K=8$ against the start at $K=16$, and at $K=16$ against the start at $K=32$: match ($\ge0$ under the three judges in every seed), level (seed mean $\ge-0.03$ and 95\% upper bound $\ge0$ under each) or not reached. & R1: pass at every planner, $K$, judge and seed ($+0.12$ to $+0.62$). R2: not reached under either planner ($-0.06$ to $-0.35$ per seed, 66 to 91\% of the start's own gap closed; continued CE and Alpha diffusion $-0.48$ to $-0.91$ under GPT-2-large and Pythia). R2-32: match under both planners ($+0.05$ to $+0.19$ per seed). & \cref{tab:results8b} \\
36 & The first token: BD3-LM never masks position 0 in training and its sampler fixes it to end-of-text, while ours draws it. Every model of the main comparison sampled both ways with the same random draws; the rule fires if fixing it helps the start by more than 0.05 under both judges at some $K$, and then fixes the consequences: one convention in every table, each claim against the start stated under the convention that favors the start with both numbers, the rater studies rerun. Also a null fine-tune: BD3-LM's own objective continued for the same budget on its raw windows (three seeds), expected level with or ahead of the start, with Mix-FT ahead of it, continued CE behind it, the window convention immaterial and our comparison unchanged on raw windows. & Fires: fixing the first token helps the start by 0.05 to 0.11 at $K=8$ and 16 (it costs 0.04 to 0.07 at $K=4$) and moves the fine-tuned models by at most 0.07. $H_8$ changes sign in every seed under both judges, with one temperature and with schedules ($-0.01$ to $-0.07$); at equal steps the 8$\times$ model still beats the start by 0.53 to 0.58. In the rater rerun the 8$\times$ model at $K=8$ recovers 88\% of the halving loss (90\% before), and at equal steps the raters prefer it in 76 and 79\% of pairs (88\% before). The null is behind the start by 0.18 to 0.43 at every $K$ under both conventions (its EMA weights too), so not level or ahead as expected; Mix-FT is ahead of it by 0.50 to 0.96; continued CE against it is mixed. The window convention is not immaterial, though by the rule it is convention-dependent: on our windows the null is better in every seed, by 0.11 to 0.34 with the first token drawn, and with it fixed by 0.06 to 0.38 at $K=4$ and 8 but level in one seed at $K=16$. On our windows it is level with the start on average at $K=4$ (seed 2 behind by 0.06 to 0.10) and behind it at $K=8$ and 16 in every seed, by 0.04 to 0.28. On raw windows Mix-FT's gains move by more than 0.10, so ``unchanged'' fails: over continued CE they grow by 0.12 to 0.18 at $K=8$, and over the start they shrink by 0.10 at $K=4$ under Pythia with the first token drawn. & \cref{app:conv}, \cref{tab:controls110}g, \cref{tab:reprorater} \\
38 & Row 36's consequence for the claims against the start that row 36 did not cover, with the first token fixed for both models. (L1) Under the confidence planner, the 8$\times$ model at $K=8$ beats the start at $K=16$ if its gain is positive under both judges in every seed; (L2) the 1$\times$ model beats the start at each $K$ under that planner (seed 0); (D1) the 8$\times$ model at $K=8$ is more diverse than the start at $K=16$ if its Vendi score is higher and its self-BLEU-4 lower in every seed; at block 32 (seed 0), (B1) the 8$\times$ model at $K=16$ matches the start at $K=32$ if its gain is $\ge0$ under both judges, (B2) the 1$\times$ model beats the start and (B3) continued CE trails it at each $K$. & L1 holds under both conventions ($+0.02$ to $+0.07$ fixed, $+0.15$ to $+0.22$ drawn). L2 better at $K=4$, 8 and 16 ($+0.29$ to $+0.48$). D1 holds. B1 does not hold under the start's convention ($-0.10$, $-0.09$; $+0.02$, $+0.05$ drawn). B2 better at every $K$ ($+0.14$ to $+0.25$); B3 behind at $K=8$ and 16. With its seeds 1 and 2 added, the 1$\times$ model beats the start at $K=4$, 8 and 16 in every seed under both judges (\cref{app:conv}). & \cref{app:conv} \\
39 & The 16$\times$ rung, three seeds: the 8$\times$ recipe with $M=512$ and 16$\times$ the budget (358 updates, 4,416,288 FE, 0.25\% of pretraining), read under both first-token conventions. (R1) ``Match'' if $H_8$ is $\ge0$ under both judges in every seed; ``level on average'' if not, but the 95\% interval of the seed mean lies above $-0.03$ under both judges. & Drawn: match ($+0.06$ to $+0.10$; Pythia $+0.09$ to $+0.13$). Fixed: no match; the seed means are $-0.005$ under GPT-2-large and $+0.016$ under Pythia, and the GPT-2-large interval, $[-0.036,+0.028]$, misses the margin by 0.006. Each seed gains 0.03 to 0.08 over 8$\times$ under either convention. & \cref{tab:h8} \\
40 & Text8, exact likelihood, run after the 110M results. (T1) Mix-FT on two-step plans against the four-step recipe at equal compute, both evaluated with the $K=4$ sampler: ``two-step ahead'' if its exact $K=4$ NLL is lower and its samples better (judge gap $G$) in every seed. We were unsure: at $B=8$ a two-step plan reveals half a block per step. (M1) The exact $K=4$ NLL along $M=8$, 32, 128 and 256 plans at fixed updates: expected to keep improving or to flatten, not to get worse at the top. Reported: continued CE given the compute of $M=256$ (seed 0). & T1: two-step ahead in every seed (0.002 to 0.005\bpc{}; $G$ $+0.06$ to $+0.11$). At equal updates its likelihood is 0.005\bpc{} worse in every seed and its samples still better. M1: improves, then flattens ($M=256$ against $M=128$: mean $-0.002$). Continued CE with the compute of $M=256$: much lower exact NLL, worse samples. & \cref{tab:t8abl}b, c \\
41 & Blinded pairwise raters on the 8B half-step result, before any of its samples were drawn: Mix-FT at $K=16$ against LLaDA at $K=32$ with low-confidence decoding, new samples at each model's matched temperature, 288 pairs over three seeds, both orders. ``Preferred'' if the 95\% lower bound of Mix-FT's win rate exceeds 0.50, ``noninferior'' if it exceeds 0.45; a rater whose control lower bound is at most 0.60 is not read. & ChatGPT: preferred, 0.69 $[0.63, 0.75]$, every seed's lower bound above 0.50; control 0.84 $[0.72, 0.95]$. Claude Sonnet: preferred, 0.71 $[0.66, 0.77]$, every seed's lower bound above 0.50; control 0.84 $[0.73, 0.95]$. & \cref{tab:8b} \\
\bottomrule
\end{xltabular}
}

\paragraph{How the two-step model became the main model.} Stage 6 trained it to close SDTT's lead at $K=2$, and that rule was killed. A later look at its 128-sample files showed it ahead of Alpha diffusion and the four-step model at $K=2$, 4 and 8. Stage 12 then tested it on the 512-sample protocol with the expectation above written down first. The 512-sample files at $K=2$, 4 and 8 and temperatures 1 and 1.1 reuse the sampling seeds of the 128-sample files, so their first 128 samples are the ones already seen. All 36 such files, the two-step model's at the three step counts and the $K=2$ files of the four-step Mix-FT, SDTT-4 and Alpha diffusion, each at both temperatures and in three seeds, reproduce their 128-sample files exactly. The other four temperatures and all of $K=16$ are new.

\end{document}